\PassOptionsToPackage{dvipsnames,table}{xcolor}
\documentclass{article} % For LaTeX2e
\usepackage{iclr2026_conference,times}

\usepackage{amsmath,amsfonts,bm}

\def\eqref#1{equation~\ref{#1}}
\def\1{\bm{1}}

\DeclareMathAlphabet{\mathsfit}{\encodingdefault}{\sfdefault}{m}{sl}
\SetMathAlphabet{\mathsfit}{bold}{\encodingdefault}{\sfdefault}{bx}{n}

\usepackage{hyperref}
\usepackage{url}
\usepackage{booktabs}
\usepackage{graphicx}
\usepackage{array}
\usepackage{adjustbox}
\usepackage{caption}
\usepackage{subcaption}
\usepackage{enumitem}
\usepackage{tikz}
\usepackage{float}
\usepackage{wrapfig}
\usepackage{tcolorbox}
\tcbuselibrary{breakable}
\usepackage{dblfloatfix}
\usepackage{booktabs,longtable}

\title{TopoBench: Benchmarking LLMs on Hard Topological Reasoning}

\author{%
\textbf{Mayug Maniparambil}$^{*,1}$\thanks{Equal contribution. Corresponding author: \texttt{mayug.maniparambil@intercom.io}} \quad
\textbf{Nils Hoehing}$^{*,2}$ \quad
\textbf{Janak Kapuriya}$^{3}$ \quad
\textbf{Arjun Karuvally}$^{4}$ \\
\textbf{Ellen Rushe}$^{5}$ \quad
\textbf{Anthony Ventresque}$^{6}$ \quad
\textbf{Noel O'Connor}$^{5}$ \quad
\textbf{Fergal Reid}$^{1}$ \\[6pt]
\begin{minipage}{\linewidth}
{\small
$^1$Intercom Research \quad
$^2$University College Dublin \quad
$^3$University of Galway \quad \\
$^4$ Salk Institute \quad
$^5$Dublin City University \quad
$^6$Trinity College Dublin
}
\end{minipage}
}

\iclrfinalcopy % Uncomment for camera-ready version, but NOT for submission.
\begin{document}
\raggedbottom

\maketitle
\lhead{Workshop on Logical Reasoning of Large Language Models at ICLR 2026}

\begin{abstract}
% Solving topological grid puzzles requires reasoning over global spatial invariants such as connectivity, loop closure, and region symmetry and remains challenging for even the most powerful large language models (LLMs). To study these abilities under controlled settings, we introduce TopoBench, a benchmark of six puzzle families across three difficulty levels. We evaluate strong reasoning LLMs on TopoBench and find that even frontier models solve fewer than one quarter of hard instances, with two families nearly unsolved. To further understand the failure modes, we annotate 750 chain of thought traces with an error taxonomy that surfaces four candidate causal failure modes, then test them with targeted interventions simulating each error type. These interventions show that certain error patterns like premature commitment (going down a wrong solution path early on) and constraint forgetting (moves that violate rules) have a direct impact on the ability to solve the puzzle while repeated-reasoning (re-trying the same reasoning path without meaningful variation) is a benign effect of search. Finally we study mitigation strategies aimed at reducing these error patterns including prompt guidance, cell-aligned grid representations and tool-based constraint checking, finding that the bottleneck lies in extracting constraints from spatial representations and not in reasoning over them.

Solving topological grid puzzles requires reasoning over global spatial invariants such as connectivity, loop closure, and region symmetry and remains challenging for even the most powerful large language models (LLMs). To study these abilities under controlled settings, we introduce TopoBench, a benchmark of six puzzle families across three difficulty levels. We evaluate strong reasoning LLMs on TopoBench and find that even frontier models solve fewer than one quarter of hard instances, with two families nearly unsolved. To investigate whether these failures stem from reasoning limitations or from difficulty extracting and maintaining spatial constraints, we annotate 750 chain of thought traces with an error taxonomy that surfaces four candidate causal failure modes, then test them with targeted interventions simulating each error type. These interventions show that certain error patterns like premature commitment (going down a wrong solution path early on) and constraint forgetting (moves that violate rules) have a direct impact on the ability to solve the puzzle, while repeated reasoning (re-trying the same reasoning path without meaningful variation) is a benign effect of search. Finally we study mitigation strategies including prompt guidance, cell-aligned grid representations and tool-based constraint checking, finding that the bottleneck lies in extracting constraints from spatial representations and not in reasoning over them.
Code and data are available at \url{github.com/mayug/topobench-benchmark}.

\end{abstract}

\section{Introduction}
\label{sec:introduction}

% Recent LLMs have achieved strong performance on algebraic~\citep{cobbe2021gsm8k,hendrycks2021math}, symbolic~\citep{suzgun2022bigbenchhard} and textual reasoning benchmarks \citep{wang2019superglue, nie-etal-2020-adversarial}. However they struggle on tasks that require maintaining global spatial invariants such as topological connectivity or geometric logic through a sequence of state updates ~\citep{toshniwal2022chess,chollet2019arc}. Retaining a global spatial understanding through sequences of local updates arises in circuit layout, route planning, and molecular structure analysis, where a single violated constraint can invalidate an entire solution. 
% Therefore topology focused grid puzzles
% % whose solutions depend on global spatial constraints such as path/network connectivity, loop closure, rotational symmetry and contiguity over a two-dimensional grid, 
% provide a natural test bed for this capability, which is underrepresented in current puzzle benchmarks ~\citep{long2025puzzleplex,ren2025vgrp,tyagi2024gridpuzzle,giadikiaroglou2024puzzlesurvey}. Moreover current puzzle benchmarks typically only report accuracy, but rarely disentangle whether model failures arise from the reasoning process itself or from limitations in representing and manipulating spatial information.

% Auto-generated by benchpress_adapter/run_predictability_table.py
% Requires \usepackage{booktabs,longtable}

Recent LLMs have achieved strong performance on algebraic~\citep{cobbe2021gsm8k,hendrycks2021math}, symbolic~\citep{suzgun2022bigbenchhard} and textual reasoning benchmarks \citep{wang2019superglue, nie-etal-2020-adversarial}. However, they struggle on tasks that require maintaining global spatial invariants such as topological connectivity or geometric logic through a sequence of state updates~\citep{toshniwal2022chess,chollet2019arc}. Retaining a global spatial understanding through sequences of local updates arises in circuit layout, route planning, and molecular structure analysis, where a single violated constraint can invalidate an entire solution. 
Despite their practical importance, the ability of LLMs to reason over global constraints is under-studied in current benchmarks~\citep{long2025puzzleplex,ren2025vgrp,tyagi2024gridpuzzle,giadikiaroglou2024puzzlesurvey}, and topology-focused grid puzzles provide a natural test bed to evaluate this capability while abstracting away the domain-specific details. Furthermore, current puzzle benchmarks typically only report accuracy, but rarely disentangle whether model failures arise from the reasoning process itself or from limitations in representing and manipulating spatial information. We address both gaps in this work.

\begin{figure}[]
\centering
\resizebox{0.7\textwidth}{!}{%
\includegraphics{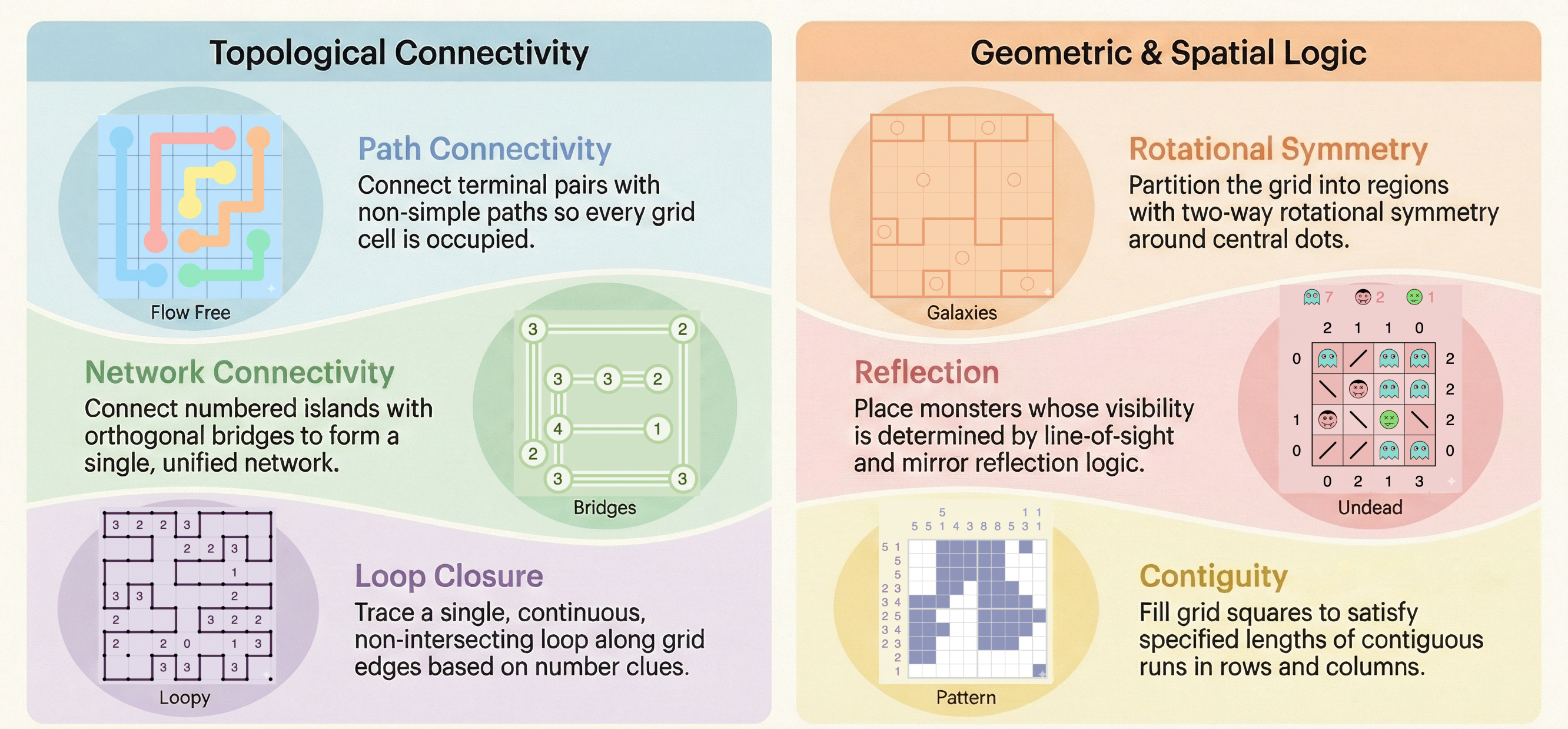}%
}
\caption{The six TopoBench puzzle families organized by the global spatial constraint each targets: path connectivity (Flow Free), network connectivity (Bridges), loop closure (Loopy), region partitioning under rotational symmetry (Galaxies), visibility through reflection (Undead), and contiguity across intersecting axes (Pattern).}
\label{fig:topobench}
\vspace{-1.5em}
\end{figure}

We introduce \textbf{TopoBench}, a benchmark comprising six topology-focused puzzle families. Each puzzle family is selected to test specific topological/geometric constraints such as connectivity (Flow Free, Bridges), loop closure (Loopy), symmetry (Galaxies), reflection (Undead) or contiguity (Pattern). We provide three difficulty tiers per puzzle family to study how performance degrades with task complexity. Rule-based verifiers are included to check correctness and solvers which expose intermediate steps are employed to enable fine-grained diagnosis. We evaluate nine reasoning LLMs including two leading closed source models and seven open source ones. Even the strongest model tested (GPT-5-mini-high) achieves just 0.24 accuracy on the hard tier, while the best open-weight model (DeepSeek V3.2) reaches just 0.10. 

To identify the factors underlying reasoning failure, we conduct a two-stage diagnosis that combines observational frequency with causal interventions. First, using an LLM-as-judge protocol, we annotate reasoning traces to classify distinct error types and estimate their frequency. Second, we perform intervention experiments by injecting these error patterns into partial gold solution prefixes and measure the resulting change in downstream accuracy. We find that premature commitment and constraint forgetting each produce substantial accuracy drops when injected (approximately 11\,pp on Bridges and Undead). In contrast, repeated reasoning is among the most frequent patterns observed in traces yet has no measurable causal effect on downstream accuracy. Constraint forgetting, by comparison, appears in only 2--7\% of traces but is among the most damaging interventions. Overall, we find that the frequency of an error is a poor predictor of how much harm it actually causes.

% Finally, we evaluate mitigations targeting each diagnosed bottleneck. Cell-aligned integer representations improve robustness by aligning tokenization with grid structure, improving accuracies across most puzzles families. To isolate whether the reasoning failures arise from initial parsing or from constraint extraction during multi-step reasoning, we introduce a tool-augmented setting in which an external engine maintains the board state and provides structured constraint information as a tool call. On Bridges hard difficulty, we see that accuracy improves by 10\% with structured constraint information. Our ablation studies reveal that these gains are driven by access to structured fields rather than spatial renderings i.e. adding a rendered ASCII grid degrades accuracy, whereas structured summaries of remaining degrees and connectivity yield consistent improvements. Together, these results indicate that the primary bottleneck lies in extracting structured constraint information from spatial representations rather than in reasoning over them. We also test prompt-level changes that encourage planning and backtracking to reduce premature commitment errors, but do not see any meaningful improvements.

Finally, we evaluate mitigations targeting each diagnosed error type. First, we test input representations that always tokenise each input row into an equal number of tokens and find that it improves accuracies across most puzzle families. Second, to isolate whether the reasoning failures arise from initial parsing or from constraint extraction during multi-step reasoning, we introduce a tool-augmented setting in which an external engine maintains the board state and provides structured constraint information as a tool call. On Bridges hard difficulty, we see that accuracy improves by 10\% with structured constraint information. Our ablation studies show that these gains are driven by access to structured state information rather than spatial renderings — specifically, adding a current ASCII-grid tool degrades accuracy, whereas structured summaries of remaining degrees and connectivity yield consistent improvements. Together, these results indicate that the primary bottleneck lies in extracting structured constraint information from spatial representations rather than in reasoning over them. Third, we test prompt-level changes that encourage planning and backtracking to reduce premature commitment errors, but do not see any meaningful improvements indicating that such behaviors are not reliably elicited through prompting alone.

\textbf{Contributions:}
\begin{enumerate}[nosep]
\item \textbf{TopoBench}, a benchmark of six topology-focused puzzle families
across three difficulty tiers with puzzle-specific verifiers, evaluated on nine
frontier and open-weight reasoning models.
\item A \textbf{diagnostic pipeline} that classifies errors in model solution
traces and tests their causal role through controlled interventions on gold
solution prefixes.
\item \textbf{Targeted mitigations} (cell-aligned encodings, structured
constraint queries, and prompt-level interventions), with ablations
identifying spatial constraint extraction as the primary bottleneck.
\end{enumerate}
%\input{sections/introduction_v4}
%\input{sections/introduction_v3}

% ============================================
\section{Related Work}
\label{sec:related_work}
% ============================================

\textbf{Reasoning benchmarks and puzzle evaluation.}
Existing reasoning benchmarks typically test mathematical or logical reasoning while few explicitly focus on topological/geometric logic. GSM8K~\citep{cobbe2021gsm8k}, MATH~\citep{hendrycks2021math} test algebraic reasoning, while SATBench~\citep{satbench2025} tests multi-step logical reasoning. ARC~\citep{chollet2025arcprize} and BIG-Bench Hard~\citep{suzgun2022bigbenchhard} target abstraction and compositional generalization, while
game settings such as chess~\citep{toshniwal2022chess,harang2025chessstate} and Sokoban~\citep{sokobench2026} focus on planning under sequential state updates, but do not isolate topological/geometric reasoning. Notably, \citet{estermann2024puzzles} trained RL algorithms on the full suite of Simon Tatham puzzles and found that topology-focused puzzles like Loopy were intractable for the algorithms tested. Recently, grid-based puzzle benchmarks have emerged as a test-bed for evaluating LLMs on structured spatial reasoning tasks. Sudoku-Bench~\citep{sudokubench2025} evaluates Latin-square variants,
CrossWordBench~\citep{leng2025crosswordbench} targets language-constrained grids,
and VGRP-Bench~\citep{vgrpbench2025}, Enigmata~\citep{chen2025enigmata}, and KORGym~\citep{shi2025korgym} provide visual-grid and interactive evaluation settings~\citep[see also][]{giadikiaroglou2024puzzlesurvey}. These benchmarks typically test local pattern matching or cell-level arithmetic, but do not require maintaining global invariants such as path connectivity, loop closure, or region symmetry across an entire grid.
GridPuzzle~\citep{tyagi2024gridpuzzle} introduced fine-grained error taxonomies with automated annotation of reasoning chains, but stops at observational classification.
In contrast, TopoBench specifically targets topological constraints and goes further by pairing observational error annotation with causal validation through controlled interventions on gold solution prefixes.

\textbf{Diagnosing reasoning failures.}
Chain-of-thought prompting~\citep{wei2022cot} and its zero-shot variant~\citep{kojima2022zeroshotcot} have made intermediate reasoning steps visible, enabling post-hoc analysis of where and how models fail.
Two lines of work use the chain of thought for error diagnosis and robustness testing. First, LLM-as-a-judge protocols~\citep{g_eval,prometheus} provide scalable annotation of reasoning traces, while perturbation-based methods such as Contrast Sets~\citep{gardner2020contrastsets} test robustness by applying controlled edits to inputs. Our diagnostic pipeline combines both, using LLM-judge annotation to build an observational error taxonomy and controlled interventions on gold solution path prefixes to test whether each error type is causally linked to downstream accuracy.
By design, the two stages separate error frequency from causal impact, a distinction that purely observational taxonomies cannot make.

\textbf{Representation sensitivity and tool augmentation.}
Model performance can be highly sensitive to input representation format. For instance, digit tokenization affects arithmetic~\citep{singh2024tokenization,nogueira2021arithmetic}, and single-character format changes can cause large accuracy swings~\citep{su2025singlechar}.
We extend this line of inquiry to topological puzzles by testing whether input representations that tokenize evenly across grid rows reduce parsing errors and hence improve performance.
Rather than only changing how inputs are encoded, tool-augmented reasoning frameworks such as ReAct~\citep{yao2023react}, Toolformer~\citep{schick2024toolformer}, and ART~\citep{paranjape2023art} offload computation to external tools entirely. We use tool augmentation to disentangle state tracking, and constraint calculation to understand exactly where LLM reasoning failure occurs when solving topological puzzles.
This question is motivated by work on state tracking in language models, from procedural text~\citep{rezaee2025statetracking} and dialogue~\citep{budzianowski2018multiwoz} to board representations in Othello-GPT~\citep{li2023othello,nanda2023othello}, where performance consistently improves when structured state extraction is offloaded or bypassed.

\section{Benchmark and Methodology}
\label{sec:benchmark}

We introduce TopoBench, a puzzle benchmark that targets multi-step \emph{topological} and \emph{geometric} reasoning in LLMs. Models must maintain and update global spatial invariants like path connectivity, region partitioning and visibility propagation through purely textual reasoning within its chain-of-thought. We do not allow access to external code execution or solvers as the goal is to test inherent topological/geometric reasoning and not coding ability or tool-based computation. The benchmark is designed such that even frontier reasoning models find hard instances challenging, but skilled humans find solving them routine. We then apply a diagnostic pipeline to identify the sources of model failures on TopoBench, and test targeted mitigations to address them; these are described in Section~\ref{sec:experiments}.

% \textbf{Task Families.} We select six puzzle families that collectively span six distinct classes of global spatial constraint (Table~\ref{tab:puzzles}; Figure~\ref{fig:topobench}). The selection is deliberately heterogeneous. Bridges requires maintaining network connectivity under degree and crossing constraints while connecting numbered isla, Flow Free demands routing non-intersecting paths that collectively fill the grid, Galaxies enforces region partitioning under global rotational symmetry, Undead requires multi-step visibility reasoning through reflections, Pattern enforces contiguity constraints across intersecting row and column axes, and Loopy requires constructing a single closed loop consistent with per-cell edge counts. These constraint types target distinct global invariants that rarely co-occur in existing reasoning or puzzle benchmarks.

 \textbf{Task Families.} We select six puzzle families that cover different categories depending on the global constraints that must be satisfied to solve them (Figure~\ref{fig:topobench}). The selection is intentionally heterogeneous to enable a diverse evaluation of topological and geometric constraints. Bridges requires connecting numbered islands with bridges such that the number of bridges satisfies the number denoted on each island and all islands are connected without any loops (maintaining network connectivity under degree and crossing constraints). FlowFree is solved by connecting similarly colored dots without any paths intersecting each other (path connectivity). Galaxies requires expanding regions while requiring each region to be rotationally symmetric around the center (rotational symmetry). 
Undead is solved by filling the grid with monsters satisfying total monster counts and visibility constraints placed on the row/column edges. While Undead might not initially appear to be a topological puzzle, the presence of mirrors introduces visual structure, requiring the model to track lines of sight through multiple reflections (reflection and visibility). For Pattern, one needs to fill a binary grid with contiguous elements such that constraints on row/column edges are satisfied (contiguity across axes). Finally, Loopy requires constructing a single closed loop such that each cell has a number of edges that equal the specified number (loop closure). Table~\ref{tab:puzzles} summarizes the task descriptions and spatial constraints. 

\begin{table}[]
\centering
\small
\caption{TopoBench puzzle families and the global spatial constraint each targets.}
\label{tab:puzzles}
\resizebox{0.9\textwidth}{!}{%
\begin{tabular}{lp{6.5cm}l}
\toprule
\textbf{Puzzle} & \textbf{Task description} & \textbf{Spatial constraint} \\
\midrule
Flow Free            & Route non-intersecting paths between color-matched endpoints, filling every cell.          & Path connectivity \\
Bridges (Hashiwokakero) & Connect numbered islands with bridges; degree, crossing, and connectivity constraints.    & Network connectivity \\
Galaxies (Tentai Show)  & Partition grid into regions each rotationally symmetric around a marked center.            & Rotational symmetry \\
Undead               & Place monsters so line-of-sight counts through mirrors match clues.                         & Reflection \& visibility \\
Pattern (Nonogram)   & Fill a binary grid to match row/column run-length clues.                                    & Contiguity across axes \\
Loopy (Slitherlink)  & Draw a single closed loop on grid edges satisfying per-cell edge-count clues.               & Loop closure \\
\bottomrule
\end{tabular}}
\vspace{-0.7em}
\end{table}

\begin{table}[]
\centering
\small
\caption{Models evaluated on TopoBench. \emph{Active} denotes active parameters for MoE models. Architecture: D\,=\,dense, MoE\,=\,mixture-of-experts, Diff\,=\,diffusion, $\dagger =$  Not publicly disclosed.}
\label{tab:models}
\resizebox{0.9\textwidth}{!}{%
\begin{tabular}{llccccc}
\toprule
\textbf{Model} & \textbf{Source} & \textbf{Params (Active)} & \textbf{Arch.} & \textbf{Open-weight} & \textbf{Reasoning} & \textbf{Multimodal} \\
\midrule
\multicolumn{7}{l}{\emph{Reasoning models}} \\
GPT-5-mini-high         & OpenAI    & $\dagger$       & $\dagger$ & --  & \checkmark & \checkmark \\
Gemini-3-Flash-Preview  & Google    & $\dagger$       & $\dagger$   & --  & \checkmark & \checkmark \\
DeepSeek V3.2           & DeepSeek  & 671B (37B)      & MoE   & \checkmark & \checkmark & -- \\
Qwen3-32B               & Alibaba   & 32B             & D     & \checkmark & \checkmark & -- \\
OLMo-3.1-32B-Think      & AI2       & 32B             & D     & \checkmark & \checkmark & -- \\
Qwen3-235B-A22B-Thinking-2507 & Alibaba & 235B (22B) & MoE & \checkmark & \checkmark & -- \\
\midrule
\multicolumn{7}{l}{\emph{Non-reasoning baselines}} \\
LLaMA-4-Maverick        & Meta      & 400B (17B)      & MoE   & \checkmark & --         & -- \\
Mercury                 & Inception & $\dagger$       & Diff  & -- & --         & -- \\
GLM-4.7-Flash           & Zhipu AI  & 30B (3B)       & MoE & \checkmark & \checkmark         & -- \\
\bottomrule
% \multicolumn{7}{l}{\footnotesize $^\dagger$~Not publicly disclosed.}
\end{tabular}%
}
\vspace{-1.5em}
\end{table}

\textbf{Dataset Construction.} We construct 50 puzzles per family and difficulty (easy, medium, hard) for a total of \textbf{900} instances. Puzzle difficulty is controlled along two complementary axes. The first is board size: most families use $5{\times}5$, $7{\times}7$, and $10{\times}10$ for easy, medium, and hard respectively, ranging up to $12{\times}12$ for hard Flow Free.
Undead is the exception, using smaller boards ($4{\times}4$, $5{\times}5$, $7{\times}7$) because its constraint structure causes complexity to grow faster with grid size.
The second axis is the generator's internal difficulty dial, which governs the deductive depth needed to solve an instance without backtracking. By scaling both axes simultaneously, we ensure that harder tiers demand deeper spatial reasoning, not merely larger grids. We de-duplicate within each puzzle$\times$tier split, ensuring that no two instances share an identical board configuration or solution.
Full engine parameter strings for every puzzle$\times$tier combination appear in Table~\ref{tab:engine_params} (Appendix).

\textbf{Human Reference.}
The PUZZLES benchmark reports that a human expert solves 100\% of puzzles at their easiest-human presets~\citep{estermann2024puzzles}.
Our easy tier uses comparable or lower difficulty settings, so these instances are well within reach for experienced human solvers.
This makes the models' low accuracies on the easy-tier  all the more striking: puzzles that humans find routine remain challenging for most models we evaluate.

\textbf{Input Representation.}
Similar to \citep{chen2025enigmata,shi2025korgym}, puzzles are presented as ASCII plain-text grids, a format that preserves spatial layout but requires models to parse two-dimensional structure from a linear token stream.
We treat this as the default encoding and explore alternatives (integer-based and multimodal formats) as interventions in Section~\ref{subsec:format-intervention}.

\textbf{Evaluated Models.} We evaluate nine models spanning closed- and open-source families, dense and mixture-of-experts architectures, and scales from 17B active to 671B total parameters (Table~\ref{tab:models}). The selection prioritizes reasoning-capable models, as TopoBench’s multi-step spatial reasoning drives non-reasoning models to near-zero accuracy on medium and hard tiers. Within this focus, we include models with strong performance on established reasoning benchmarks (AIME, MATH, ARC-AGI) and various architectures (dense, MoE, diffusion) to assess whether topology-heavy reasoning depends on model design. We additionally include LLaMA-4-Maverick as non-reasoning baseline to quantify the gap.

\textbf{Prompting.} We use one-shot prompting: each prompt includes a complete rule specification for the puzzle family and a single worked example showing input and final solution in the same ASCII format. The worked example is fixed across all instances of a given family, and no intermediate reasoning traces are provided, so models must infer their own solving strategy from the rules alone. We allow up to \textbf{100\,k tokens} per evaluation to accommodate extended chains of thought.
Full prompts for each puzzle family are provided in Appendix~\ref{app:eval_prompts}.

\textbf{Execution.}
We evaluate models using single-attempt (pass@1) scoring with all models run with their respective default decoding hyperparameters. Reasoning mode is activated at the highest available level where supported.
GPT-5-mini-high, Gemini-3-Flash, DeepSeek and Qwen3-235B are accessed via their native APIs while the remaining models are accessed via OpenRouter and Groq with providers pinned for reproducibility.
Crucially, \emph{no external code execution is permitted} as we are interested in measuring reasoning in the chain of thought alone and not the ability to offload computation to external solvers.

\textbf{Verification.}
Since the puzzles can have multiple solutions, each puzzle type has a dedicated constraint-checking verifier. Verification is binary (correct/incorrect) with no partial credit.
Models output a JSON object encoding their proposed solution and the verifier reconstructs the grid and checks it against the full constraint set.
To maximize extraction rates, we parse leniently: extracting JSON from markdown code fences, repairing malformed output via the \texttt{json\_repair} library, and falling back to Python literal evaluation when standard parsing fails.
Instances that yield no valid grid are scored as incorrect.

\section{Experiments and Analysis}
\label{sec:experiments}

\begin{table}[]
\centering
\small
\caption{Accuracy (proportion of solved puzzles) on TopoBench by model and puzzle type across difficulties. Cell colors use a low-to-high heat scale over accuracy values (lighter to darker/greener indicates higher accuracy). Rows with avg accuracy $\leq$ 0.01 are omitted for readability. Full results in Tables \ref{tab:accuracy_full_easy},\ref{tab:accuracy_full_medium},\ref{tab:accuracy_full_hard} (Appendix).}
\label{tab:accuracy_combined_v2}
\resizebox{0.82\textwidth}{!}{%
\begin{tabular}{lrrrrrrr}
\toprule
Model & FlowFree & Bridges & Loopy & Galaxies & Undead & Pattern & Avg \\
\midrule
\multicolumn{8}{l}{\textbf{Easy} (n=50)} \\
\midrule
Gemini 3 Flash & \cellcolor[HTML]{AFE7D2} 0.72 & \cellcolor[HTML]{89D5B4} 0.86 & \cellcolor[HTML]{FFECC9} 0.18 & \cellcolor[HTML]{FFEECF} 0.16 & \cellcolor[HTML]{9FDFC5} 0.78 & \cellcolor[HTML]{7FD0AB} 0.90 & \cellcolor[HTML]{C7E7CA} 0.60 \\
GPT-5 Mini & \cellcolor[HTML]{BCE9D2} 0.66 & \cellcolor[HTML]{7FD0AB} 0.90 & \cellcolor[HTML]{FBE0A8} 0.32 & \cellcolor[HTML]{CBE7C8} 0.58 & \cellcolor[HTML]{84D3B0} 0.88 & \cellcolor[HTML]{79CEA7} 0.92 & \cellcolor[HTML]{B2E8D4} 0.71 \\
DeepSeek V3.2 & \cellcolor[HTML]{C0E8CF} 0.64 & \cellcolor[HTML]{74CBA3} 0.94 & \cellcolor[HTML]{FFE4B1} 0.26 & \cellcolor[HTML]{FFFAF3} 0.04 & \cellcolor[HTML]{A4E2CA} 0.76 & \cellcolor[HTML]{94DABD} 0.82 & \cellcolor[HTML]{CBE6C7} 0.58 \\
Qwen3-235b& \cellcolor[HTML]{DDE4BC} 0.48 & \cellcolor[HTML]{AFE7D2} 0.72 & \cellcolor[HTML]{FFFFFF} 0.00 & \cellcolor[HTML]{FFFFFF} 0.00 & \cellcolor[HTML]{FFE6B7} 0.24 & \cellcolor[HTML]{E5E3B7} 0.44 & \cellcolor[HTML]{FCE0A7} 0.31 \\
GLM-4.7 Flash & \cellcolor[HTML]{FFF0D5} 0.14 & \cellcolor[HTML]{FFFFFF} 0.00 & \cellcolor[HTML]{FFFAF3} 0.04 & \cellcolor[HTML]{FFFFFF} 0.00 & \cellcolor[HTML]{FFFFFF} 0.00 & \cellcolor[HTML]{FFFCF9} 0.02 & \cellcolor[HTML]{FFFBF5} 0.03 \\
Mercury & \cellcolor[HTML]{FFF4E1} 0.10 & \cellcolor[HTML]{FFEAC3} 0.20 & \cellcolor[HTML]{FFFFFF} 0.00 & \cellcolor[HTML]{FFFFFF} 0.00 & \cellcolor[HTML]{FFFCF9} 0.02 & \cellcolor[HTML]{FFF6E7} 0.08 & \cellcolor[HTML]{FFF8EB} 0.07 \\
OLMo 3.1 32B & \cellcolor[HTML]{FFE6B7} 0.24 & \cellcolor[HTML]{FFF6E7} 0.08 & \cellcolor[HTML]{FFFCF9} 0.02 & \cellcolor[HTML]{FFFFFF} 0.00 & \cellcolor[HTML]{FFFFFF} 0.00 & \cellcolor[HTML]{FFF6E7} 0.08 & \cellcolor[HTML]{FFF7EA} 0.07 \\
Qwen3-32B & \cellcolor[HTML]{FFECC9} 0.18 & \cellcolor[HTML]{FFF8ED} 0.06 & \cellcolor[HTML]{FFFFFF} 0.00 & \cellcolor[HTML]{FFFFFF} 0.00 & \cellcolor[HTML]{FFFCF9} 0.02 & \cellcolor[HTML]{FFEECF} 0.16 & \cellcolor[HTML]{FFF7EA} 0.07 \\
\midrule
\multicolumn{8}{l}{\textbf{Medium} (n=50)} \\
\midrule
Gemini 3 Flash & \cellcolor[HTML]{D2E6C3} 0.54 & \cellcolor[HTML]{BCE9D2} 0.66 & \cellcolor[HTML]{FFFFFF} 0.00 & \cellcolor[HTML]{FFFFFF} 0.00 & \cellcolor[HTML]{FFF4E1} 0.10 & \cellcolor[HTML]{9FDFC5} 0.78 & \cellcolor[HTML]{F6E1AB} 0.35 \\
GPT-5 Mini & \cellcolor[HTML]{FFE8BD} 0.22 & \cellcolor[HTML]{AFE7D2} 0.72 & \cellcolor[HTML]{FFFFFF} 0.00 & \cellcolor[HTML]{FFFFFF} 0.00 & \cellcolor[HTML]{99DDC1} 0.80 & \cellcolor[HTML]{7FD0AB} 0.90 & \cellcolor[HTML]{E5E3B7} 0.44 \\
DeepSeek V3.2 & \cellcolor[HTML]{FFE8BD} 0.22 & \cellcolor[HTML]{84D3B0} 0.88 & \cellcolor[HTML]{FFFFFF} 0.00 & \cellcolor[HTML]{FFFFFF} 0.00 & \cellcolor[HTML]{E1E4B9} 0.46 & \cellcolor[HTML]{C0E8CF} 0.64 & \cellcolor[HTML]{F2E1AE} 0.37 \\
Qwen3-235b& \cellcolor[HTML]{FFECC9} 0.18 & \cellcolor[HTML]{F7E1AA} 0.34 & \cellcolor[HTML]{FFFFFF} 0.00 & \cellcolor[HTML]{FFFFFF} 0.00 & \cellcolor[HTML]{FFFCF9} 0.02 & \cellcolor[HTML]{FFEAC3} 0.20 & \cellcolor[HTML]{FFF2DA} 0.12 \\
% GLM-4.7 Flash & \cellcolor[HTML]{FFFFFF} 0.00 & \cellcolor[HTML]{FFFFFF} 0.00 & \cellcolor[HTML]{FFFCF9} 0.02 & \cellcolor[HTML]{FFFFFF} 0.00 & \cellcolor[HTML]{FFFFFF} 0.00 & \cellcolor[HTML]{FFFFFF} 0.00 & \cellcolor[HTML]{FFFEFE} 0.00 \\
% OLMo 3.1 32B & \cellcolor[HTML]{FFFFFF} 0.00 & \cellcolor[HTML]{FFFFFF} 0.00 & \cellcolor[HTML]{FFFFFF} 0.00 & \cellcolor[HTML]{FFFFFF} 0.00 & \cellcolor[HTML]{FFFFFF} 0.00 & \cellcolor[HTML]{FFFAF3} 0.04 & \cellcolor[HTML]{FFFEFD} 0.01 \\
\midrule
\multicolumn{8}{l}{\textbf{Hard} (n=50)} \\
\midrule
Gemini 3 Flash & \cellcolor[HTML]{FFFCF9} 0.02 & \cellcolor[HTML]{FFE8BD} 0.22 & \cellcolor[HTML]{FFFFFF} 0.00 & \cellcolor[HTML]{FFFFFF} 0.00 & \cellcolor[HTML]{FFFFFF} 0.00 & \cellcolor[HTML]{FFE0A6} 0.30 & \cellcolor[HTML]{FFF5E4} 0.09 \\
GPT-5 Mini & \cellcolor[HTML]{FFFCF9} 0.02 & \cellcolor[HTML]{E5E3B7} 0.44 & \cellcolor[HTML]{FFFFFF} 0.00 & \cellcolor[HTML]{FFFFFF} 0.00 & \cellcolor[HTML]{D6E5C0} 0.52 & \cellcolor[HTML]{E5E3B7} 0.44 & \cellcolor[HTML]{FFE6B8} 0.24 \\
DeepSeek V3.2 & \cellcolor[HTML]{FFFFFF} 0.00 & \cellcolor[HTML]{ECE2B2} 0.40 & \cellcolor[HTML]{FFFFFF} 0.00 & \cellcolor[HTML]{FFFFFF} 0.00 & \cellcolor[HTML]{FFF4E1} 0.10 & \cellcolor[HTML]{FFF2DB} 0.12 & \cellcolor[HTML]{FFF4E0} 0.10 \\
% qwen3-235b& \cellcolor[HTML]{FFFFFF} 0.00 & \cellcolor[HTML]{FFFAF3} 0.04 & \cellcolor[HTML]{FFFFFF} 0.00 & \cellcolor[HTML]{FFFFFF} 0.00 & \cellcolor[HTML]{FFFFFF} 0.00 & \cellcolor[HTML]{FFFCF9} 0.02 & \cellcolor[HTML]{FFFDFC} 0.01 \\
% GLM-4.7 Flash & \cellcolor[HTML]{FFFFFF} 0.00 & \cellcolor[HTML]{FFFFFF} 0.00 & \cellcolor[HTML]{FFFFFF} 0.00 & \cellcolor[HTML]{FFFFFF} 0.00 & \cellcolor[HTML]{FFFCF9} 0.02 & \cellcolor[HTML]{FFFCF9} 0.02 & \cellcolor[HTML]{FFFEFD} 0.01 \\
% Mercury & \cellcolor[HTML]{FFFFFF} 0.00 & \cellcolor[HTML]{FFFFFF} 0.00 & \cellcolor[HTML]{FFFFFF} 0.00 & \cellcolor[HTML]{FFFFFF} 0.00 & \cellcolor[HTML]{FFFFFF} 0.00 & \cellcolor[HTML]{FFFCF9} 0.02 & \cellcolor[HTML]{FFFEFE} 0.00 \\
% OLMo 3.1 32B & \cellcolor[HTML]{FFFFFF} 0.00 & \cellcolor[HTML]{FFFFFF} 0.00 & \cellcolor[HTML]{FFFFFF} 0.00 & \cellcolor[HTML]{FFFFFF} 0.00 & \cellcolor[HTML]{FFFFFF} 0.00 & \cellcolor[HTML]{FFFAF3} 0.04 & \cellcolor[HTML]{FFFEFD} 0.01 \\
\bottomrule
\end{tabular}}
\vspace{-1.5em}
\end{table}

\subsection{Main Results}
\label{subsec:main-results}

Table~\ref{tab:accuracy_combined_v2} reports verifier-checked accuracy by model, puzzle family, and difficulty. Three models (GPT-5 Mini, Gemini 3 Flash, and DeepSeek V3.2) form a clear frontier tier across the benchmark, with substantially higher easy-tier performance than the rest. Performance degrades sharply from easy to medium/hard across nearly all models and families, and puzzle families are not equally difficult. Not all global constraints pose the same challenge: Galaxies (rotational symmetry) and Loopy (loop closure) remain near zero for every model beyond the easy tier, whereas Bridges (network connectivity) and Pattern (contiguity) retain measurable accuracy into medium and hard. Constraints that require verifying a single global invariant over the entire grid, such as the closed-loop property in Loopy or region symmetry in Galaxies, appear harder than those that decompose into semi-local checks.
The large gap between frontier and non-frontier models on easy tiers, combined with broad hard-tier collapse, indicates a sharp capability boundary rather than smooth scaling. Across tiers, constraint types remain the same and board sizes grow only modestly; what substantially increases is the deduction depth required by the solver (Appendix~\ref{app:dataset_params}), so the performance collapse is not explained by the puzzles being categorically different problems. Our model set also spans diverse architectures. Mercury~\citep{mercury2025}, a diffusion-based language model, and the MoE models LLaMA-4-Maverick~\citep{llama4} and Qwen3-235B all fall well below the frontier tier, suggesting that neither discrete-diffusion decoding nor sparse expert routing gives an obvious advantage on topology-heavy reasoning in the current generation of models. These results align with the view that topology-heavy puzzle solving demands global consistency tracking that remains brittle across architectures. As mentioned above, many easy-tier instances are routine for experienced human solvers~\citep{estermann2024puzzles}.
Taken together, the benchmark shows strong easy-tier separation between frontier and non-frontier models, substantial medium-tier instability, and broad hard-tier failure. This combination supports the benchmark's role as a stress test for global, topology-constrained reasoning.

\subsection{Diagnosing Failure Modes}
\label{sec:diagnosis}

To understand \emph{why} models fail on topology-heavy reasoning, we first perform a systematic error analysis on full chains of thought, then design targeted interventions to test which failure modes genuinely degrade accuracy.
We focus on DeepSeek V3.2, the strongest reasoning model among those in our evaluation that expose their complete chain of thought.

\textbf{Observational method.}
Following LLM-as-a-judge protocols~\citep{g_eval,prometheus}, we use GPT-5-mini to label each chain of thought according to an error taxonomy of eleven categories, of which seven are discussed in the main body (Table~\ref{tab:error_taxonomy_judge}) and the remainder in Appendix~\ref{app:full_error_taxonomy}.
This is a simpler task than puzzle solving: the judge reads a complete reasoning trace and identifies behavioral patterns, rather than constructing a solution from scratch.
We developed categories iteratively, first running a coarse labeling pass to surface prevalent behavioral patterns, then manually refining these into formal definitions.
The judge tags each trace with up to three labels and returns supporting excerpts for manual verification (prompt in Appendix~\ref{app:judge_prompt}).
In total we collect 750 traces across 5 puzzle types and 3 difficulty tiers (50 per combination), excluding Loopy because even frontier models score near zero beyond the easy tier.
Since traces that reach the correct answer may contain recovered mistakes, we restrict our analysis to the 455 incorrect traces to focus on genuine failures. Per-difficulty and per-puzzle breakdowns on the full set are reported in Appendix~\ref{app:per_puzzle_errors}.

\begin{table*}[t]
\centering
\small
\setlength{\tabcolsep}{5pt}
\renewcommand{\arraystretch}{1.1}
\caption{Error taxonomy for chain-of-thought analysis.
The first four categories (above the mid-rule) are plausibly \emph{causal} contributors to failure; the remaining categories are \emph{symptomatic} of an already-degraded reasoning state.
Four additional low-frequency categories (UE, HV, TD, OTH) are reported with the full taxonomy in Appendix~\ref{app:full_error_taxonomy}.}
\label{tab:error_taxonomy_judge}
\resizebox{0.82\textwidth}{!}{%
\begin{tabular}{cl p{0.68\textwidth}}
\toprule
\textbf{Abbr.} & \textbf{Error type} & \textbf{Definition} \\
\midrule
RR & Repeated Reasoning &
Near-identical move sequences repeated from the same board position without meaningful variation. \\
\midrule
STF & State-Tracking Failure &
Claimed board state diverges from the cumulative effect of the model's own stated actions. \\
\midrule
CF & Constraint Forgetting &
An explicit action that directly violates structural puzzle rules (overcounting, crossing, overwriting fixed cells). \\
\midrule
PC & Premature Commitment &
Commits to a provably incorrect configuration and persists for 3+ steps. \\
\midrule
\midrule
ES & Explicit Surrender &
Explicitly gives up, requests a solver, or states it cannot proceed. \\
IO & Incomplete Output &
Fails to produce a complete, valid final answer. \\
RD & Representation Drift &
Internal representation of the problem shifts over time, causing inconsistent reasoning. \\
\bottomrule
\end{tabular}}
\vspace{-0em}
\end{table*}

\begin{figure}[t]
\centering
\includegraphics[width=0.75\textwidth]{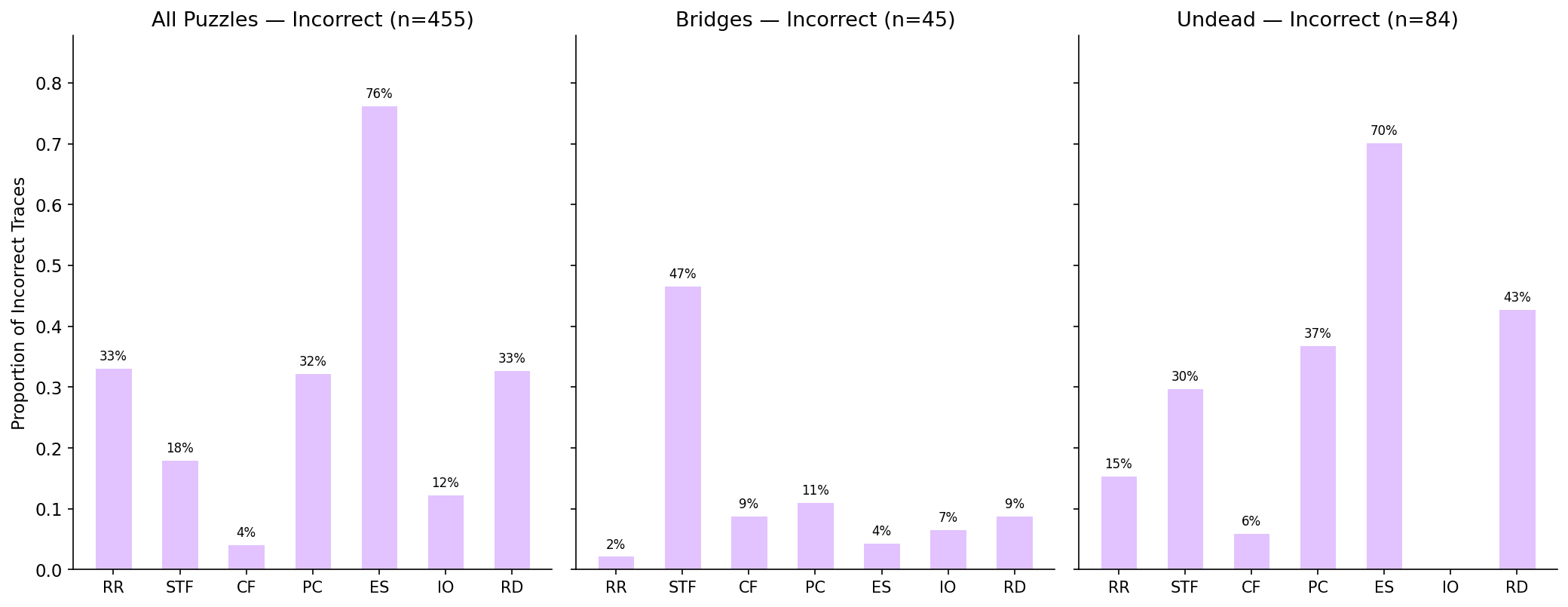}
\caption{Prevalence of the seven main error categories among incorrect traces, pooled across difficulty tiers: all five\protect\footnotemark puzzle types ($n{=}455$, left), Bridges ($n{=}45$, center), and Undead ($n{=}84$, right).
ES dominates the aggregate and Undead panels as a downstream symptom of failure.
STF is the leading category on Bridges (47\%), while PC (37\%) and RD (43\%) are prominent on Undead.
CF is the rarest category across all panels.
Four additional low-frequency categories are reported in Appendix~\ref{app:full_error_taxonomy}.}
\label{fig:error_trio}
\vspace{-1.5em}
\end{figure}

\textbf{Observational results.}
Figure~\ref{fig:error_trio} summarizes the prevalence of each error category among incorrect traces, both aggregated across all five puzzle types (left) and broken down for Bridges (center) and Undead (right).

Explicit surrender (ES) is the most frequent label overall, appearing in 76\% of failed traces.
However, ES reflects a downstream consequence of failure rather than an initiating cause: the model abandons the problem \emph{after} its reasoning has already degraded. Observationally, ES traces use on average 42k output tokens compared to 35k for other failed traces, and cluster near the token ceiling with notably lower variance (standard deviation 8.7k versus 15.4k for other failures). Moreover, no ES trace occurs below 22k tokens, suggesting the model always engages extensively before surrendering (see Appendix~\ref{app:es_tokens}).
Incomplete output (IO, 12\%) plays a similar symptomatic role.
Among the remaining categories, repeated reasoning (RR, 33\%), premature commitment (PC, 32\%), and representation drift (RD, 33\%) occur at comparable rates in the aggregate, while state-tracking failure (STF, 18\%) and constraint forgetting (CF, 4\%) are less common.

The puzzle-specific plots reveal sharper contrasts.
On Bridges, STF dominates failed traces at 47\%, consistent with the rich state that bridge-count and connectivity constraints impose.
On Undead, ES (70\%) and RD (43\%) are the most frequent labels, followed by PC (37\%) and STF (30\%), reflecting the combinatorial difficulty of tracking monster types and directional visibility constraints simultaneously.
CF is the least frequent category across all puzzles, but as the intervention experiments below show, its low frequency hides a disproportionately large causal effect.

\footnotetext{Loopy omitted because even frontier models score zero on medium and hard.}

\begin{wrapfigure}{R}{0.5\textwidth}
\centering
\vspace{-1.2em}
\includegraphics[width=0.48\textwidth]{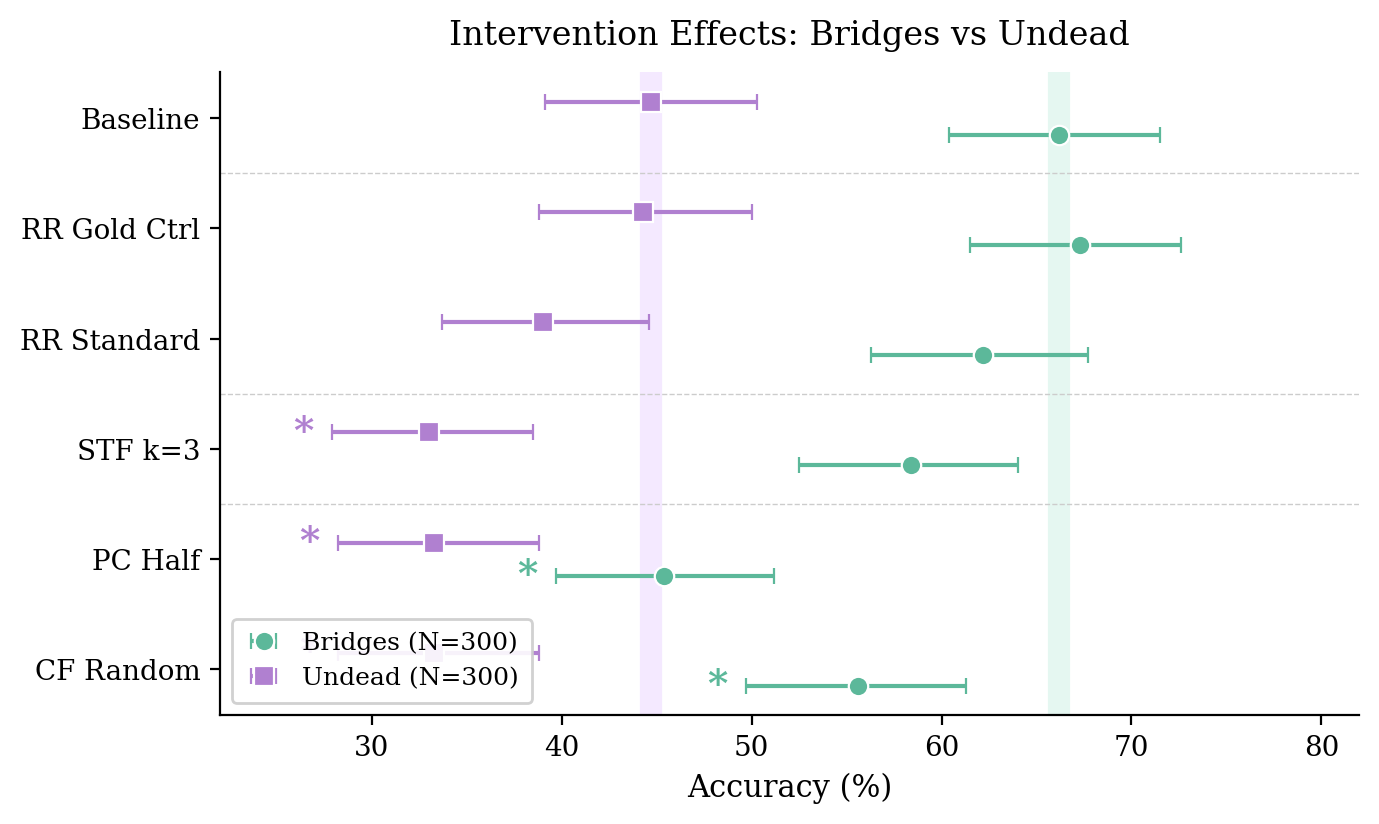}
\caption{Intervention effects on Bridges (circles, $N{=}300$) and Undead (squares, $N{=}300$).
Points show total accuracy with 95\% Wilson CIs; shaded bands mark the baseline.
PC and CF produce large, significant drops on both puzzles; STF reaches significance only on Undead; RR variants are indistinguishable from baseline.}
\label{fig:intervention_forest}
\vspace{-1em}
\end{wrapfigure}

\textbf{Intervention setup.}
To disentangle frequency from impact, we select four error types with distinct mechanistic profiles and test each via targeted injection. Premature commitment and repeated reasoning both waste the token budget, the former by pursuing dead-end branches and the latter by cycling through near-identical analyses that prevent forward progress.
Meanwhile, state-tracking failure and constraint forgetting both corrupt the model's representation of the puzzle, with state-tracking failure causing the model to lose track of the board state entirely, while constraint forgetting introduces rule violations that the model cannot detect, leaving it to build on a flawed foundation.
To test whether these mechanisms genuinely drive failure and whether observational frequency predicts causal impact, we inject each error type into partial gold solution paths and measure downstream accuracy.
We evaluate DeepSeek V3.2 on Bridges across three difficulty tiers (Easy/5$\times$5, Medium/7$\times$7, Hard/10$\times$10) with 100 puzzles per difficulty tier (300 per condition, 2,100 total API calls across 7 conditions), and replicate on Undead ($N{=}300$).
Each puzzle is presented with an unmodified partial gold path covering ${\sim}$15\% of the solution and intervention conditions apply a targeted modification before asking the model to continue.
We verify in Appendix~\ref{app:baseline_calibration} that the unmodified baseline is competitive with DeepSeek V3.2's standard evaluation accuracy, confirming that the gold prefix does not introduce a meaningful distribution shift.
All accuracy figures use 95\% Wilson score confidence intervals.

To generate intervention material, we expand the solver's solution paths into partial game trees by sampling legal-but-wrong moves at branching points and following wrong branches until a constraint violation terminates them.
Candidate prefixes that match any future gold state are discarded to prevent information leakage.
The four intervention types are:\textbf{RR}, $k{=}2$ wrong-path-then-backtrack cycles before the continuation point; \textbf{RR Gold Ctrl}, a length-matched control that replaces the wrong-path segments with repetitions from the partial solution path, isolating whether increased context length alone affects performance; \textbf{STF}, 1 bridge cell flipped in each of the last $k{=}3$ intermediate grids while the action log is left intact; \textbf{CF}, one constraint violation (overcounting, crossing, or count modification) injected at the last step with no indication of an error; and \textbf{PC}, the gold prefix replaced with a wrong-path branch diverging at the first move, truncated before any obvious violation.
Full definitions and parameters are in Appendix~\ref{app:intervention_details}.

\textbf{Intervention results.}
Figure~\ref{fig:intervention_forest} summarizes accuracy across all conditions for Bridges and Undead.
PC and CF produce the largest accuracy drops, with PC falling by 20.8\,pp on Bridges and 11.3\,pp on Undead, and CF by 10.6\,pp and 11.3\,pp respectively. In both cases, confidence intervals exclude the baseline.
STF has a weaker and less consistent effect, producing a borderline 7.8\,pp drop on Bridges but a significant 11.7\,pp drop on Undead, where the model must track richer state.
RR shows no effect on either puzzle. Length-matched gold controls confirm that added context alone does not degrade performance, so repeated reasoning in incorrect traces is a symptom of search, not a cause of failure.

\textbf{Observational frequency is a poor proxy for causal impact.}
CF is among the rarest errors in the observational analysis yet produces one of the largest causal effects when injected.
The contrast with STF helps explain why. STF introduces a \emph{syntactic} inconsistency between the grid and the action log, and the model can partially recover by cross-referencing the uncorrupted text.
CF, by contrast, produces an internally consistent state that violates puzzle rules, a \emph{semantic} error detectable only through active constraint verification.
The model largely lacks this capability, which explains why even rare constraint violations are so damaging.
Per-difficulty breakdowns are in Appendix~\ref{app:intervention_details}.

\subsection{Positive Interventions}
\label{sec:positive_interventions}
\label{sec:mitigations}

Having identified premature commitment, constraint forgetting, and state-tracking failure as causal bottlenecks (Section~\ref{sec:diagnosis}), we test whether targeted interventions can recover performance. We evaluate input-format changes, tool-augmented constraint access, and prompt-level strategy guidance, finding that changing the interface the model reasons over is more effective than changing how it approaches the problem.
% --- Input Format Intervention (v2) ---

\subsubsection{Input Format Intervention}
\label{subsec:format-intervention}

\begin{figure}[]
    \centering
    \includegraphics[height=2.2cm]{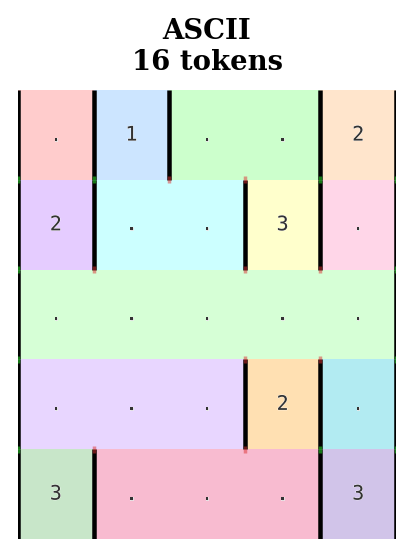}
    \hspace{0.3cm}
    \includegraphics[height=2.2cm]{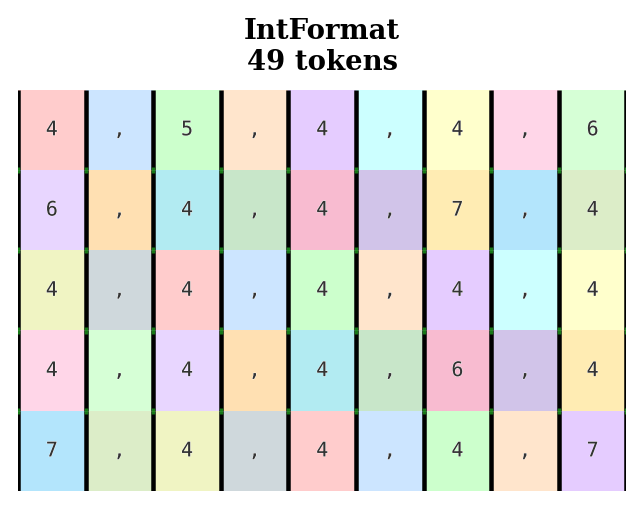}
    \hspace{0.3cm}
    \includegraphics[height=2.2cm]{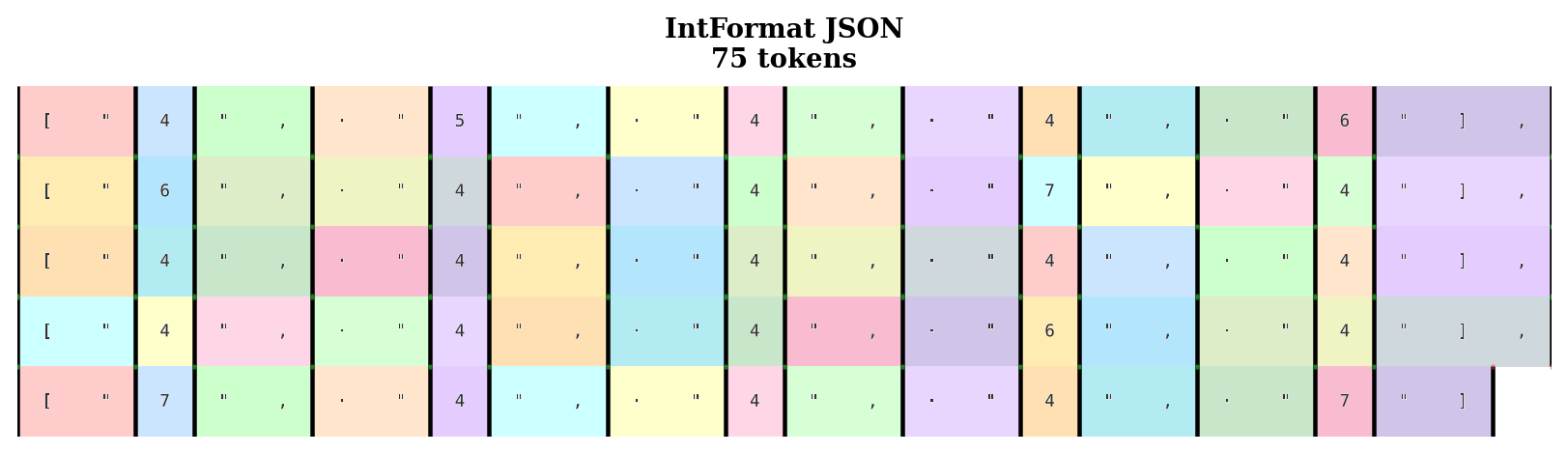}
    \caption{Tokenization of a Bridges puzzle with GPT-5-mini's tokenizer~\citep{tiktoken}. ASCII (left) produces ragged boundaries that straddle grid cells; IntFormat (center) and IntFormat-JSON (right) yield uniform, cell-aligned tokens preserving board structure.}
    \label{fig:rugged_boundaries}
\end{figure}

The two-dimensional structure of puzzles tokenizes poorly under standard BPE tokenizers. Many tokens straddle cell boundaries, merging parts of adjacent cells so that rows occupying identical numbers of grid cells map to different numbers of tokens (Figure~\ref{fig:rugged_boundaries}). Because models process input as a flat token sequence, they rely on consistent alignment across rows to reconstruct the 2D grid. Ragged tokenization disrupts this, making it difficult to identify cell boundaries, align coordinates, or extract constraints that depend on spatial adjacency. This suggests that part of the observed accuracy gap may arise from input parsing rather than reasoning limitations. We therefore hypothesize that tokenization-induced structure loss accounts for a substantial share of model failures, and re-evaluate \emph{identical} puzzle instances under four alternative input encodings.
\noindent

\begin{description}\setlength{\itemsep}{0pt}\setlength{\parskip}{0pt}\setlength{\parsep}{0pt}
  \item[\textbf{ASCII}] Plain-text grid resulting in irregular cell boundaries from tokenization.
  \item[\textbf{IntFormat}] Comma-separated integer encoding leading to more uniform, cell-aligned tokenization.
  \item[\textbf{IntFormat-JSON}] IntFormat as 2D JSON list, similar to ARC-AGI's format~\citep{chollet2025arcagi2}.
  \item[\textbf{ASCII + Image}] ASCII grid + programmatically rendered puzzle image.
\end{description}

% \textbf{ASCII:} Plain-text grid with standard tokenization, resulting in irregular cell boundaries. \\
% \textbf{IntFormat:} Cells encoded as numeric values separated by commas, producing more uniform, cell-aligned tokenization. \\
% \textbf{IntFormat-JSON:} IntFormat wrapped in a two-dimensional JSON list, providing explicit structural boundaries (similar to ARC-AGI~\citep{chollet2025arcagi2}). \\
% \textbf{ASCII+image:} ASCII grid accompanied by a programmatically rendered image of the puzzle.

IntFormat and IntFormat-JSON require more tokens overall but preserve consistent cell-level alignment in the token sequence, maintaining the board’s spatial structure during processing.

% \textbf{Results.}
% Table~\ref{tab:format_comparison_per_model} reports per-model accuracy changes relative to each model's own ASCII baseline.
% Cell-aligned formats produce large gains on Bridges ($+$12 to $+$38.7\,pp across models) and Galaxies ($+$15.3 to $+$32\,pp), where integer encodings consistently outperform ASCII.
% FlowFree sees moderate improvements ($+$3 to $+$10\,pp).
% For Loopy, IntFormat alone barely helps, but IntFormat-JSON yields meaningful gains (up to $+$16\,pp for GPT-5-mini), suggesting that JSON structure provides additional parsing benefit beyond cell alignment.

% Undead and Pattern are exceptions: all three models regress on Undead, though the magnitude varies widely (DeepSeek $-$26\,pp vs.\ GPT-5-mini $-$8\,pp with IntFormat).
% On Pattern, GPT-5-mini drops $-$21.8\,pp while Gemini gains $+$3.3\,pp under the same format change.
% The direction and magnitude of these regressions vary substantially across models, indicating that format effects interact with model-specific factors such as tokenizer design and pretraining distribution.
% Adding an image alongside ASCII does not consistently help and degrades accuracy on Pattern ($-$17.8\,pp for GPT-5-mini) and Undead ($-$5.2\,pp), suggesting that current vision-language integration does not aid structured constraint reasoning.

\textbf{Results.}
Table~\ref{tab:format_comparison_per_model_main} reports per-model accuracy changes relative to each model’s ASCII baseline. Cell-aligned integer encodings yield large gains on several puzzle families, particularly Bridges and Galaxies, with improvements of roughly $+30$ to $+40$\,pp. Flow Free shows smaller but consistent improvements. 
% For Loopy, simple cell alignment offers little benefit, but adding JSON structure produces meaningful gains, suggesting that explicit structural fields provide advantages beyond grid alignment. 
However, the effects are not uniform. On Undead and Pattern, integer encodings often degrade performance, in some cases substantially (e.g., $-26$\,pp on Undead for DeepSeek). The direction and magnitude of these changes vary across models, indicating that format interventions interact with tokenizer design and pretraining distribution.

Adding an image alongside ASCII input does not reliably help and can reduce accuracy, particularly on Pattern and Undead. This suggests that current multimodal encoders do not improve structured constraint extraction and may interfere with algebraic reasoning over grid-based problems.

\begin{table}[]
\centering
\caption{Per-model accuracy (\%) by input format, average of all difficulties. Each model's $\Delta$ rows show change relative to its own Plain ASCII baseline.$^\ddagger$ In each section the biggest delta is marked (positive in green and negative in red). Full table in the Appendix in Figure \ref{tab:format_comparison_per_model}.}
\label{tab:format_comparison_per_model_main}
\resizebox{0.85\textwidth}{!}{%
\begin{tabular}{lrrrrrr}
\toprule
Model / Format & FlowFree & Bridges & Loopy & Galaxies & Undead & Pattern \\
\midrule
% \textbf{Qwen3-235B Thinking} & 22.0 & 36.7 & 0.0 & 0.0 & 8.7 & 22.0 \\
% \quad $\Delta$ Int-Format & \cellcolor[HTML]{D5F5E3} +9.3 & +20.7 & \cellcolor[HTML]{D5F5E3} +0.0 & \cellcolor[HTML]{D5F5E3} +8.0 & \cellcolor[HTML]{FADBD8} -8.7 & -13.3 \\
% \quad $\Delta$ Int-Format JSON & +7.3 & \cellcolor[HTML]{D5F5E3} +22.0 & \cellcolor[HTML]{D5F5E3} +0.0 & +4.0 & -8.0 & \cellcolor[HTML]{D5F5E3} +2.0 \\
% \midrule
\textbf{DeepSeek V3.2} & 28.7 & 74.0 & 8.7 & 1.3 & 44.0 & 52.7 \\
\quad $\Delta$ Int-Format & \cellcolor[HTML]{D5F5E3} +5.3 & \cellcolor[HTML]{D5F5E3} +17.3 & \cellcolor[HTML]{D5F5E3} +7.3 & \cellcolor[HTML]{D5F5E3} +28.0 & -26.0 & \cellcolor[HTML]{FADBD8} -8.7 \\
\quad $\Delta$ Int-Format JSON & +3.3 & +14.0 & +3.3 & +23.3 & \cellcolor[HTML]{FADBD8} -36.7 & -8.0 \\
\midrule
\textbf{GPT-5 Mini} & 30.0 & 68.7 & 10.7 & 19.3 & 73.3 & 75.3 \\
\quad $\Delta$ With Image & +0.7 & -0.7 & +2.0 & -0.7 & -5.3 & -18.0 \\
\quad $\Delta$ Int-Format & \cellcolor[HTML]{D5F5E3} +8.0 & +12.0 & +12.0 & +15.3 & \cellcolor[HTML]{FADBD8} -8.7 & \cellcolor[HTML]{FADBD8} -22.0 \\
\quad $\Delta$ Int-Format JSON & +4.7 & \cellcolor[HTML]{D5F5E3} +15.3 & \cellcolor[HTML]{D5F5E3} +13.3 & \cellcolor[HTML]{D5F5E3} +17.3 & -7.3 & -12.7 \\
\midrule
\textbf{Gemini 3 Flash} & 42.7 & 58.0 & 6.0 & 5.3 & 29.3 & 66.0 \\
\quad $\Delta$ With Image & +4.7 & +7.3 & -3.3 & +0.0 & -4.7 & \cellcolor[HTML]{D5F5E3} +6.7 \\
\quad $\Delta$ Int-Format & +8.0 & +37.3 & +7.3 & \cellcolor[HTML]{D5F5E3} +32.0 & -6.0 & +3.3 \\
\quad $\Delta$ Int-Format JSON & \cellcolor[HTML]{D5F5E3} +10.0 & \cellcolor[HTML]{D5F5E3} +38.7 & \cellcolor[HTML]{D5F5E3} +12.0 & +29.3 & \cellcolor[HTML]{FADBD8} -25.3 & +4.7 \\
\bottomrule
\multicolumn{7}{l}{\footnotesize $^\ddagger$DeepSeek V3.2 is text-only; image condition not evaluated.}
\end{tabular}}
\vspace{-1.5em}
\end{table}

The format results suggest that spatial parsing is a first-order bottleneck; we next test this more directly by externalizing constraint extraction entirely.

% --- State Externalization Intervention (v4: compact, grid-removal-led) ---

\subsubsection{Tool-Augmented Reasoning}
\label{subsec:state-externalization}

The format intervention leaves open whether the bottleneck lies in parsing the initial input or in extracting constraints during multi-step reasoning, since changing the input format affects both simultaneously.
To disentangle these, we keep the initial puzzle in ASCII but give the model an iterative tool-call interface: an external engine maintains the authoritative board state and provides the current state information on demand.
We evaluate DeepSeek-v3.2 on hard Bridges ($10{\times}10$, $N{=}50$), pairing the strongest open-CoT reasoning model with the tier where baseline accuracy (40\%) is high enough to measure gains yet low enough to leave room for improvement.
% The model accesses five tools in two categories.
% \emph{State-mutation tools} (\texttt{make\_move}, \texttt{render\_board}) apply moves and return the board as an ASCII grid.
% \emph{Constraint-query tools} (\texttt{state\_summary}, \texttt{neighbors}, \texttt{components}) return pre-computed structured data---remaining bridge counts, legal moves, connected components---as JSON (specifications in Appendix~\ref{app:tool_augmented}).
% The key design distinction is that state-mutation tools return spatial output, while constraint-query tools return structured numeric fields; no tool solves the puzzle or suggests moves.

The model has access to five tools.
\texttt{make\_move} applies a proposed move to an authoritative external board state.
The current state can be observed in two forms: as a spatial ASCII grid via \texttt{render\_board}, or as structured JSON via \texttt{state\_summary}, \texttt{neighbors}, and \texttt{components}, which report pre-computed state/constraint information such as remaining bridge counts, legal moves, and connected components (specifications in Appendix~\ref{app:tool_augmented}).
These tools expose state but do not solve the puzzle or suggest moves.

\begin{wraptable}{r}{0.42\textwidth}
\vspace{-1.2em}
\centering
\small
\caption{Tool ablation, hard Bridges ($N{=}50$). Full table in Appendix~\ref{app:tool_augmented}.}
\label{tab:tool_ablation_main}
\begin{tabular}{lcc}
\toprule
Condition & Acc.\ & Valid \\
\midrule
Baseline (no tools) & 40\% & 50\% \\
Structured only & 46\% & 100\% \\
\;\; + \texttt{render\_board} & 42\% & 100\% \\
Full structured suite & 50\% & 100\% \\
\bottomrule
\end{tabular}
\vspace{-1em}
\end{wraptable}

Table~\ref{tab:tool_ablation_main} summarizes the key comparisons.
With only structured tools (\texttt{make\_move} and \texttt{state\_summary}, no spatial grid), accuracy rises to 46\%, a 6\,pp gain over the no-tool baseline, while board-validity errors are eliminated entirely.
The full constraint-query suite reaches 50\%.
To isolate whether the spatial grid contributes, we toggle \texttt{render\_board} while holding everything else constant: adding it to the two-tool configuration \emph{drops} accuracy by 4\,pp (46\%\,$\to$\,42\%), and with the full suite it has no effect (50\% either way).
When \texttt{render\_board} is available, the model calls it eight times per puzzle on average regardless of instructions to the contrary, and these spatial renderings appear to interfere with the algebraic reasoning that \texttt{state\_summary} enables.
The spatial grid is at best redundant and at worst harmful.

% Inspection of baseline (no-tool) reasoning traces sheds light on why structured tools help.
% The model's natural strategy is to convert the ASCII grid into an algebraic constraint system at the start of its chain of thought:- labeling islands, enumerating possible connections, and setting up explicit degree equations, then reason entirely over that structured representation without re-reading the grid.
% Structured tools provide a reliable, updatable version of exactly this algebraic representation; the error-prone manual conversion is bypassed.

Inspection of baseline (no-tool) reasoning traces suggests a mechanism for why structured tools help. The model typically starts by translating the ASCII grid into an algebraic constraint system by labeling islands, enumerating candidate connections, and writing degree equations. It then carries out multi-step reasoning over this structured representation without re-printing the grid. We hypothesize that structured tools help because they provide a reliable and updatable version of this algebraic form, which bypasses the error-prone manual conversion step.

Taken together with the format intervention, these results converge on a consistent finding for Bridges: \textbf{the primary bottleneck is extracting constraints from spatial representations, not reasoning over them.}
The model can solve hard Bridges at 40\% even without tools, demonstrating that the underlying reasoning capability exists.
What it cannot do reliably is compile spatial layouts into the algebraic form it needs for constraint propagation, which is a step that better input formats ease and structured tools eliminate.

\textbf{Prompt-level interventions.}
We also tested whether prompt-based interventions, like few-shot worked examples, explicit planning instructions, and demonstrations of mistake recovery, can address premature commitment on Bridges puzzles.
Across 11 conditions evaluated on DeepSeek V3.2, no prompt intervention significantly outperformed baseline on hard problems. Most interventions degraded accuracy, with longer prompts correlating with worse hard-tier performance.
The model's extended-thinking process appears to dominate over prompt-level strategy guidance.
Full results and analysis are reported in Appendix~\ref{app:prompt_interventions}.

% \section{Conclusion}
% \label{sec:conclusion}

% TopoBench shows that topology-heavy spatial reasoning remains a genuine open problem: even the strongest models we evaluate solve fewer than a quarter of hard instances, and two puzzle families (Galaxies, Loopy) are essentially unsolved at all difficulties.
% Our diagnostic pipeline reveals that the dominant bottlenecks are premature commitment and constraint forgetting, while more visually salient patterns like repeated reasoning turn out to be symptomatic rather than causal; correlational error frequency is a poor guide to what actually breaks performance.
% On the mitigation side, cell-aligned representations and tool-augmented constraint queries each address a specific diagnosed failure mode, whereas prompt-level strategy guidance has no effect on reasoning models.
% Our causal analysis is limited to one model (DeepSeek-v3.2) and two puzzle families (Bridges, Undead), and all mitigations operate at inference time with pass@1 evaluation; extending the intervention framework to more models and puzzles, and testing whether majority voting or best-of-N sampling changes the key findings, are natural next steps.
% The finding that constraint queryability drives accuracy gains more than state externalization raises a broader question: whether models can learn to internalize constraint verification through training (e.g., reinforcement learning with process-level rewards that penalize constraint violations), or whether tool-augmented reasoning is the more scalable path for tasks that demand sustained global-state maintenance.

\section{Conclusion}
\label{sec:conclusion}

TopoBench shows that topology-heavy spatial reasoning remains challenging for state-of-the-art models, with performance staying below 25\% on hard instances.
Using a diagnostic pipeline that combines trace-level error annotation with causal interventions on gold solution prefixes, we find that error frequency is not a reliable proxy for causal importance.
Instead, two failure modes dominate: premature commitment and constraint forgetting, both of which yield large accuracy drops when injected. On the mitigation side, input representations and tool augmentation can improve performance, but gains are uneven across puzzle families.
For tool use, ablations show that improvements come from structured constraint information rather than from repeated spatial renderings.
Overall, the evidence from tool-augmented experiments on Bridges points to a consistent bottleneck: models struggle more with extracting and maintaining constraints from spatial representations than with reasoning once those constraints are available.

\textbf{Limitations.} Our causal analysis covers one model and two puzzle families, tool augmentation is evaluated on a single model and puzzle family and all mitigations are inference-time with pass@1. Our error taxonomy is derived from chain-of-thought traces, which may not always faithfully reflect the model's internal computation~\citep{turpin2023language,lanham2023measuring}; errors visible in the trace may not be the true cause of failure.
Future work should extend interventions and tools to more settings, and test whether best-of-$N$ or voting changes the conclusions.
More broadly, our results raise the question of whether models can be trained to internalize constraint verification (e.g., via process-level rewards for violations of global invariants) or whether tool-augmented reasoning is a more scalable approach for tasks requiring sustained maintenance of spatial constraints.

%    used only for main submission


% TO REMOVE
% \nocite{*}

\bibliography{cleaned_references}
\bibliographystyle{iclr2026_conference}

\clearpage
\appendix
\section{Appendix}

\section{Accuracy with Confidence Intervals and Format Comparison}

% end benchmark predictability table

\begin{table}[htbp]
\centering
\small
\caption{Accuracy by model and puzzle type across difficulties. Rows with all-zero accuracy are omitted.}
\label{tab:accuracy_combined}
\resizebox{\textwidth}{!}{%
\begin{tabular}{lrrrrrrr}
\toprule
Model & FlowFree & Bridges & Loopy & Galaxies & Undead & Pattern & Avg \\
\midrule
\multicolumn{8}{l}{\textbf{Easy}} \\
\midrule
Gemini 3 Flash & \cellcolor[HTML]{AFE7D2} 0.72\,{\scriptsize $\pm$0.12} & \cellcolor[HTML]{89D5B4} 0.86\,{\scriptsize $\pm$0.10} & \cellcolor[HTML]{FFECC9} 0.18\,{\scriptsize $\pm$0.11} & \cellcolor[HTML]{FFEECF} 0.16\,{\scriptsize $\pm$0.10} & \cellcolor[HTML]{9FDFC5} 0.78\,{\scriptsize $\pm$0.11} & \cellcolor[HTML]{7FD0AB} 0.90\,{\scriptsize $\pm$0.09} & \cellcolor[HTML]{C7E7CA} 0.60\,{\scriptsize $\pm$0.06} \\
GPT-5 Mini & \cellcolor[HTML]{BCE9D2} 0.66\,{\scriptsize $\pm$0.13} & \cellcolor[HTML]{7FD0AB} 0.90\,{\scriptsize $\pm$0.09} & \cellcolor[HTML]{FBE0A8} 0.32\,{\scriptsize $\pm$0.13} & \cellcolor[HTML]{CBE7C8} 0.58\,{\scriptsize $\pm$0.13} & \cellcolor[HTML]{84D3B0} 0.88\,{\scriptsize $\pm$0.09} & \cellcolor[HTML]{79CEA7} 0.92\,{\scriptsize $\pm$0.08} & \cellcolor[HTML]{B2E8D4} 0.71\,{\scriptsize $\pm$0.05} \\
DeepSeek V3.2 & \cellcolor[HTML]{C0E8CF} 0.64\,{\scriptsize $\pm$0.13} & \cellcolor[HTML]{74CBA3} 0.94\,{\scriptsize $\pm$0.07} & \cellcolor[HTML]{FFE4B1} 0.26\,{\scriptsize $\pm$0.12} & \cellcolor[HTML]{FFFAF3} 0.04\,{\scriptsize $\pm$0.06} & \cellcolor[HTML]{A4E2CA} 0.76\,{\scriptsize $\pm$0.12} & \cellcolor[HTML]{94DABD} 0.82\,{\scriptsize $\pm$0.11} & \cellcolor[HTML]{CBE6C7} 0.58\,{\scriptsize $\pm$0.06} \\
qwen3-235b-a22b-thinking-2507 & \cellcolor[HTML]{DDE4BC} 0.48\,{\scriptsize $\pm$0.13} & \cellcolor[HTML]{AFE7D2} 0.72\,{\scriptsize $\pm$0.12} & \cellcolor[HTML]{FFFFFF} 0.00\,{\scriptsize $\pm$0.04} & \cellcolor[HTML]{FFFFFF} 0.00\,{\scriptsize $\pm$0.04} & \cellcolor[HTML]{FFE6B7} 0.24\,{\scriptsize $\pm$0.12} & \cellcolor[HTML]{E5E3B7} 0.44\,{\scriptsize $\pm$0.13} & \cellcolor[HTML]{FCE0A7} 0.31\,{\scriptsize $\pm$0.05} \\
GLM-4.7 Flash & \cellcolor[HTML]{FFF0D5} 0.14\,{\scriptsize $\pm$0.10} & \cellcolor[HTML]{FFFFFF} 0.00\,{\scriptsize $\pm$0.04} & \cellcolor[HTML]{FFFAF3} 0.04\,{\scriptsize $\pm$0.06} & \cellcolor[HTML]{FFFFFF} 0.00\,{\scriptsize $\pm$0.04} & \cellcolor[HTML]{FFFFFF} 0.00\,{\scriptsize $\pm$0.04} & \cellcolor[HTML]{FFFCF9} 0.02\,{\scriptsize $\pm$0.05} & \cellcolor[HTML]{FFFBF5} 0.03\,{\scriptsize $\pm$0.02} \\
Mercury & \cellcolor[HTML]{FFF4E1} 0.10\,{\scriptsize $\pm$0.09} & \cellcolor[HTML]{FFEAC3} 0.20\,{\scriptsize $\pm$0.11} & \cellcolor[HTML]{FFFFFF} 0.00\,{\scriptsize $\pm$0.04} & \cellcolor[HTML]{FFFFFF} 0.00\,{\scriptsize $\pm$0.04} & \cellcolor[HTML]{FFFCF9} 0.02\,{\scriptsize $\pm$0.05} & \cellcolor[HTML]{FFF6E7} 0.08\,{\scriptsize $\pm$0.08} & \cellcolor[HTML]{FFF8EB} 0.07\,{\scriptsize $\pm$0.03} \\
OLMo 3.1 32B Think & \cellcolor[HTML]{FFE6B7} 0.24\,{\scriptsize $\pm$0.12} & \cellcolor[HTML]{FFF6E7} 0.08\,{\scriptsize $\pm$0.08} & \cellcolor[HTML]{FFFCF9} 0.02\,{\scriptsize $\pm$0.05} & \cellcolor[HTML]{FFFFFF} 0.00\,{\scriptsize $\pm$0.04} & \cellcolor[HTML]{FFFFFF} 0.00\,{\scriptsize $\pm$0.04} & \cellcolor[HTML]{FFF6E7} 0.08\,{\scriptsize $\pm$0.08} & \cellcolor[HTML]{FFF7EA} 0.07\,{\scriptsize $\pm$0.03} \\
Qwen3-32B & \cellcolor[HTML]{FFECC9} 0.18\,{\scriptsize $\pm$0.11} & \cellcolor[HTML]{FFF8ED} 0.06\,{\scriptsize $\pm$0.07} & \cellcolor[HTML]{FFFFFF} 0.00\,{\scriptsize $\pm$0.04} & \cellcolor[HTML]{FFFFFF} 0.00\,{\scriptsize $\pm$0.04} & \cellcolor[HTML]{FFFCF9} 0.02\,{\scriptsize $\pm$0.05} & \cellcolor[HTML]{FFEECF} 0.16\,{\scriptsize $\pm$0.10} & \cellcolor[HTML]{FFF7EA} 0.07\,{\scriptsize $\pm$0.03} \\
\midrule
\multicolumn{8}{l}{\textbf{Medium}} \\
\midrule
Gemini 3 Flash & \cellcolor[HTML]{D2E6C3} 0.54\,{\scriptsize $\pm$0.13} & \cellcolor[HTML]{BCE9D2} 0.66\,{\scriptsize $\pm$0.13} & \cellcolor[HTML]{FFFFFF} 0.00\,{\scriptsize $\pm$0.04} & \cellcolor[HTML]{FFFFFF} 0.00\,{\scriptsize $\pm$0.04} & \cellcolor[HTML]{FFF4E1} 0.10\,{\scriptsize $\pm$0.09} & \cellcolor[HTML]{9FDFC5} 0.78\,{\scriptsize $\pm$0.11} & \cellcolor[HTML]{F6E1AB} 0.35\,{\scriptsize $\pm$0.05} \\
GPT-5 Mini & \cellcolor[HTML]{FFE8BD} 0.22\,{\scriptsize $\pm$0.11} & \cellcolor[HTML]{AFE7D2} 0.72\,{\scriptsize $\pm$0.12} & \cellcolor[HTML]{FFFFFF} 0.00\,{\scriptsize $\pm$0.04} & \cellcolor[HTML]{FFFFFF} 0.00\,{\scriptsize $\pm$0.04} & \cellcolor[HTML]{99DDC1} 0.80\,{\scriptsize $\pm$0.11} & \cellcolor[HTML]{7FD0AB} 0.90\,{\scriptsize $\pm$0.09} & \cellcolor[HTML]{E5E3B7} 0.44\,{\scriptsize $\pm$0.06} \\
DeepSeek V3.2 & \cellcolor[HTML]{FFE8BD} 0.22\,{\scriptsize $\pm$0.11} & \cellcolor[HTML]{84D3B0} 0.88\,{\scriptsize $\pm$0.09} & \cellcolor[HTML]{FFFFFF} 0.00\,{\scriptsize $\pm$0.04} & \cellcolor[HTML]{FFFFFF} 0.00\,{\scriptsize $\pm$0.04} & \cellcolor[HTML]{E1E4B9} 0.46\,{\scriptsize $\pm$0.13} & \cellcolor[HTML]{C0E8CF} 0.64\,{\scriptsize $\pm$0.13} & \cellcolor[HTML]{F2E1AE} 0.37\,{\scriptsize $\pm$0.05} \\
qwen3-235b-a22b-thinking-2507 & \cellcolor[HTML]{FFECC9} 0.18\,{\scriptsize $\pm$0.11} & \cellcolor[HTML]{F7E1AA} 0.34\,{\scriptsize $\pm$0.13} & \cellcolor[HTML]{FFFFFF} 0.00\,{\scriptsize $\pm$0.04} & \cellcolor[HTML]{FFFFFF} 0.00\,{\scriptsize $\pm$0.04} & \cellcolor[HTML]{FFFCF9} 0.02\,{\scriptsize $\pm$0.05} & \cellcolor[HTML]{FFEAC3} 0.20\,{\scriptsize $\pm$0.11} & \cellcolor[HTML]{FFF2DA} 0.12\,{\scriptsize $\pm$0.04} \\
GLM-4.7 Flash & \cellcolor[HTML]{FFFFFF} 0.00\,{\scriptsize $\pm$0.04} & \cellcolor[HTML]{FFFFFF} 0.00\,{\scriptsize $\pm$0.04} & \cellcolor[HTML]{FFFCF9} 0.02\,{\scriptsize $\pm$0.05} & \cellcolor[HTML]{FFFFFF} 0.00\,{\scriptsize $\pm$0.04} & \cellcolor[HTML]{FFFFFF} 0.00\,{\scriptsize $\pm$0.04} & \cellcolor[HTML]{FFFFFF} 0.00\,{\scriptsize $\pm$0.04} & \cellcolor[HTML]{FFFEFE} 0.00\,{\scriptsize $\pm$0.01} \\
OLMo 3.1 32B Think & \cellcolor[HTML]{FFFFFF} 0.00\,{\scriptsize $\pm$0.04} & \cellcolor[HTML]{FFFFFF} 0.00\,{\scriptsize $\pm$0.04} & \cellcolor[HTML]{FFFFFF} 0.00\,{\scriptsize $\pm$0.04} & \cellcolor[HTML]{FFFFFF} 0.00\,{\scriptsize $\pm$0.04} & \cellcolor[HTML]{FFFFFF} 0.00\,{\scriptsize $\pm$0.04} & \cellcolor[HTML]{FFFAF3} 0.04\,{\scriptsize $\pm$0.06} & \cellcolor[HTML]{FFFEFD} 0.01\,{\scriptsize $\pm$0.01} \\
\midrule
\multicolumn{8}{l}{\textbf{Hard}} \\
\midrule
Gemini 3 Flash & \cellcolor[HTML]{FFFCF9} 0.02\,{\scriptsize $\pm$0.05} & \cellcolor[HTML]{FFE8BD} 0.22\,{\scriptsize $\pm$0.11} & \cellcolor[HTML]{FFFFFF} 0.00\,{\scriptsize $\pm$0.04} & \cellcolor[HTML]{FFFFFF} 0.00\,{\scriptsize $\pm$0.04} & \cellcolor[HTML]{FFFFFF} 0.00\,{\scriptsize $\pm$0.04} & \cellcolor[HTML]{FFE0A6} 0.30\,{\scriptsize $\pm$0.12} & \cellcolor[HTML]{FFF5E4} 0.09\,{\scriptsize $\pm$0.03} \\
GPT-5 Mini & \cellcolor[HTML]{FFFCF9} 0.02\,{\scriptsize $\pm$0.05} & \cellcolor[HTML]{E5E3B7} 0.44\,{\scriptsize $\pm$0.13} & \cellcolor[HTML]{FFFFFF} 0.00\,{\scriptsize $\pm$0.04} & \cellcolor[HTML]{FFFFFF} 0.00\,{\scriptsize $\pm$0.04} & \cellcolor[HTML]{D6E5C0} 0.52\,{\scriptsize $\pm$0.13} & \cellcolor[HTML]{E5E3B7} 0.44\,{\scriptsize $\pm$0.13} & \cellcolor[HTML]{FFE6B8} 0.24\,{\scriptsize $\pm$0.05} \\
DeepSeek V3.2 & \cellcolor[HTML]{FFFFFF} 0.00\,{\scriptsize $\pm$0.04} & \cellcolor[HTML]{ECE2B2} 0.40\,{\scriptsize $\pm$0.13} & \cellcolor[HTML]{FFFFFF} 0.00\,{\scriptsize $\pm$0.04} & \cellcolor[HTML]{FFFFFF} 0.00\,{\scriptsize $\pm$0.04} & \cellcolor[HTML]{FFF4E1} 0.10\,{\scriptsize $\pm$0.09} & \cellcolor[HTML]{FFF2DB} 0.12\,{\scriptsize $\pm$0.09} & \cellcolor[HTML]{FFF4E0} 0.10\,{\scriptsize $\pm$0.03} \\
qwen3-235b-a22b-thinking-2507 & \cellcolor[HTML]{FFFFFF} 0.00\,{\scriptsize $\pm$0.04} & \cellcolor[HTML]{FFFAF3} 0.04\,{\scriptsize $\pm$0.06} & \cellcolor[HTML]{FFFFFF} 0.00\,{\scriptsize $\pm$0.04} & \cellcolor[HTML]{FFFFFF} 0.00\,{\scriptsize $\pm$0.04} & \cellcolor[HTML]{FFFFFF} 0.00\,{\scriptsize $\pm$0.04} & \cellcolor[HTML]{FFFCF9} 0.02\,{\scriptsize $\pm$0.05} & \cellcolor[HTML]{FFFDFC} 0.01\,{\scriptsize $\pm$0.01} \\
GLM-4.7 Flash & \cellcolor[HTML]{FFFFFF} 0.00\,{\scriptsize $\pm$0.04} & \cellcolor[HTML]{FFFFFF} 0.00\,{\scriptsize $\pm$0.04} & \cellcolor[HTML]{FFFFFF} 0.00\,{\scriptsize $\pm$0.04} & \cellcolor[HTML]{FFFFFF} 0.00\,{\scriptsize $\pm$0.04} & \cellcolor[HTML]{FFFCF9} 0.02\,{\scriptsize $\pm$0.05} & \cellcolor[HTML]{FFFCF9} 0.02\,{\scriptsize $\pm$0.05} & \cellcolor[HTML]{FFFEFD} 0.01\,{\scriptsize $\pm$0.01} \\
Mercury & \cellcolor[HTML]{FFFFFF} 0.00\,{\scriptsize $\pm$0.04} & \cellcolor[HTML]{FFFFFF} 0.00\,{\scriptsize $\pm$0.04} & \cellcolor[HTML]{FFFFFF} 0.00\,{\scriptsize $\pm$0.04} & \cellcolor[HTML]{FFFFFF} 0.00\,{\scriptsize $\pm$0.04} & \cellcolor[HTML]{FFFFFF} 0.00\,{\scriptsize $\pm$0.04} & \cellcolor[HTML]{FFFCF9} 0.02\,{\scriptsize $\pm$0.05} & \cellcolor[HTML]{FFFEFE} 0.00\,{\scriptsize $\pm$0.01} \\
OLMo 3.1 32B Think & \cellcolor[HTML]{FFFFFF} 0.00\,{\scriptsize $\pm$0.04} & \cellcolor[HTML]{FFFFFF} 0.00\,{\scriptsize $\pm$0.04} & \cellcolor[HTML]{FFFFFF} 0.00\,{\scriptsize $\pm$0.04} & \cellcolor[HTML]{FFFFFF} 0.00\,{\scriptsize $\pm$0.04} & \cellcolor[HTML]{FFFFFF} 0.00\,{\scriptsize $\pm$0.04} & \cellcolor[HTML]{FFFAF3} 0.04\,{\scriptsize $\pm$0.06} & \cellcolor[HTML]{FFFEFD} 0.01\,{\scriptsize $\pm$0.01} \\
\bottomrule
\end{tabular}}
\end{table}

\begin{table}[ht]
\centering
\caption{Per-model accuracy (\%) by input format. Each model's $\Delta$ rows show change relative to its own Plain ASCII baseline.}
\label{tab:format_comparison_per_model}
\resizebox{\textwidth}{!}{%
\begin{tabular}{lrrrrrr}
\toprule
Model / Format & FlowFree & Bridges & Loopy & Galaxies & Undead & Pattern \\
\midrule
\textbf{DeepSeek V3.2} & 28.7\,{\scriptsize $\pm$7.2} & 74.0\,{\scriptsize $\pm$7.0} & 8.7\,{\scriptsize $\pm$4.6} & 1.3\,{\scriptsize $\pm$2.2} & 44.0\,{\scriptsize $\pm$7.8} & 52.7\,{\scriptsize $\pm$7.9} \\
\quad $\Delta$ With Image & -- & -- & -- & -- & -- & -- \\
\quad $\Delta$ Int-Format & \cellcolor[HTML]{D5F5E3} +5.3\,{\scriptsize $\pm$10.4} & \cellcolor[HTML]{D5F5E3} +17.3\,{\scriptsize $\pm$8.4} & \cellcolor[HTML]{D5F5E3} +7.3\,{\scriptsize $\pm$7.5} & \cellcolor[HTML]{D5F5E3} +28.0\,{\scriptsize $\pm$7.6} & -26.0\,{\scriptsize $\pm$10.0} & \cellcolor[HTML]{FADBD8} -8.7\,{\scriptsize $\pm$11.1} \\
\quad $\Delta$ Int-Format JSON & +3.3\,{\scriptsize $\pm$10.3} & +14.0\,{\scriptsize $\pm$8.8} & +3.3\,{\scriptsize $\pm$7.1} & +23.3\,{\scriptsize $\pm$7.3} & \cellcolor[HTML]{FADBD8} -36.7\,{\scriptsize $\pm$9.0} & -8.0\,{\scriptsize $\pm$11.1} \\
\midrule
\textbf{Gemini 3 Flash} & 42.7\,{\scriptsize $\pm$7.8} & 58.0\,{\scriptsize $\pm$7.8} & 6.0\,{\scriptsize $\pm$3.9} & 5.3\,{\scriptsize $\pm$3.7} & 29.3\,{\scriptsize $\pm$7.2} & 66.0\,{\scriptsize $\pm$7.5} \\
\quad $\Delta$ With Image & +4.7\,{\scriptsize $\pm$11.1} & +7.3\,{\scriptsize $\pm$10.9} & -3.3\,{\scriptsize $\pm$5.1} & +0.0\,{\scriptsize $\pm$5.5} & -4.7\,{\scriptsize $\pm$10.0} & \cellcolor[HTML]{D5F5E3} +6.7\,{\scriptsize $\pm$10.3} \\
\quad $\Delta$ Int-Format & +8.0\,{\scriptsize $\pm$11.1} & +37.3\,{\scriptsize $\pm$8.6} & +7.3\,{\scriptsize $\pm$6.9} & \cellcolor[HTML]{D5F5E3} +32.0\,{\scriptsize $\pm$8.6} & -6.0\,{\scriptsize $\pm$9.9} & +3.3\,{\scriptsize $\pm$10.5} \\
\quad $\Delta$ Int-Format JSON & \cellcolor[HTML]{D5F5E3} +10.0\,{\scriptsize $\pm$11.1} & \cellcolor[HTML]{D5F5E3} +38.7\,{\scriptsize $\pm$8.5} & \cellcolor[HTML]{D5F5E3} +12.0\,{\scriptsize $\pm$7.4} & +29.3\,{\scriptsize $\pm$8.5} & \cellcolor[HTML]{FADBD8} -25.3\,{\scriptsize $\pm$8.0} & +4.7\,{\scriptsize $\pm$10.4} \\
\midrule
\textbf{GLM-4.7 Flash} & 4.7\,{\scriptsize $\pm$3.5} & 0.0\,{\scriptsize $\pm$1.2} & 2.0\,{\scriptsize $\pm$2.5} & 0.0\,{\scriptsize $\pm$1.2} & 0.7\,{\scriptsize $\pm$1.8} & 1.3\,{\scriptsize $\pm$2.2} \\
\quad $\Delta$ With Image & -- & -- & -- & -- & -- & -- \\
\quad $\Delta$ Int-Format & \cellcolor[HTML]{D5F5E3} +6.7\,{\scriptsize $\pm$6.4} & \cellcolor[HTML]{D5F5E3} +0.0\,{\scriptsize $\pm$2.5} & -0.7\,{\scriptsize $\pm$3.7} & \cellcolor[HTML]{D5F5E3} +0.0\,{\scriptsize $\pm$2.5} & +0.0\,{\scriptsize $\pm$3.1} & \cellcolor[HTML]{FADBD8} -1.3\,{\scriptsize $\pm$3.0} \\
\quad $\Delta$ Int-Format JSON & +3.3\,{\scriptsize $\pm$5.9} & \cellcolor[HTML]{D5F5E3} +0.0\,{\scriptsize $\pm$2.5} & \cellcolor[HTML]{FADBD8} -1.3\,{\scriptsize $\pm$3.5} & \cellcolor[HTML]{D5F5E3} +0.0\,{\scriptsize $\pm$2.5} & \cellcolor[HTML]{D5F5E3} +1.3\,{\scriptsize $\pm$3.5} & -0.7\,{\scriptsize $\pm$3.3} \\
\midrule
\textbf{GPT-5 Mini} & 30.0\,{\scriptsize $\pm$7.3} & 68.7\,{\scriptsize $\pm$7.3} & 10.7\,{\scriptsize $\pm$5.0} & 19.3\,{\scriptsize $\pm$6.3} & 73.3\,{\scriptsize $\pm$7.0} & 75.3\,{\scriptsize $\pm$6.8} \\
\quad $\Delta$ With Image & +0.7\,{\scriptsize $\pm$10.3} & -0.7\,{\scriptsize $\pm$10.4} & +2.0\,{\scriptsize $\pm$7.4} & -0.7\,{\scriptsize $\pm$8.9} & -5.3\,{\scriptsize $\pm$10.2} & -18.0\,{\scriptsize $\pm$10.4} \\
\quad $\Delta$ Int-Format & \cellcolor[HTML]{D5F5E3} +8.0\,{\scriptsize $\pm$10.6} & +12.0\,{\scriptsize $\pm$9.7} & +12.0\,{\scriptsize $\pm$8.4} & +15.3\,{\scriptsize $\pm$9.8} & \cellcolor[HTML]{FADBD8} -8.7\,{\scriptsize $\pm$10.3} & \cellcolor[HTML]{FADBD8} -22.0\,{\scriptsize $\pm$10.5} \\
\quad $\Delta$ Int-Format JSON & +4.7\,{\scriptsize $\pm$10.5} & \cellcolor[HTML]{D5F5E3} +15.3\,{\scriptsize $\pm$9.4} & \cellcolor[HTML]{D5F5E3} +13.3\,{\scriptsize $\pm$8.5} & \cellcolor[HTML]{D5F5E3} +17.3\,{\scriptsize $\pm$9.9} & -7.3\,{\scriptsize $\pm$10.3} & -12.7\,{\scriptsize $\pm$10.3} \\
\midrule
\textbf{Llama 4 Maverick} & 0.0\,{\scriptsize $\pm$1.2} & 0.0\,{\scriptsize $\pm$1.2} & 0.0\,{\scriptsize $\pm$1.2} & 0.0\,{\scriptsize $\pm$1.2} & 0.0\,{\scriptsize $\pm$1.2} & 0.0\,{\scriptsize $\pm$1.2} \\
\quad $\Delta$ With Image & -- & -- & -- & -- & -- & -- \\
\quad $\Delta$ Int-Format & \cellcolor[HTML]{D5F5E3} +3.3\,{\scriptsize $\pm$3.7} & \cellcolor[HTML]{D5F5E3} +0.0\,{\scriptsize $\pm$2.5} & \cellcolor[HTML]{D5F5E3} +0.0\,{\scriptsize $\pm$2.5} & \cellcolor[HTML]{D5F5E3} +0.0\,{\scriptsize $\pm$2.5} & \cellcolor[HTML]{D5F5E3} +0.0\,{\scriptsize $\pm$2.5} & \cellcolor[HTML]{D5F5E3} +0.0\,{\scriptsize $\pm$2.5} \\
\quad $\Delta$ Int-Format JSON & +0.7\,{\scriptsize $\pm$2.8} & \cellcolor[HTML]{D5F5E3} +0.0\,{\scriptsize $\pm$2.5} & \cellcolor[HTML]{D5F5E3} +0.0\,{\scriptsize $\pm$2.5} & \cellcolor[HTML]{D5F5E3} +0.0\,{\scriptsize $\pm$2.5} & \cellcolor[HTML]{D5F5E3} +0.0\,{\scriptsize $\pm$2.5} & \cellcolor[HTML]{D5F5E3} +0.0\,{\scriptsize $\pm$2.5} \\
\midrule
\textbf{Mercury} & 3.3\,{\scriptsize $\pm$3.1} & 6.7\,{\scriptsize $\pm$4.1} & 0.0\,{\scriptsize $\pm$1.2} & 0.0\,{\scriptsize $\pm$1.2} & 0.7\,{\scriptsize $\pm$1.8} & 3.3\,{\scriptsize $\pm$3.1} \\
\quad $\Delta$ With Image & -- & -- & -- & -- & -- & -- \\
\quad $\Delta$ Int-Format & \cellcolor[HTML]{D5F5E3} +6.7\,{\scriptsize $\pm$5.9} & \cellcolor[HTML]{D5F5E3} +22.0\,{\scriptsize $\pm$8.3} & \cellcolor[HTML]{D5F5E3} +0.0\,{\scriptsize $\pm$2.5} & \cellcolor[HTML]{D5F5E3} +0.0\,{\scriptsize $\pm$2.5} & \cellcolor[HTML]{FADBD8} -0.7\,{\scriptsize $\pm$2.8} & \cellcolor[HTML]{FADBD8} -3.3\,{\scriptsize $\pm$3.7} \\
\quad $\Delta$ Int-Format JSON & +4.7\,{\scriptsize $\pm$5.6} & +14.0\,{\scriptsize $\pm$7.7} & \cellcolor[HTML]{D5F5E3} +0.0\,{\scriptsize $\pm$2.5} & \cellcolor[HTML]{D5F5E3} +0.0\,{\scriptsize $\pm$2.5} & \cellcolor[HTML]{FADBD8} -0.7\,{\scriptsize $\pm$2.8} & \cellcolor[HTML]{FADBD8} -3.3\,{\scriptsize $\pm$3.7} \\
\midrule
\textbf{OLMo 3.1 32B Think} & 8.0\,{\scriptsize $\pm$4.4} & 2.7\,{\scriptsize $\pm$2.8} & 0.7\,{\scriptsize $\pm$1.8} & 0.0\,{\scriptsize $\pm$1.2} & 0.0\,{\scriptsize $\pm$1.2} & 5.3\,{\scriptsize $\pm$3.7} \\
\quad $\Delta$ With Image & -- & -- & -- & -- & -- & -- \\
\quad $\Delta$ Int-Format & \cellcolor[HTML]{D5F5E3} +4.7\,{\scriptsize $\pm$7.1} & \cellcolor[HTML]{D5F5E3} +0.7\,{\scriptsize $\pm$4.5} & +2.2\,{\scriptsize $\pm$3.9} & \cellcolor[HTML]{D5F5E3} +0.0\,{\scriptsize $\pm$2.5} & \cellcolor[HTML]{D5F5E3} +0.0\,{\scriptsize $\pm$2.5} & \cellcolor[HTML]{FADBD8} -5.3\,{\scriptsize $\pm$4.2} \\
\quad $\Delta$ Int-Format JSON & +0.7\,{\scriptsize $\pm$6.5} & \cellcolor[HTML]{D5F5E3} +0.7\,{\scriptsize $\pm$4.5} & \cellcolor[HTML]{D5F5E3} +2.7\,{\scriptsize $\pm$4.0} & \cellcolor[HTML]{D5F5E3} +0.0\,{\scriptsize $\pm$2.5} & \cellcolor[HTML]{D5F5E3} +0.0\,{\scriptsize $\pm$2.5} & -4.7\,{\scriptsize $\pm$4.4} \\
\midrule
\textbf{Qwen3-235B Thinking} & 22.0\,{\scriptsize $\pm$6.6} & 36.7\,{\scriptsize $\pm$7.6} & 0.0\,{\scriptsize $\pm$1.2} & 0.0\,{\scriptsize $\pm$1.2} & 8.7\,{\scriptsize $\pm$4.6} & 22.0\,{\scriptsize $\pm$6.6} \\
\quad $\Delta$ With Image & -- & -- & -- & -- & -- & -- \\
\quad $\Delta$ Int-Format & \cellcolor[HTML]{D5F5E3} +9.3\,{\scriptsize $\pm$9.9} & +20.7\,{\scriptsize $\pm$10.9} & \cellcolor[HTML]{D5F5E3} +0.0\,{\scriptsize $\pm$2.5} & \cellcolor[HTML]{D5F5E3} +8.0\,{\scriptsize $\pm$4.8} & \cellcolor[HTML]{FADBD8} -8.7\,{\scriptsize $\pm$5.0} & -13.3\,{\scriptsize $\pm$8.1} \\
\quad $\Delta$ Int-Format JSON & +7.3\,{\scriptsize $\pm$9.8} & \cellcolor[HTML]{D5F5E3} +22.0\,{\scriptsize $\pm$10.9} & \cellcolor[HTML]{D5F5E3} +0.0\,{\scriptsize $\pm$2.5} & +4.0\,{\scriptsize $\pm$3.9} & -8.0\,{\scriptsize $\pm$5.1} & \cellcolor[HTML]{D5F5E3} +2.0\,{\scriptsize $\pm$9.5} \\
\midrule
\textbf{Qwen3-32B} & 6.0\,{\scriptsize $\pm$3.9} & 2.0\,{\scriptsize $\pm$2.5} & 0.0\,{\scriptsize $\pm$1.2} & 0.0\,{\scriptsize $\pm$1.2} & 0.7\,{\scriptsize $\pm$1.8} & 5.3\,{\scriptsize $\pm$3.7} \\
\quad $\Delta$ With Image & -- & -- & -- & -- & -- & -- \\
\quad $\Delta$ Int-Format & \cellcolor[HTML]{D5F5E3} +2.0\,{\scriptsize $\pm$6.1} & \cellcolor[HTML]{D5F5E3} +15.3\,{\scriptsize $\pm$6.7} & \cellcolor[HTML]{D5F5E3} +0.0\,{\scriptsize $\pm$2.6} & \cellcolor[HTML]{D5F5E3} +0.0\,{\scriptsize $\pm$2.5} & \cellcolor[HTML]{FADBD8} -0.7\,{\scriptsize $\pm$2.8} & \cellcolor[HTML]{FADBD8} -5.3\,{\scriptsize $\pm$4.2} \\
\quad $\Delta$ Int-Format JSON & +0.7\,{\scriptsize $\pm$5.9} & +14.7\,{\scriptsize $\pm$6.6} & \cellcolor[HTML]{D5F5E3} +0.0\,{\scriptsize $\pm$2.6} & \cellcolor[HTML]{D5F5E3} +0.0\,{\scriptsize $\pm$2.5} & \cellcolor[HTML]{FADBD8} -0.7\,{\scriptsize $\pm$2.8} & \cellcolor[HTML]{FADBD8} -5.3\,{\scriptsize $\pm$4.2} \\
\bottomrule
\end{tabular}}
\end{table}

\subsection{Dataset Generation Parameters}
\label{app:dataset_params}

Table~\ref{tab:engine_params} lists the full engine parameter strings passed to each puzzle generator at each difficulty tier.
For the five Tatham-family puzzles, the string encodes both the board dimensions and the solver's deduction-depth setting; higher difficulty requires deeper logical chains to solve without backtracking.
Pattern is the exception: the Tatham generator does not expose a separate difficulty dial, so tiers differ only in board size.
Flow Free instances are produced by a separate generator where difficulty scales with board size and color count.

\begin{table}[h]
\centering
\small
\caption{Engine parameter strings for each puzzle family and difficulty tier.
Notation follows the Simon Tatham generator conventions: dimensions precede the difficulty code.
For Loopy, \texttt{t0} denotes the square grid type.
Difficulty codes: Bridges \texttt{d0}/\texttt{d1}/\texttt{d2} = increasing constraint complexity;
Galaxies \texttt{dn}/\texttt{du} = normal/unreasonable solver;
Loopy \texttt{de}/\texttt{dt}/\texttt{dh} = easy/tricky/hard deduction;
Undead \texttt{de}/\texttt{dn}/\texttt{dt} = easy/normal/tricky deduction.}
\label{tab:engine_params}
\begin{tabular}{llll}
\toprule
\textbf{Puzzle} & \textbf{Easy} & \textbf{Medium} & \textbf{Hard} \\
\midrule
Bridges   & \texttt{5x5d0}     & \texttt{7x7d1}     & \texttt{10x10d2}    \\
Galaxies  & \texttt{5x5dn}     & \texttt{7x7du}     & \texttt{10x10du}    \\
Loopy     & \texttt{5x5t0de}   & \texttt{7x7t0dt}   & \texttt{10x10t0dh}  \\
Pattern   & \texttt{5x5}       & \texttt{7x7}       & \texttt{10x10}      \\
Undead    & \texttt{4x4de}     & \texttt{5x5dn}     & \texttt{7x7dt}      \\
Flow Free & 5$\times$5--6$\times$6 & 7$\times$7--8$\times$8 & 9$\times$9--12$\times$12 \\
\bottomrule
\end{tabular}
\end{table}

\subsection{Orthogonality to Existing Benchmarks}
\label{subsec:orthogonality}

A natural question is whether TopoBench measures reasoning abilities that are already captured by established benchmarks, or whether it targets a genuinely distinct capability. To investigate this, we compare model performance on TopoBench against performance on two existing puzzle benchmarks (KORGym~\citep{shi2025korgym} and Enigmata~\citep{chen2025enigmata}) and several widely used reasoning benchmarks (ARC-AGI-1, ARC-AGI-2, AIME 2025, and AA Intelligence~\citep{artificialanalysis2026}).

We compute Spearman rank correlations between model scores on each benchmark pair. Existing puzzle benchmarks (KORGym, Enigmata) exhibit strong correlations with the general reasoning benchmarks, suggesting that they largely measure the same underlying capability axis. TopoBench, on average, shows weaker correlation with these same general reasoning benchmarks, indicating that the skills required to solve topology-heavy puzzles under our evaluation protocol---sustained state tracking, global constraint propagation, and representation-sensitive spatial inference---are not as well predicted by performance on standard reasoning tasks.

This finding has two implications. First, it validates TopoBench as a \emph{complementary} evaluation tool: models that score highly on existing reasoning benchmarks may still fall short on TopoBench, and vice versa. Second, topological and spatial reasoning under high state-tracking load appears to constitute a partially independent capability dimension that current benchmarks underrepresent.

% Additionally, TopoBench's hard tier is substantially more challenging than the instances in existing puzzle benchmarks. On hard-tier puzzles, even the best closed-source models achieve low aggregate accuracy (Section~\ref{subsec:positioning}), while open-source models are near zero. This difficulty ceiling means TopoBench can differentiate between model capabilities at a finer granularity than benchmarks where top models are already near saturation.

\begin{figure}[ht]
\centering
\includegraphics[width=\textwidth]{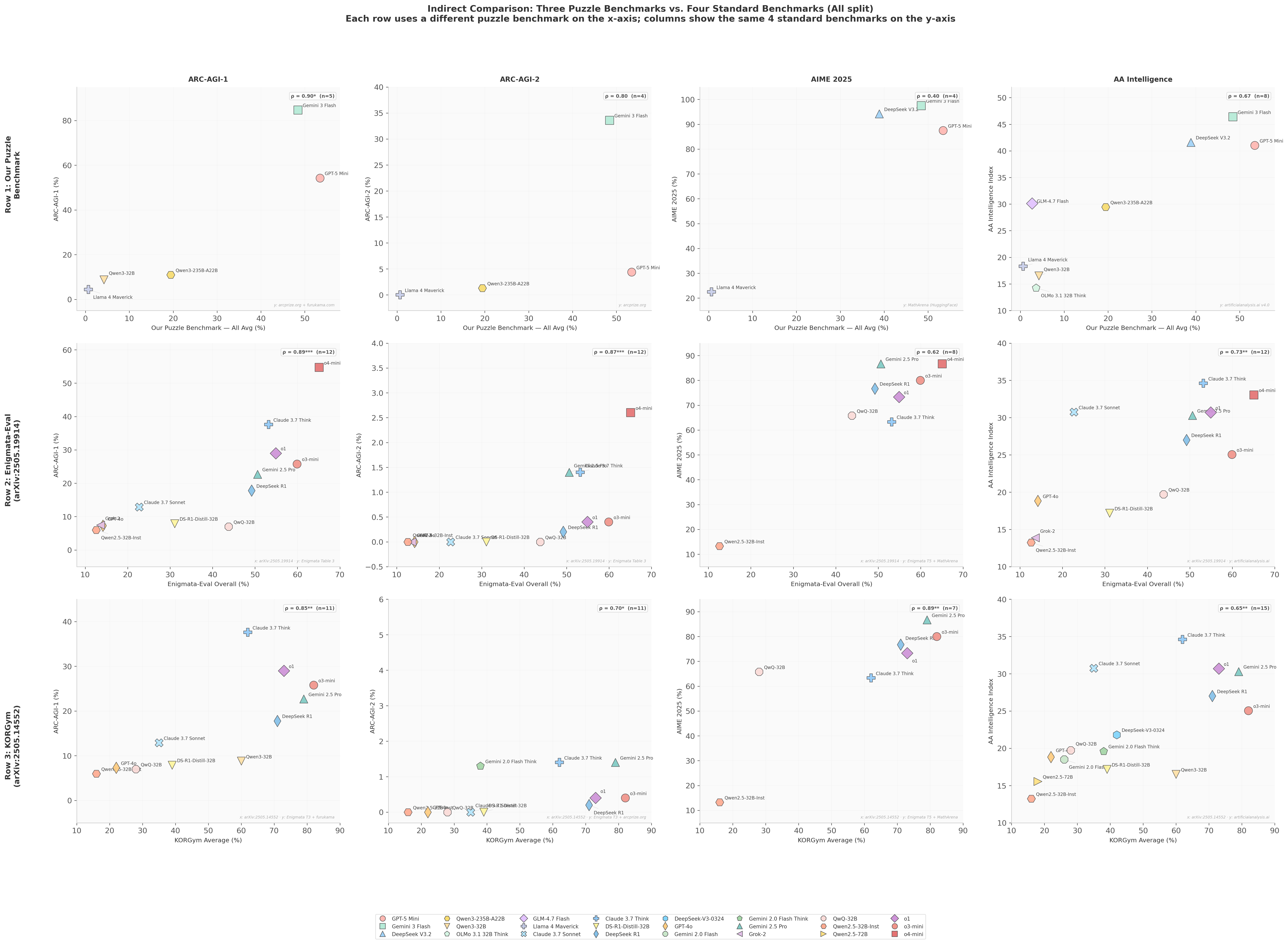}
\caption{Spearman rank correlations between model performance on TopoBench, existing puzzle benchmarks (KORGym, Enigmata), and general reasoning benchmarks (ARC-AGI-1/2, AIME 2025, AA Intelligence). All puzzle benchmarks correlate with existing benchmarks.
%Existing puzzle benchmarks correlate strongly with general reasoning; TopoBench shows weaker correlation, suggesting it targets a partially independent capability axis.
}
\label{fig:orthogonality}
\end{figure}

\definecolor{loopycolor}{RGB}{217,234,211}
\definecolor{undeadcolor}{RGB}{242,220,219}
\definecolor{flowfreecolor}{RGB}{207,226,243}
\definecolor{bridgescolor}{RGB}{255,242,204}
\definecolor{galaxiescolor}{RGB}{234,209,220}
\definecolor{patterncolor}{RGB}{208,224,227}

\newcommand{\PuzzleFigure}[3]{
\subsection{#2}
\vspace{-2mm}
\noindent
\begin{minipage}[t]{0.48\textwidth}
\centering
\begin{tikzpicture}
\node[fill=#3, rounded corners=8pt, inner sep=10pt]{
\includegraphics[width=0.8\linewidth]{figures/puzzle_images_easy/#1_problem}
};
\end{tikzpicture}

\vspace{2mm}
\textbf{Problem}
\end{minipage}
\hfill
\begin{minipage}[t]{0.48\textwidth}
\centering
\begin{tikzpicture}
\node[fill=#3, rounded corners=8pt, inner sep=10pt]{
\includegraphics[width=0.8\linewidth]{figures/puzzle_images_easy/#1_solution}
};
\end{tikzpicture}

\vspace{2mm}
\textbf{Solution}
\end{minipage}
\par
\vspace{2mm}
}

\section{Puzzle Examples}

\PuzzleFigure{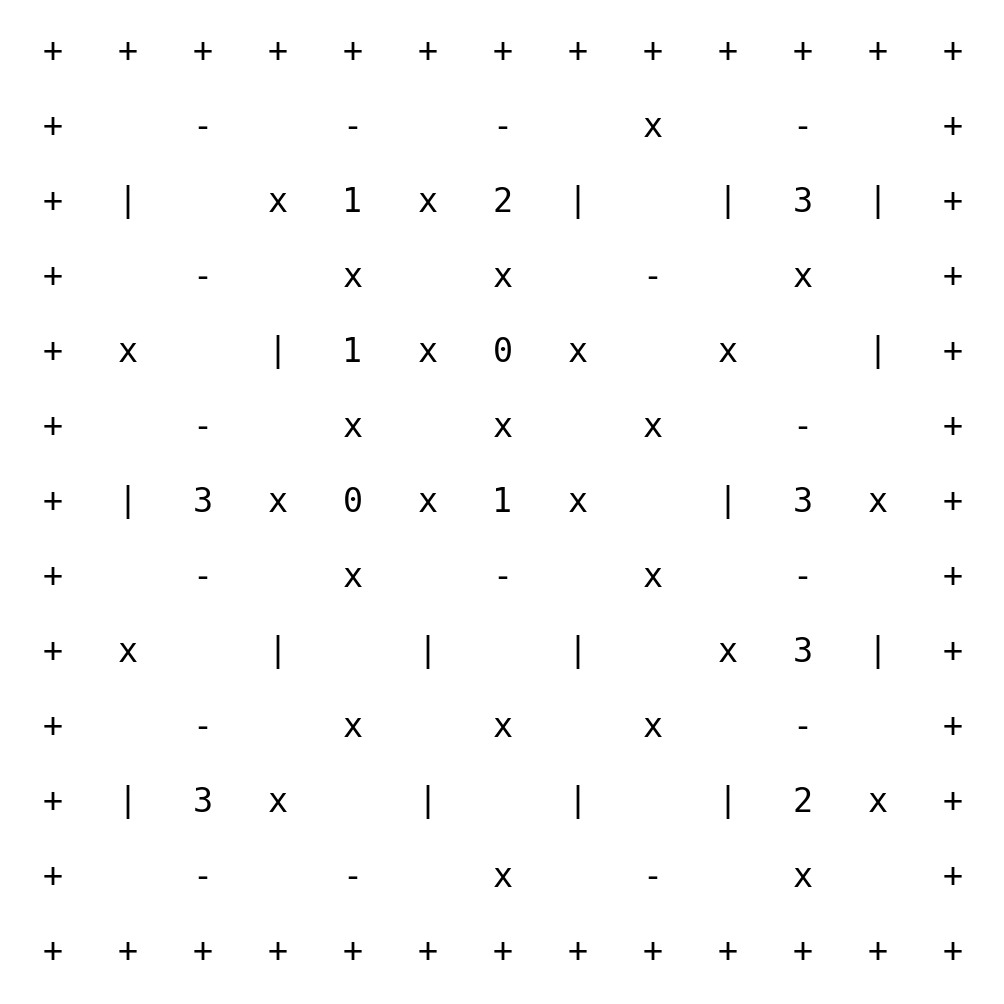}{Loopy}{loopycolor}

\PuzzleFigure{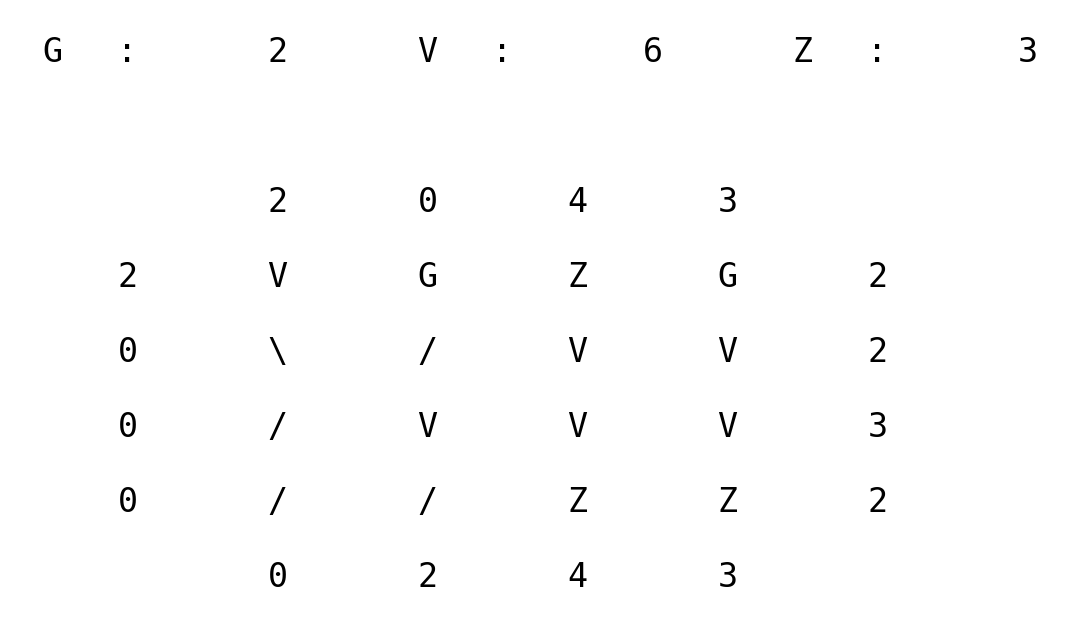}{Undead}{undeadcolor}

\PuzzleFigure{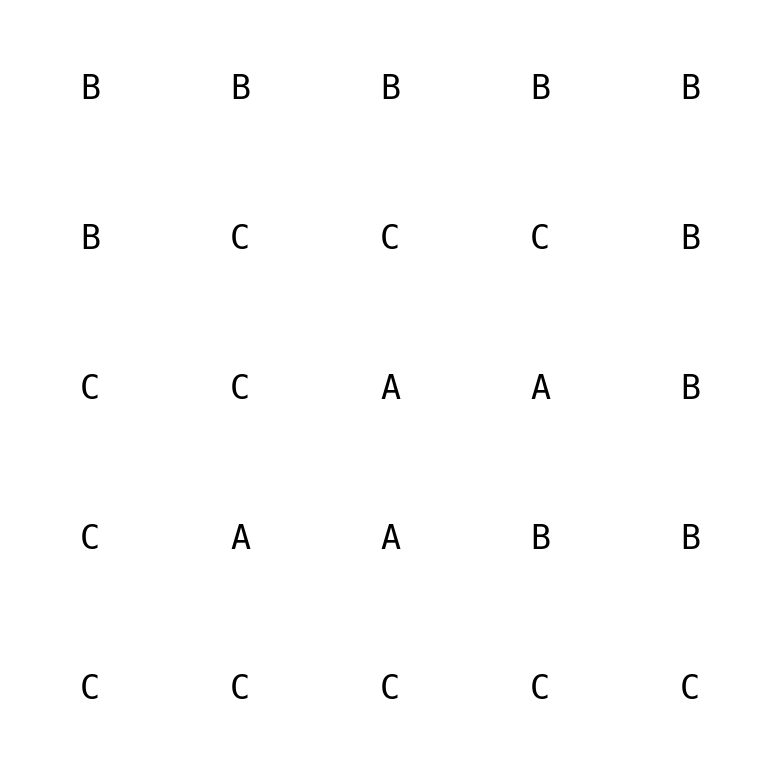}{Flow Free}{flowfreecolor}

\PuzzleFigure{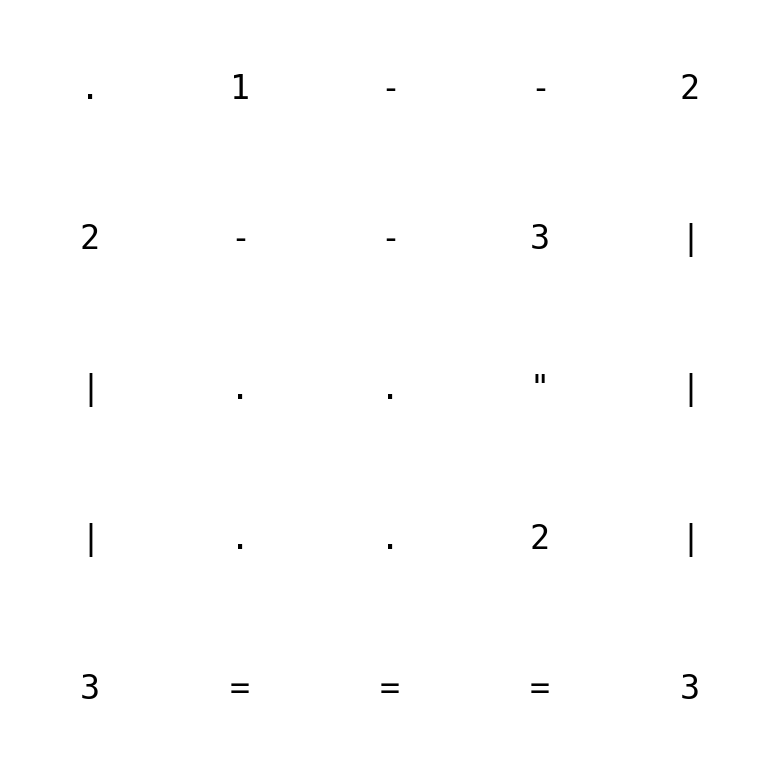}{Bridges}{bridgescolor}

\PuzzleFigure{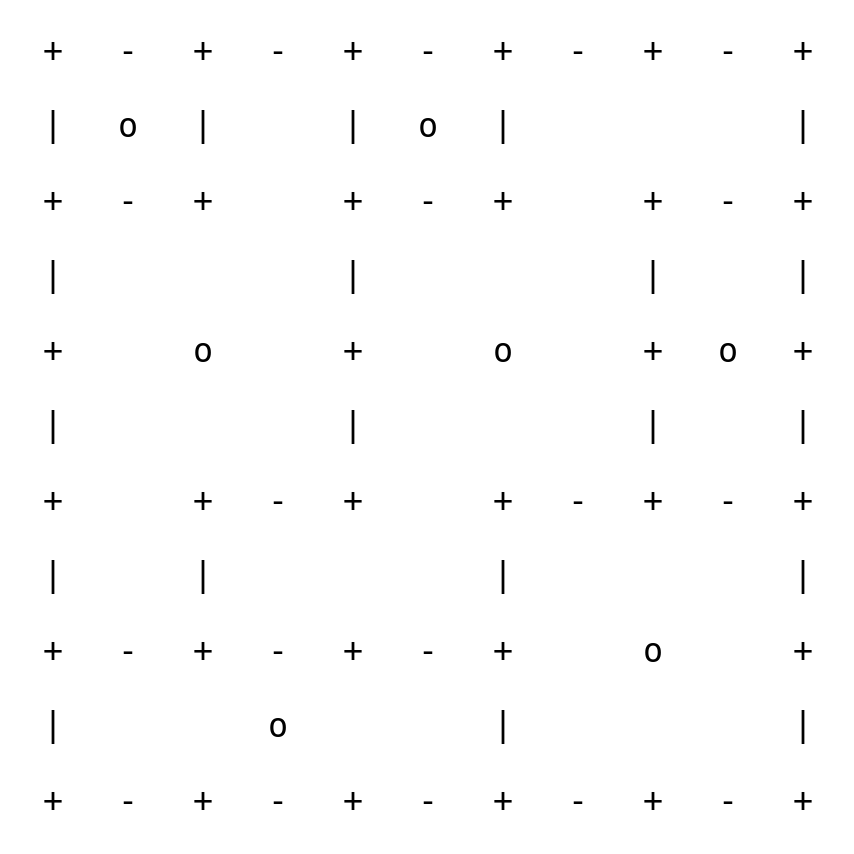}{Galaxies}{galaxiescolor}

\PuzzleFigure{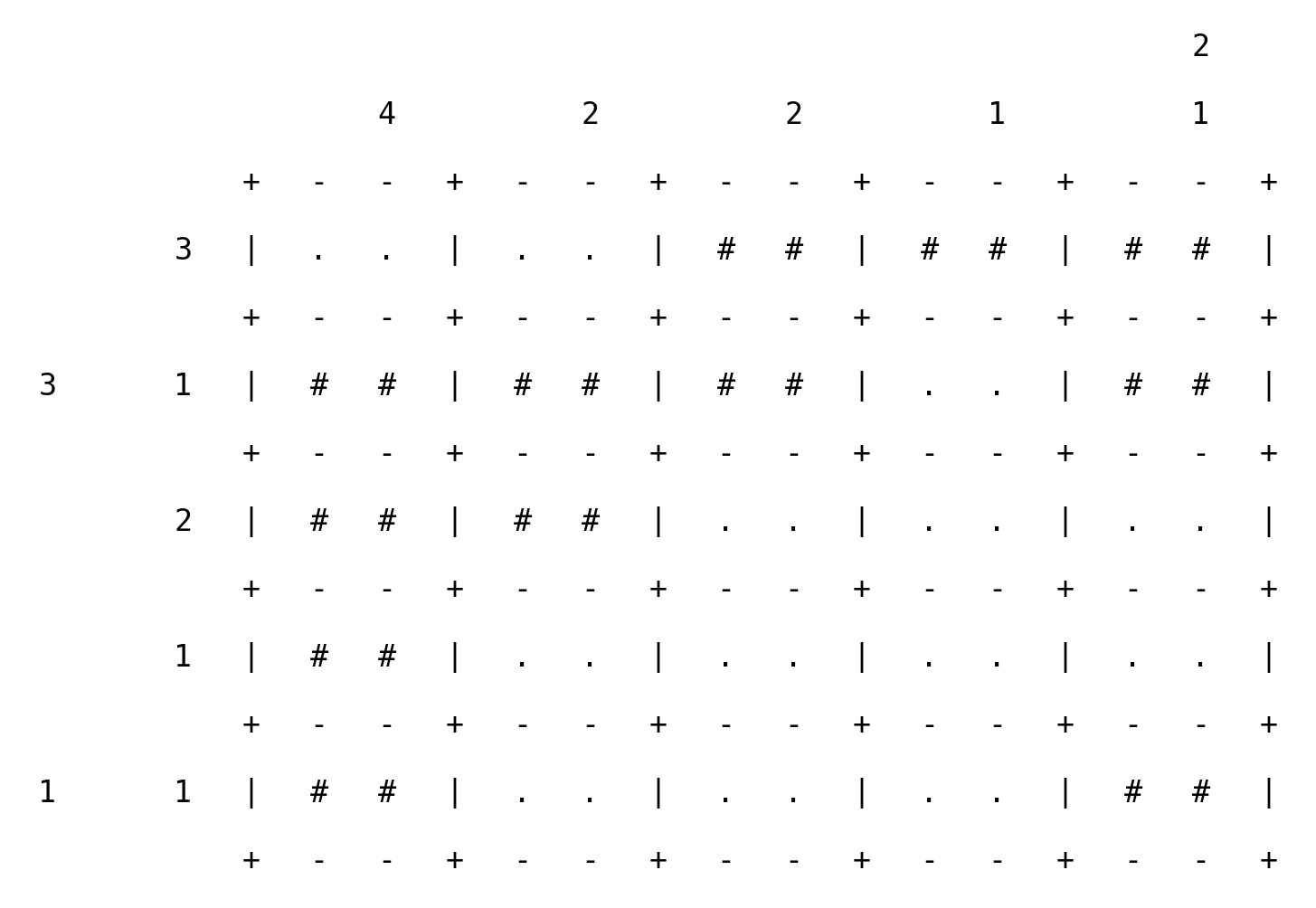}{Pattern}{patterncolor}

\section{Full Error Taxonomy}
\label{app:full_error_taxonomy}

Table~\ref{tab:error_taxonomy_full} lists all eleven error categories used in the LLM-as-judge analysis.
The seven categories shown in Table~\ref{tab:error_taxonomy_judge} drive the main narrative; the four additional categories below (UE, HV, TD, OTH) are low-frequency or show no clear difficulty trend and are omitted from the main-body figures for clarity.

\begin{table*}[h]
\centering
\small
\setlength{\tabcolsep}{5pt}
\renewcommand{\arraystretch}{1.25}
\caption{Full eleven-category error taxonomy.
Categories above the first double rule are plausibly \emph{causal}; those below are \emph{symptomatic}.
The last four rows (below the second double rule) are omitted from the main-body figures.}
\label{tab:error_taxonomy_full}
\begin{tabular}{cl p{0.68\textwidth}}
\toprule
\textbf{Abbr.} & \textbf{Error type} & \textbf{Definition} \\
\midrule
RR & Repeated Reasoning &
Near-identical move sequences repeated from the same board position without meaningful variation. \\
\midrule
STF & State-Tracking Failure &
Claimed board state diverges from the cumulative effect of the model's own stated actions. \\
\midrule
CF & Constraint Forgetting &
An explicit action that directly violates structural puzzle rules (overcounting, crossing, overwriting fixed cells). \\
\midrule
PC & Premature Commitment &
Commits to a provably incorrect configuration and persists for 3+ steps. \\
\midrule
\midrule
ES & Explicit Surrender &
Explicitly gives up, requests a solver, or states it cannot proceed. \\
IO & Incomplete Output &
Fails to produce a complete, valid final answer. \\
RD & Representation Drift &
Internal representation of the problem shifts over time, causing inconsistent reasoning. \\
\midrule
\midrule
UE & Unjustified Elimination &
Prematurely eliminates valid possibilities without sufficient justification. \\
HV & Hallucinated Validation &
Asserts a solution is correct without proper verification or despite contradictions. \\
TD & Topic Drift &
Shifts focus away from the original problem. \\
OTH & Other &
Failure mode outside the above categories. \\
\bottomrule
\end{tabular}
\end{table*}

\section{Per-Puzzle Error Analysis}
\label{app:per_puzzle_errors}

\begin{figure}[h]
\centering
\includegraphics[width=\textwidth]{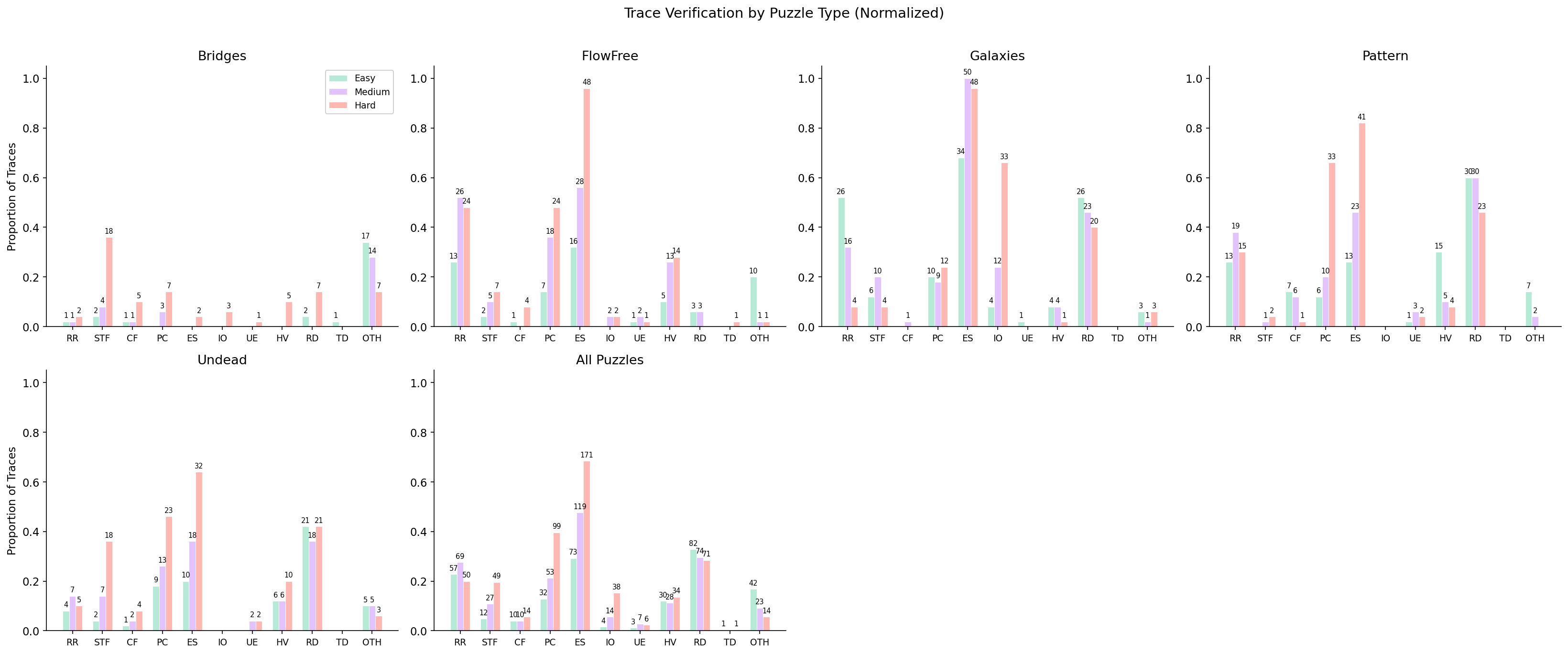}
\caption{Error category prevalence by puzzle type and difficulty tier (50 traces per cell).
Galaxies and FlowFree are dominated by ES even at lower difficulties.
FlowFree has the strongest PC scaling.
Pattern exhibits elevated RR and extreme RD across all tiers.
Bridges has the lowest ES/RR rates but shows concentrated STF and PC at hard.}
\label{fig:error_per_puzzle}
\end{figure}

Figure~\ref{fig:error_per_puzzle} provides the full per-puzzle breakdown of error category prevalence across all five puzzle families and three difficulty tiers (50 traces per cell, 750 total).

Several puzzle-specific patterns complement the aggregate trends reported in Section~\ref{sec:diagnosis}.

\textbf{FlowFree.}
FlowFree exhibits the strongest PC scaling (14\% $\rightarrow$ 36\% $\rightarrow$ 48\% from easy to hard), consistent with the path-routing nature of the puzzle where an early wrong path blocks later flows.
ES climbs to 96\% at hard, and RR peaks at 52\% at medium before declining to 48\% at hard.

\textbf{Galaxies.}
Galaxies reaches 68\% ES even at easy and 100\% at medium, the earliest saturation of any puzzle.
PC peaks at hard (24\%) but remains moderate, because the model surrenders before making meaningful commitments.
RR peaks at easy (52\%) then collapses to 8\% at hard, following the same surrender-driven decline.

\textbf{Pattern.}
Pattern exhibits extreme representation drift rates (60\% at easy, 60\% at medium, 46\% at hard), far above any other puzzle type.
Pattern puzzles require reasoning simultaneously about row and column constraints, which may naturally lead to shifting between different ways of referring to grid elements.
RR peaks at 38\% at medium, and PC shows a sharp jump to 66\% at hard.

\textbf{Undead.}
Undead shows a sharp PC jump to 46\% at hard, and elevated STF across all tiers (4\% $\rightarrow$ 14\% $\rightarrow$ 36\%), reflecting the richer state involving monster types and visibility constraints.

\textbf{Bridges.}
Bridges retains the lowest ES and RR rates across all difficulties, but the hard tier reveals substantial STF (36\%), PC (14\%), and RD (14\%).
The clean easy/medium profile paired with concentrated hard-tier failures makes Bridges a natural target for causal interventions.

\subsection{Early Stopping and Token Usage}
\label{app:es_tokens}

To support the claim that Explicit Surrender (ES) is a downstream consequence of reasoning degradation rather than an arbitrary abandonment, we compare output token counts across trace groups.
We use 900 DeepSeek V3.2 traces (50 per puzzle $\times$ difficulty cell) with GPT-5-mini judge labels, joining on token counts from the original evaluation log.

\begin{table}[h]
\centering
\small
\caption{Output token counts and reasoning trace length by outcome group. ES traces cluster near the token ceiling with notably low variance; no ES trace occurs below 22k tokens.}
\label{tab:es_tokens}
\begin{tabular}{lrrrr}
\toprule
\textbf{Group} & \textbf{N} & \textbf{Mean tokens} & \textbf{Median tokens} & \textbf{Std} \\
\midrule
Correct       & 308 & 21,801 & 18,368 & 12,174 \\
Other failure & 128 & 34,632 & 34,924 & 15,423 \\
ES failure    & 464 & 42,425 & 42,326 &  8,713 \\
\bottomrule
\end{tabular}
\end{table}

On average, ES traces use 1.23$\times$ more output tokens than other failed traces, and the minimum token count for any ES trace is 22,305 — with no short ES traces observed.
The low standard deviation for ES (8,713 compared to 15,423 for other failures) is consistent with a model approaching its reasoning limit and surrendering, rather than failing at random points in the solution process.
Reasoning trace length (raw thinking text) shows a similar pattern: ES traces average 122k characters compared to 96k for other failures.

\textbf{Caveat.} This analysis does not fully disentangle puzzle difficulty from ES tendency: harder instances produce both more ES labels and longer traces.
A within-cell comparison of ES against other failures at the same puzzle and difficulty is limited by small other-failure sample sizes in most cells.

\section{Error Prevalence by Outcome}
\label{app:correctness_delta}

\begin{figure}[h]
\centering
\includegraphics[width=\textwidth]{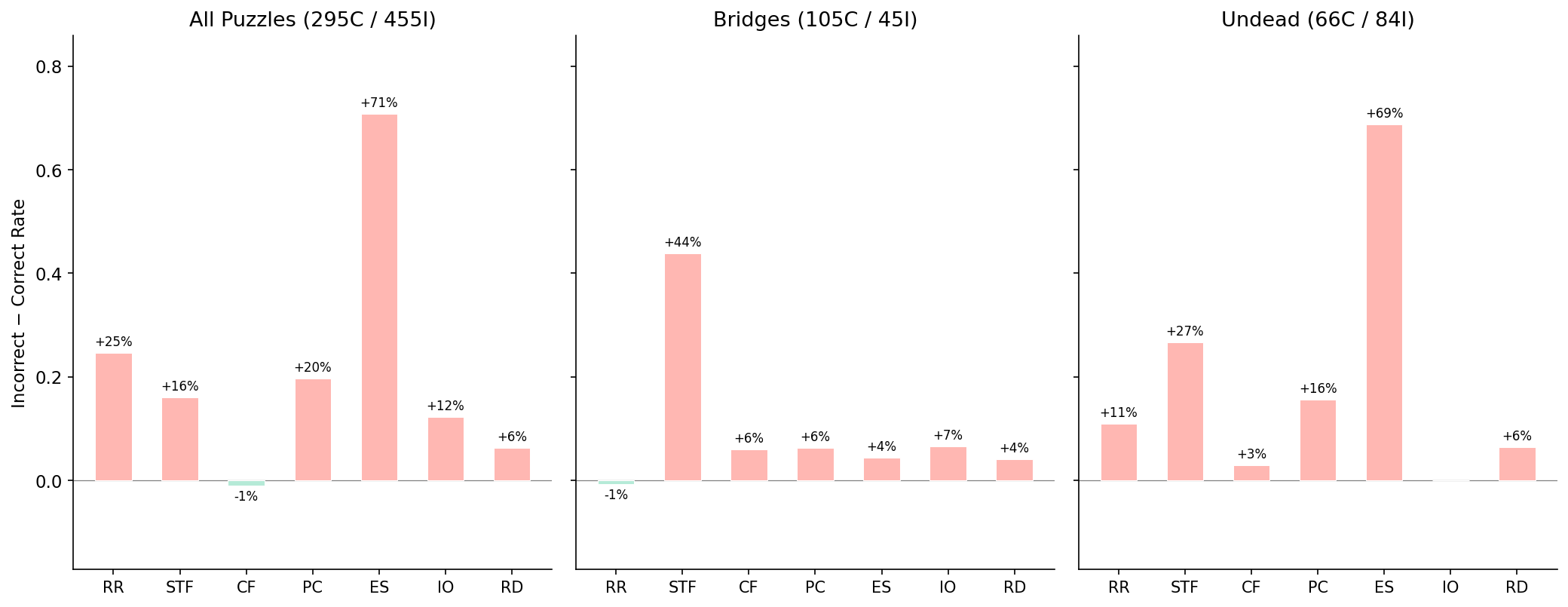}
\caption{Difference in error prevalence between incorrect and correct traces (incorrect rate $-$ correct rate), aggregated across difficulty tiers.
Positive bars indicate errors more common in failed solves.
ES dominates the aggregate because it is a downstream symptom of failure.
On Bridges, STF is the strongest discriminator (+44\,pp); on Undead, ES (+69\,pp) and STF (+27\,pp) lead.
CF is near zero across all panels despite its large causal effect in interventions (Section~\ref{sec:diagnosis}), reinforcing that observational association---like raw frequency---is a poor proxy for causal impact.}
\label{fig:delta_aggregated}
\end{figure}

To assess which error categories discriminate between correct and incorrect solutions, we compute the difference in error rates between the two outcome groups.
For each error category, we calculate its prevalence among incorrect traces and subtract its prevalence among correct traces, pooling across all three difficulty tiers.
A positive value indicates the error is more frequent when the model fails; a negative value indicates it appears more often in successful solves.
Figure~\ref{fig:delta_aggregated} reports these deltas for the same 750 traces analyzed in Section~\ref{sec:diagnosis}.

The ranking by correctness-association (ES $\gg$ RR $>$ PC $>$ STF $>$ CF in the aggregate) bears little resemblance to the causal ranking established by interventions (PC $\approx$ CF $>$ STF $\gg$ RR).
ES and RR rank highest here because they co-occur with failure rather than cause it: the model surrenders or cycles \emph{after} its reasoning has already degraded.
CF is the starkest case---it is essentially non-discriminating between outcomes ($-1$\,pp aggregate), yet injecting a single constraint violation produces an 10--11\,pp accuracy drop.
This provides a complementary perspective to the frequency-based analysis in Section~\ref{sec:diagnosis}: neither how often an error appears, nor how strongly it correlates with failure, reliably predicts its causal role.

\section{Intervention Details}
\label{app:intervention_details}

This appendix provides per-difficulty breakdowns, the full cross-puzzle comparison, and detailed per-condition analysis for the intervention experiments summarized in Figure~\ref{fig:intervention_forest}.
We include these detailed tables for completeness; the main text summarizes the key findings.

\subsection{Bridges Per-Difficulty Breakdown}

Table~\ref{tab:intervention_results_bridges} reports accuracy by difficulty tier for Bridges (100 puzzles per tier, 300 per condition).

\begin{table}[h]
\centering
\small
\caption{Intervention results on Bridges (accuracy \%, 100 puzzles per difficulty).
95\% Wilson CIs shown for the Total column.
$^*$Significant at $\alpha{=}0.05$ (95\% Newcombe CI for $\Delta$ excludes zero).}
\label{tab:intervention_results_bridges}
\begin{tabular}{lccccc}
\toprule
\textbf{Condition} & \textbf{Easy} & \textbf{Medium} & \textbf{Hard} & \textbf{Total} & $\boldsymbol{\Delta}$ \\
\midrule
Baseline            & 91.7 & 82.0 & 32.0 & 66.2 {\scriptsize [60.4, 71.5]} & -- \\
\midrule
RR Gold Ctrl        & 91.7 & 85.0 & 32.0 & 67.3 {\scriptsize [61.5, 72.6]} & $+$1.1 \\
RR Rand Ctrl        & 93.5 & 80.0 & 30.0 & 65.7 {\scriptsize [59.9, 71.0]} & $-$0.5 \\
RR Standard         & 94.7 & 76.0 & 24.0 & 62.2 {\scriptsize [56.3, 67.7]} & $-$4.0 \\
\midrule
STF $k{=}3$         & 88.9 & 73.0 & 19.0 & 58.4 {\scriptsize [52.5, 64.0]} & $-$7.8 \\
\midrule
PC Half             & 93.9 & 43.0 &  8.0 & 45.4 {\scriptsize [39.7, 51.2]} & $-$20.8$^*$ \\
CF Random           & 87.3 & 66.0 & 20.0 & 55.6 {\scriptsize [49.7, 61.3]} & $-$10.6$^*$ \\
\bottomrule
\end{tabular}
\end{table}

\subsection{Undead Per-Difficulty Breakdown}

Table~\ref{tab:intervention_results_undead} reports accuracy by difficulty tier for Undead (100 puzzles per tier, 300 per condition).

\begin{table}[h]
\centering
\small
\caption{Intervention results on Undead (accuracy \%, 100 puzzles per difficulty).
95\% Wilson CIs shown for the Total column.
$^*$Significant at $\alpha{=}0.05$ (95\% Newcombe CI for $\Delta$ excludes zero).}
\label{tab:intervention_results_undead}
\begin{tabular}{lccccc}
\toprule
\textbf{Condition} & \textbf{Easy} & \textbf{Medium} & \textbf{Hard} & \textbf{Total} & $\boldsymbol{\Delta}$ \\
\midrule
Baseline            & 81.0 & 44.0 &  9.0 & 44.7 {\scriptsize [39.1, 50.3]} & -- \\
\midrule
RR Gold Ctrl        & 82.0 & 45.0 &  6.0 & 44.3 {\scriptsize [38.8, 50.0]} & $-$0.3 \\
RR Rand Ctrl        & 71.0 & 33.0 &  3.0 & 35.7 {\scriptsize [30.5, 41.2]} & $-$9.0$^*$ \\
RR Standard         & 70.0 & 45.0 &  2.0 & 39.0 {\scriptsize [33.7, 44.6]} & $-$5.7 \\
\midrule
STF $k{=}3$         & 63.0 & 35.0 &  1.0 & 33.0 {\scriptsize [27.9, 38.5]} & $-$11.7$^*$ \\
\midrule
PC Half             & 74.0 & 26.0 &  0.0 & 33.3 {\scriptsize [28.2, 38.8]} & $-$11.3$^*$ \\
CF Random           & 67.0 & 28.0 &  5.0 & 33.3 {\scriptsize [28.2, 38.8]} & $-$11.3$^*$ \\
\bottomrule
\end{tabular}
\end{table}

\subsection{Cross-Puzzle Comparison: Bridges vs Undead}

To test whether the causal findings from Bridges generalize beyond a single puzzle type, we replicate the intervention experiments on Undead puzzles across three difficulty tiers (Easy/4$\times$4, Medium/5$\times$5, Hard/7$\times$7), with 100 puzzles per tier and 300 per condition.
The same six conditions are evaluated using the same model (DeepSeek-v3.2).
Undead CF violation subtypes are adapted to the puzzle mechanics: visibility count alteration, monster count alteration, and extra monster placement.

\begin{table}[h]
\centering
\small
\caption{Cross-puzzle comparison of intervention effects (accuracy \%).
Both puzzles: 100 puzzles per difficulty ($N{=}300$).
95\% Wilson CIs for individual proportions; Newcombe hybrid score CIs for $\Delta$.
$^*$Significant at $\alpha{=}0.05$ (95\% CI for $\Delta$ excludes zero).}
\label{tab:undead_intervention_results}
\begin{tabular}{lcccc}
\toprule
\textbf{Condition} & \multicolumn{2}{c}{\textbf{Bridges} ($N{=}300$)} & \multicolumn{2}{c}{\textbf{Undead} ($N{=}300$)} \\
\cmidrule(lr){2-3} \cmidrule(lr){4-5}
 & \textbf{Total} & $\boldsymbol{\Delta}$ & \textbf{Total} & $\boldsymbol{\Delta}$ \\
\midrule
Baseline            & 66.2 {\scriptsize [60.4, 71.5]} & --      & 44.7 {\scriptsize [39.1, 50.3]} & -- \\
\midrule
RR Gold Ctrl        & 67.3 {\scriptsize [61.5, 72.6]} & $+$1.1  & 44.3 {\scriptsize [38.8, 50.0]} & $-$0.3 \\
RR Standard         & 62.2 {\scriptsize [56.3, 67.7]} & $-$4.0  & 39.0 {\scriptsize [33.7, 44.6]} & $-$5.7 \\
\midrule
STF $k{=}3$         & 58.4 {\scriptsize [52.5, 64.0]} & $-$7.8  & 33.0 {\scriptsize [27.9, 38.5]} & $-$11.7$^*$ \\
\midrule
PC Half             & 45.4 {\scriptsize [39.7, 51.2]} & $-$20.8$^*$ & 33.3 {\scriptsize [28.2, 38.8]} & $-$11.3$^*$ \\
CF Random           & 55.6 {\scriptsize [49.7, 61.3]} & $-$10.6$^*$ & 33.3 {\scriptsize [28.2, 38.8]} & $-$11.3$^*$ \\
\bottomrule
\end{tabular}
\end{table}

\subsection{Detailed Per-Condition Analysis}

\textbf{Premature commitment.}
PC produces a $-$20.8pp accuracy drop on Bridges, the largest causal effect in our study, with its Newcombe 95\% CI [$-$28.6, $-$12.6] cleanly excluding zero.
Because the wrong path diverges from the very first move, PC tests whether the model can detect an inherited wrong trajectory and course-correct; the results show it largely cannot.
PC replaces the entire gold prefix---a more severe perturbation than the last-step modifications used by STF and CF---but the per-difficulty breakdown (Table~\ref{tab:intervention_results_bridges}) shows Easy accuracy is unaffected (93.9\% vs.\ 91.7\% baseline), while Medium drops sharply (43.0\% vs.\ 82.0\%) and Hard collapses to 8.0\%.
The effect concentrates on tiers where the search space is larger and recovery from a wrong branch requires more extensive backtracking.

\textbf{Constraint forgetting.}
CF produces a $-$10.6pp drop on Bridges (Newcombe CI [$-$18.6, $-$2.5]) and $-$11.3pp on Undead, both significant.
For structural violations (overcounting, crossing), the action text describes the illegal move and the grid updates consistently with it---the corruption is internally consistent and detectable only through semantic verification of puzzle constraints (e.g., ``does this island already have its full bridge count?'').
Once an illegal state is introduced, the model trusts and builds on it without verifying legality.

\textbf{State-tracking failure.}
STF $k{=}3$ ($-$7.8pp on Bridges) is borderline: its Newcombe 95\% CI is [$-$15.7, $+$0.3], with the upper bound just barely above zero.
On Undead, STF reaches significance at $-$11.7pp.
The contrast with CF is informative: STF introduces a \emph{syntactic} inconsistency between two representations of the same state (the corrupted grid versus the intact action log), whereas CF produces an \emph{internally consistent} state that violates puzzle rules---a \emph{semantic} error detectable only by active constraint checking.
The model's partial resilience to STF suggests it can leverage the uncorrupted action history to detect grid corruption, while the larger CF effect ($-$11pp vs.\ $-$8pp on Bridges) indicates that verifying move legality against puzzle constraints is harder.

\textbf{Repeated reasoning.}
RR Standard ($-$4.0pp on Bridges, $-$5.7pp on Undead) does not differ significantly from the baseline; Newcombe CIs contain zero on both puzzles.
Length-matched gold ($+$1.1pp) and random ($-$0.5pp) controls also show no significant effect, confirming that neither the specific wrong-path content nor the mere extension of context length causally affects performance.
We conclude that the repeated reasoning patterns observed in incorrect chains of thought are more likely a \emph{symptom} of the model searching for a solution rather than a causal contributor to failure.

\subsection{Undead-Specific Findings}

\textbf{Consistent findings.}
CF and PC are significant on both puzzles, confirming that constraint violations and premature commitment are robustly harmful across puzzle types.
RR is a null result on both puzzles: RR Standard does not differ meaningfully from baseline, and Gold Ctrl ($-$0.3pp on Undead, $+$1.1pp on Bridges) confirms that added context length alone does not degrade performance.

\textbf{STF is puzzle-dependent.}
STF $k{=}3$ produces a significant $-$11.7pp effect on Undead but only a borderline $-$7.8pp trend on Bridges (Newcombe CI upper bound $+$0.3pp).
Undead puzzles require tracking a richer state (monster types across a grid with visibility constraints), which may make the model more sensitive to grid/text inconsistencies introduced by STF.
On Undead, STF, PC, and CF all converge around $-$11pp, whereas on Bridges the effects are more differentiated: PC ($-$20.8pp) separates clearly from CF ($-$10.6pp) and STF ($-$7.8pp).

\textbf{Limitations.}
The Undead baseline is substantially lower than Bridges (44.7\% vs.\ 66.2\%), and hard Undead accuracy is near floor (9\% baseline).
RR Standard shows a negative trend on both puzzles ($-$4 to $-$6pp), though it does not reach significance on either puzzle.

\subsection{Baseline Calibration Against Standard Evaluation}
\label{app:baseline_calibration}

A potential concern with the intervention methodology is that presenting the model with a gold partial trace (the first ${\sim}$15\% of the solution) could shift the solving distribution relative to the model's self-generated reasoning.
If the gold prefix substantially altered behavior, the measured intervention effects might reflect an interaction between the prefix and the injected error rather than the error alone.

To quantify this gap, we compare the intervention baseline (gold prefix, no injection) against DeepSeek-v3.2's accuracy on the same puzzle families under the standard benchmark prompt, where the model solves from scratch with no prefix.
Both evaluations use identical grid sizes per difficulty tier (Bridges: $5{\times}5$/$7{\times}7$/$10{\times}10$, Undead: $4{\times}4$/$5{\times}5$/$7{\times}7$).
The benchmark evaluation uses 50 puzzles per tier ($N{=}150$), while the intervention baseline uses 100 per tier ($N{=}300$), drawn from independent samples produced by the same puzzle generator.

\begin{table}[h]
\centering
\small
\caption{Comparison of DeepSeek-v3.2 accuracy (\%) under the standard benchmark prompt (no prefix) and the intervention baseline (gold prefix at 15\% depth, no error injection).
Grid sizes are identical across evaluations.
Differences are within the range expected from sampling variation and minor prompt format differences.}
\label{tab:baseline_calibration}
\begin{tabular}{llcccc}
\toprule
\textbf{Puzzle} & \textbf{Condition} & \textbf{Easy} & \textbf{Medium} & \textbf{Hard} & \textbf{Total} \\
\midrule
Bridges & Standard prompt ($N{=}150$) & 94.0 & 88.0 & 40.0 & 74.0 \\
Bridges & Intervention baseline ($N{=}300$) & 91.7 & 82.0 & 32.0 & 66.2 \\
\midrule
Undead  & Standard prompt ($N{=}150$) & 76.0 & 46.0 & 10.0 & 44.0 \\
Undead  & Intervention baseline ($N{=}300$) & 81.0 & 44.0 &  9.0 & 44.7 \\
\bottomrule
\end{tabular}
\end{table}

Table~\ref{tab:baseline_calibration} shows that the two conditions produce comparable accuracy.
On Undead, per-tier differences are at most 5\,pp and the totals are within 1\,pp (44.0\% vs.\ 44.7\%).
On Bridges, the intervention baseline is somewhat lower (66.2\% vs.\ 74.0\%), with the gap concentrated at medium and hard tiers.
Two factors likely contribute to this difference.
First, the intervention prompt uses a structured output format with explicit move logging and intermediate grid rendering, which constrains generation more tightly than the standard benchmark prompt.
Second, the puzzle instances differ between the two evaluations (independent samples of 50 vs.\ 100 per tier), so sampling variability accounts for part of the spread.
The key observation is that the gold prefix does not produce a systematic boost or degradation relative to standard evaluation.
This confirms that the intervention baseline operates in a comparable performance regime, and that the accuracy drops observed under error injection (Section~\ref{sec:diagnosis}) reflect the causal effect of the injected errors rather than an artifact of the prefix format.

\section{Prompt-Based Interventions Targeting Premature Commitment}
\label{app:prompt_interventions}

Given the causal role of premature commitment identified in Section~\ref{sec:diagnosis}, we investigated whether prompt-level interventions can improve accuracy on full (non-partial) Bridges puzzles.
We evaluate DeepSeek-v3.2 on 50 medium (7$\times$7) and 50 hard (10$\times$10) puzzles per condition, testing 10 conditions that span text-only strategy instructions, few-shot worked examples, and recovery demonstrations.

\textbf{Conditions.}
Each condition augments the baseline prompt (puzzle rules plus a minimal input/output example) with one of the following:
\begin{itemize}[nosep,leftmargin=*]
\item \emph{Text backtrack instruction}: a single sentence directing the model to backtrack when a bridge placement leads to a dead end or contradiction.
\item \emph{1-shot gold path}: a 5$\times$5 solution trace showing moves and intermediate grids, without natural-language reasoning.
\item \emph{1-shot worked example}: a 7$\times$7 verbose worked example with natural-language reasoning accompanying each move.
\item \emph{2-shot worked example}: two verbose worked examples (5$\times$5 and 7$\times$7).
\item \emph{Text planning instruction}: a strategy directive to count neighbors, identify forced moves, propagate constraints, and defer guessing until forced moves are exhausted.
\item \emph{1-shot planned example}: a worked example preceded by explicit constraint analysis, demonstrating the planning strategy on a 5$\times$5 puzzle.
\item \emph{Text self-correct instruction}: a post-hoc verification directive asking the model to check each island's bridge count after producing a solution and fix any violated constraints.
\item \emph{1-shot PC recovery (concise)}: a concise example drawn from real game-tree data showing a wrong bridge placement that leads to a contradiction, followed by explicit backtracking and the correct continuation (${\sim}100$ lines of prompt).
\item \emph{1-shot PC recovery (verbose)}: the same recovery trace augmented with natural-language reasoning explaining why the wrong choice fails (${\sim}140$ lines of prompt).
\end{itemize}

\begin{table}[h]
\centering
\small
\caption{Prompt-based intervention results on Bridges (DeepSeek-v3.2).
Each condition is evaluated on 50 medium and 50 hard puzzles ($N{=}100$).
$\Delta$ is change relative to baseline.}
\label{tab:prompt_interventions}
\begin{tabular}{clcccc}
\toprule
\# & \textbf{Condition} & \textbf{Medium} & \textbf{Hard} & \textbf{Total} & $\boldsymbol{\Delta}$ \\
\midrule
1 & Baseline (rules + minimal I/O) & 80\% & 40\% & 60\% & -- \\
\midrule
2 & Text backtrack instruction & 84\% & 28\% & 56\% & $-$4 \\
3 & 1-shot gold path (no NL reasoning) & 76\% & 42\% & 59\% & $-$1 \\
4 & 1-shot worked example (verbose NL) & 86\% & 28\% & 57\% & $-$3 \\
5 & 2-shot worked example (verbose NL) & 76\% & 30\% & 53\% & $-$7 \\
6 & Text planning instruction & 78\% & 30\% & 54\% & $-$6 \\
7 & 1-shot planned example & 74\% & 30\% & 52\% & $-$8 \\
8 & Text self-correct instruction & 78\% & 24\% & 51\% & $-$9 \\
9 & 1-shot PC recovery (concise) & \textbf{86\%} & 36\% & \textbf{61\%} & $+$1 \\
10 & 1-shot PC recovery (verbose NL) & 80\% & 30\% & 55\% & $-$5 \\
\bottomrule
\end{tabular}
\end{table}

\textbf{No prompt intervention significantly outperforms baseline.}
The baseline achieves 60\% overall (80\% medium, 40\% hard).
The best intervention, concise PC recovery (condition~9), reaches 61\%, a difference well within sampling noise at $N{=}100$.
Every other condition underperforms baseline, with the largest deficits on hard puzzles.

\textbf{Hard accuracy degrades with prompt length.}
The most consistent pattern is that longer prompts correlate with worse hard-tier accuracy.
The baseline prompt achieves 40\% on hard puzzles; adding even a single instruction sentence drops hard accuracy to 24--30\%.
One-shot examples range from 28--42\% on hard, with the most compact format (gold path, no NL reasoning) performing best.
Two-shot examples, the longest prompts tested, achieve only 30\% on hard.
We hypothesize that additional prompt content competes with the model's own extended-thinking budget for context window capacity, and that this competition is most damaging on hard problems where reasoning chains are longest.

\textbf{Abstract instructions are ignored.}
All three text-only instructions (backtrack, planning, self-correct) degrade hard accuracy by 10--16pp despite adding minimal tokens.
The model's extended-thinking process appears to operate independently of prompt-level strategy advice, consistent with findings that reasoning models tend to follow their own internal problem-solving strategies rather than conforming to instructed approaches.

\textbf{Interpretation.}
For reasoning models with extended thinking, the model's internal problem-solving process dominates over prompt-level guidance.
Worked examples and strategy instructions may anchor the model to approaches from simpler examples that fail to generalize to harder instances, creating rigidity rather than flexibility.
PC-prevention approaches (planning instructions and planned examples) consistently degrade hard accuracy by 8--10pp, while the concise PC-recovery example roughly matches baseline, suggesting that demonstrating \emph{recovery from} mistakes is less harmful than demonstrating \emph{avoidance of} mistakes, though neither significantly improves performance.
Prompt-level changes alone do not elicit meaningful planning or recovery behavior from the model.
We believe that enabling robust search and backtracking on hard combinatorial problems may require training on puzzle-solving data, potentially combined with reinforcement learning, to internalize the exploration and recovery mechanisms that prompt guidance fails to activate.

\section{Tool-Augmented Reasoning: Details}
\label{app:tool_augmented}

This appendix provides full tool specifications, the complete ablation table, trace-level statistics, and an extended discussion of the spatial-vs-structured distinction for the tool-augmented reasoning experiments in Section~\ref{subsec:state-externalization}.

\subsection{Tool Specifications}

The model interacts with the puzzle engine through five tools divided into two categories.

\textbf{State-mutation tools.}
\begin{itemize}[nosep]
\item \texttt{make\_move}$(i_1, i_2, n)$: places $n$ bridges between islands $i_1$ and $i_2$.
The engine validates the move against puzzle rules (capacity, corridor clearance) and rejects illegal moves with an error message.
\item \texttt{render\_board}: returns the current board state as an ASCII grid identical in format to the original puzzle input. Bridge characters encode single (\texttt{-}, \texttt{|}) and double (\texttt{=}, \texttt{"}) connections.
\end{itemize}

\textbf{Constraint-query tools.}
\begin{itemize}[nosep]
\item \texttt{state\_summary}: for each island, reports the required bridge count, current count, and remaining count, plus the full edge list and overflow detection flags. Returns JSON.
\item \texttt{neighbors}: for each island, enumerates all legal moves with corridor clearance status and joint capacity (maximum additional bridges possible given both endpoints' remaining counts). Returns JSON.
\item \texttt{components}: returns the list of connected components, flags islands at isolation risk (components with no remaining outward capacity), and lists candidate cross-component links that could merge components. Returns JSON.
\end{itemize}

No tool solves the puzzle, suggests an optimal move, or performs search.
The model must decide which bridges to place and in what order.

\subsection{Full Ablation Results}

Table~\ref{tab:tool_ablation} reports the complete ablation on hard Bridges ($N{=}50$), extending the summary in Table~\ref{tab:tool_ablation_main}.
The additive ablation starts from state-mutation tools only and progressively adds each constraint-query tool.
The grid removal experiments test whether \texttt{render\_board} helps or hurts when structured alternatives are available.

\begin{table}[ht]
\centering
\caption{Full tool ablation on hard Bridges ($N{=}50$).
\emph{Top}: additive ablation adding one tool per row.
\emph{Bottom}: grid removal variants testing whether \texttt{render\_board} helps when structured tools are available.}
\label{tab:tool_ablation}

\resizebox{\textwidth}{!}{%
\begin{tabular}{llrrr}
\toprule
Condition & Tools available & Accuracy & Board valid & Avg tokens \\
\midrule
Baseline (no tools) & --- & 40\% & 50\% & 25{,}640 \\
\midrule
\multicolumn{5}{l}{\emph{Additive ablation}} \\
\, State mutation only & \texttt{make\_move}, \texttt{render\_board} & 26\% & 96\% & 28{,}308 \\
\, + degree bookkeeping & + \texttt{state\_summary} & 42\% & 100\% & 22{,}784 \\
\, + move enumeration & + \texttt{neighbors} & 44\% & 100\% & 19{,}570 \\
\, + connectivity & + \texttt{components} & 48\% & 98\% & 18{,}030 \\
\, Full suite & + candidate links & 50\% & 100\% & 17{,}571 \\
\midrule
\multicolumn{5}{l}{\emph{Grid removal}} \\
\, Structured only (2 tools) & \texttt{make\_move}, \texttt{state\_summary} & 46\% & 100\% & 27{,}301 \\
\, Structured only (4 tools) & all except \texttt{render\_board} & 50\% & 100\% & 22{,}294 \\
\bottomrule
\end{tabular}%
}

\end{table}
\paragraph{Additive ablation analysis.}
With only state-mutation tools (\texttt{make\_move} and \texttt{render\_board}), accuracy drops to 26\%, well below the 40\% no-tool baseline, despite board validity rising from 50\% to 96\%.
This paradox arises because the multi-turn tool interface introduces overhead---commit-and-regret spirals, repeated backtracking, growing context pollution---that spatial state information alone cannot offset.
The model calls \texttt{render\_board} 12.5 times per puzzle on average but cannot reliably extract constraint information from the rendered grid (e.g., counting bridges per island requires scanning four directions and decoding bridge characters).
Adding \texttt{state\_summary}, which provides the same constraint information as pre-computed numeric fields, produces the single largest accuracy gain in the ablation ($+$16\,pp, from 26\% to 42\%).
Subsequent constraint-query tools contribute diminishing returns: $+$2\,pp (\texttt{neighbors}), $+$4\,pp (\texttt{components}), $+$2\,pp (candidate links).

\paragraph{Medium difficulty results.}
On medium Bridges ($7{\times}7$, $N{=}50$), the full tool suite improves accuracy by $+$12\,pp (80\%\,$\to$\,92\%) while reducing token consumption from 14{,}194 to 12{,}603 (11\% reduction).
Board-validity errors, which affect 10\% of medium baseline outputs, are eliminated entirely.

\subsection{Spatial vs.\ Structured Information}

The key design distinction is that state-mutation tools return information in the same spatial format as the original puzzle (an ASCII grid), while constraint-query tools return pre-computed structured data (JSON with numeric fields).
This distinction is central to the ablation results reported in Table~\ref{tab:tool_ablation}.

The ASCII grid is a lossless encoding of game state: all information needed to verify constraint satisfaction is present in principle.
However, extracting actionable constraints from it requires spatial-algorithmic reasoning---scanning in four directions from each island, decoding bridge characters (\texttt{-}, \texttt{=}, \texttt{|}, \texttt{"}), computing degree sums, and checking corridor intersections---that the model performs unreliably.
\texttt{state\_summary} converts this spatial extraction into pre-computed numeric fields (e.g., \texttt{remaining:~2}), shifting the task from perceptual parsing to algebraic comparison.

The constraint-query tools do not merely \emph{reformat} existing information; they provide \emph{derived} quantities---remaining degree, joint capacity, connected components---that are absent from the raw grid and would require algorithmic computation to obtain.
This is why \texttt{state\_summary} produces a 16\,pp gain over state-mutation-only tools, even though both tool sets expose the same underlying board state.

\subsection{Trace-Level Statistics}

Trace analysis reveals how tool availability changes the model's behavior:

\begin{itemize}[nosep]
\item \emph{State mutation only} (\texttt{make\_move} + \texttt{render\_board}): the model calls \texttt{render\_board} 12.5 times per puzzle on average, entering multi-turn loops of trial, error, and backtracking. Counting bridges per island requires scanning four directions and decoding bridge characters from the rendered grid, which the model cannot do reliably.
\item \emph{With \texttt{state\_summary}} (no grid removal): \texttt{render\_board} calls drop to 8 per puzzle on average, but the model continues to call it regardless of prompt instructions to the contrary. These spatial renderings appear to interfere with the algebraic reasoning that \texttt{state\_summary} enables, explaining the 4\,pp accuracy gain when \texttt{render\_board} is physically removed.
\item \emph{Full tool stack}: \texttt{render\_board} calls drop further to 6 per puzzle. At this point the richer constraint tools provide sufficient structured data that the spatial grid becomes redundant (50\% accuracy with or without it).
\end{itemize}

\subsection{Convergence of Evidence}

Three independent lines of evidence from the main body converge on the same conclusion---that spatial constraint extraction, not reasoning ability, is the primary bottleneck:

\begin{enumerate}[nosep]
\item \emph{Format intervention} (Section~\ref{subsec:format-intervention}): switching from ASCII to IntFormat produces large accuracy gains on the same puzzles, by providing cell-aligned tokenization that reduces the spatial parsing burden.
\item \emph{Additive ablation} (Table~\ref{tab:tool_ablation}, top): \texttt{render\_board} (spatial state) degrades accuracy; \texttt{state\_summary} (structured constraints) recovers it. The tools expose the same underlying information in different formats---the structured format is what helps.
\item \emph{Grid removal} (Table~\ref{tab:tool_ablation}, bottom): when structured tools are available, removing the spatial grid either improves accuracy ($+$4\,pp with \texttt{state\_summary} alone) or has no effect (with the full tool stack). The spatial representation is never beneficial.
\end{enumerate}

\section{Full Result Tables}
\label{sec:full_results}

Tables~\ref{tab:accuracy_full_easy}--\ref{tab:accuracy_full_hard} report accuracy across all models, formats, and difficulty tiers.
Rows with average accuracy $\leq 0.01$ are shaded gray.

% ===== FULL TABLES (ONE PER DIFFICULTY) =====

\begin{table}[htbp]
\centering
\small
\caption{Easy accuracy results across format types. intf. json = IntFormat JSON, ascii + img = ASCII + Image. Gray rows indicate average accuracy $\leq 0.01$.}
\label{tab:accuracy_full_easy}
\begin{tabular}{llrrrrrrr}
\toprule
Model & Format & FlowFree & Bridges & Loopy & Galaxies & Undead & Pattern & Avg \\
\midrule
Gemini 3 Flash & ascii & 0.72 & 0.86 & 0.18 & 0.16 & 0.78 & 0.90 & 0.60 \\
Gemini 3 Flash & intformat & 0.70 & 1.00 & 0.40 & 0.98 & 0.60 & 0.94 & 0.77 \\
Gemini 3 Flash & intf. json & 0.72 & 1.00 & 0.54 & 0.98 & 0.12 & 0.96 & 0.72 \\
Gemini 3 Flash & ascii + img & 0.74 & 0.92 & 0.08 & 0.16 & 0.60 & 0.92 & 0.57 \\
GPT-5 Mini & ascii & 0.66 & 0.90 & 0.32 & 0.58 & 0.88 & 0.92 & 0.71 \\
GPT-5 Mini & intformat & 0.72 & 0.98 & 0.66 & 0.88 & 0.98 & 0.90 & 0.85 \\
GPT-5 Mini & intf. json & 0.70 & 0.98 & 0.72 & 0.90 & 0.92 & 0.92 & 0.86 \\
GPT-5 Mini & ascii + img & 0.62 & 0.88 & 0.38 & 0.56 & 0.96 & 0.76 & 0.69 \\
DeepSeek V3.2 & ascii & 0.64 & 0.94 & 0.26 & 0.04 & 0.76 & 0.82 & 0.58 \\
DeepSeek V3.2 & intformat & 0.66 & 1.00 & 0.48 & 0.78 & 0.32 & 0.66 & 0.65 \\
DeepSeek V3.2 & intf. json & 0.60 & 1.00 & 0.36 & 0.68 & 0.16 & 0.72 & 0.59 \\
qwen3-235b & ascii & 0.48 & 0.72 & 0.00 & 0.00 & 0.24 & 0.44 & 0.31 \\
qwen3-235b & intformat & 0.64 & 0.94 & 0.00 & 0.24 & 0.00 & 0.18 & 0.33 \\
qwen3-235b & intf. json & 0.56 & 0.94 & 0.00 & 0.12 & 0.02 & 0.50 & 0.36 \\
GLM-4.7 Flash & ascii & 0.14 & 0.00 & 0.04 & 0.00 & 0.00 & 0.02 & 0.03 \\
GLM-4.7 Flash & intformat & 0.32 & 0.00 & 0.04 & 0.00 & 0.02 & 0.00 & 0.06 \\
GLM-4.7 Flash & intf. json & 0.22 & 0.00 & 0.02 & 0.00 & 0.02 & 0.00 & 0.04 \\
\rowcolor{gray!20} Llama 4 Maverick & ascii & 0.00 & 0.00 & 0.00 & 0.00 & 0.00 & 0.00 & 0.00 \\
Llama 4 Maverick & intformat & 0.10 & 0.00 & 0.00 & 0.00 & 0.00 & 0.00 & 0.02 \\
\rowcolor{gray!20} Llama 4 Maverick & intf. json & 0.00 & 0.00 & 0.00 & 0.00 & 0.00 & 0.00 & 0.00 \\
Mercury & ascii & 0.10 & 0.20 & 0.00 & 0.00 & 0.02 & 0.08 & 0.07 \\
Mercury & intformat & 0.28 & 0.68 & 0.00 & 0.00 & 0.00 & 0.00 & 0.16 \\
Mercury & intf. json & 0.22 & 0.50 & 0.00 & 0.00 & 0.00 & 0.00 & 0.12 \\
OLMo 3.1 32B & ascii & 0.24 & 0.08 & 0.02 & 0.00 & 0.00 & 0.08 & 0.07 \\
OLMo 3.1 32B & intformat & 0.36 & 0.08 & 0.07 & 0.00 & 0.00 & 0.00 & 0.08 \\
OLMo 3.1 32B & intf. json & 0.20 & 0.08 & 0.11 & 0.00 & 0.00 & 0.00 & 0.06 \\
Qwen3-32B & ascii & 0.18 & 0.06 & 0.00 & 0.00 & 0.02 & 0.16 & 0.07 \\
Qwen3-32B & intformat & 0.22 & 0.50 & 0.00 & 0.00 & 0.00 & 0.00 & 0.12 \\
Qwen3-32B & intf. json & 0.20 & 0.48 & 0.00 & 0.00 & 0.00 & 0.00 & 0.11 \\
\bottomrule
\end{tabular}
\end{table}

\begin{table}[htbp]
\centering
\small
\caption{Medium accuracy results across format types. intf. json = IntFormat JSON, ascii + img = ASCII + Image. Gray rows indicate average accuracy $\leq 0.01$.}
\label{tab:accuracy_full_medium}
\begin{tabular}{llrrrrrrr}
\toprule
Model & Format & FlowFree & Bridges & Loopy & Galaxies & Undead & Pattern & Avg \\
\midrule
Gemini 3 Flash & ascii & 0.54 & 0.66 & 0.00 & 0.00 & 0.10 & 0.78 & 0.35 \\
Gemini 3 Flash & intformat & 0.70 & 1.00 & 0.00 & 0.14 & 0.10 & 0.78 & 0.45 \\
Gemini 3 Flash & intf. json & 0.68 & 1.00 & 0.00 & 0.06 & 0.00 & 0.86 & 0.43 \\
Gemini 3 Flash & ascii + img & 0.62 & 0.70 & 0.00 & 0.00 & 0.14 & 0.82 & 0.38 \\
GPT-5 Mini & ascii & 0.22 & 0.72 & 0.00 & 0.00 & 0.80 & 0.90 & 0.44 \\
GPT-5 Mini & intformat & 0.38 & 0.90 & 0.02 & 0.16 & 0.72 & 0.70 & 0.48 \\
GPT-5 Mini & intf. json & 0.32 & 0.90 & 0.00 & 0.20 & 0.82 & 0.82 & 0.51 \\
GPT-5 Mini & ascii + img & 0.30 & 0.78 & 0.00 & 0.00 & 0.78 & 0.80 & 0.44 \\
DeepSeek V3.2 & ascii & 0.22 & 0.88 & 0.00 & 0.00 & 0.46 & 0.64 & 0.37 \\
DeepSeek V3.2 & intformat & 0.34 & 1.00 & 0.00 & 0.10 & 0.22 & 0.52 & 0.36 \\
DeepSeek V3.2 & intf. json & 0.34 & 0.96 & 0.00 & 0.06 & 0.06 & 0.58 & 0.33 \\
qwen3-235b & ascii & 0.18 & 0.34 & 0.00 & 0.00 & 0.02 & 0.20 & 0.12 \\
qwen3-235b & intformat & 0.28 & 0.70 & 0.00 & 0.00 & 0.00 & 0.08 & 0.18 \\
qwen3-235b & intf. json & 0.30 & 0.66 & 0.00 & 0.00 & 0.00 & 0.22 & 0.20 \\
\rowcolor{gray!20} GLM-4.7 Flash & ascii & 0.00 & 0.00 & 0.02 & 0.00 & 0.00 & 0.00 & 0.00 \\
\rowcolor{gray!20} GLM-4.7 Flash & intformat & 0.02 & 0.00 & 0.00 & 0.00 & 0.00 & 0.00 & 0.00 \\
\rowcolor{gray!20} GLM-4.7 Flash & intf. json & 0.02 & 0.00 & 0.00 & 0.00 & 0.04 & 0.00 & 0.01 \\
\rowcolor{gray!20} Llama 4 Maverick & ascii & 0.00 & 0.00 & 0.00 & 0.00 & 0.00 & 0.00 & 0.00 \\
\rowcolor{gray!20} Llama 4 Maverick & intformat & 0.00 & 0.00 & 0.00 & 0.00 & 0.00 & 0.00 & 0.00 \\
\rowcolor{gray!20} Llama 4 Maverick & intf. json & 0.02 & 0.00 & 0.00 & 0.00 & 0.00 & 0.00 & 0.00 \\
\rowcolor{gray!20} Mercury & ascii & 0.00 & 0.00 & 0.00 & 0.00 & 0.00 & 0.00 & 0.00 \\
Mercury & intformat & 0.02 & 0.18 & 0.00 & 0.00 & 0.00 & 0.00 & 0.03 \\
Mercury & intf. json & 0.02 & 0.10 & 0.00 & 0.00 & 0.00 & 0.00 & 0.02 \\
\rowcolor{gray!20} OLMo 3.1 32B & ascii & 0.00 & 0.00 & 0.00 & 0.00 & 0.00 & 0.04 & 0.01 \\
\rowcolor{gray!20} OLMo 3.1 32B & intformat & 0.02 & 0.02 & 0.02 & 0.00 & 0.00 & 0.00 & 0.01 \\
OLMo 3.1 32B & intf. json & 0.06 & 0.02 & 0.00 & 0.00 & 0.00 & 0.02 & 0.02 \\
\rowcolor{gray!20} Qwen3-32B & ascii & 0.00 & 0.00 & 0.00 & 0.00 & 0.00 & 0.00 & 0.00 \\
\rowcolor{gray!20} Qwen3-32B & intformat & 0.02 & 0.02 & 0.00 & 0.00 & 0.00 & 0.00 & 0.01 \\
\rowcolor{gray!20} Qwen3-32B & intf. json & 0.00 & 0.02 & 0.00 & 0.00 & 0.00 & 0.00 & 0.00 \\
\bottomrule
\end{tabular}
\end{table}

\begin{table}[htbp]
\centering
\small
\caption{Hard accuracy results across format types. intf. json = IntFormat JSON, ascii + img = ASCII + Image. Gray rows indicate average accuracy $\leq 0.01$.}
\label{tab:accuracy_full_hard}
\begin{tabular}{llrrrrrrr}
\toprule
Model & Format & FlowFree & Bridges & Loopy & Galaxies & Undead & Pattern & Avg \\
\midrule
Gemini 3 Flash & ascii & 0.02 & 0.22 & 0.00 & 0.00 & 0.00 & 0.30 & 0.09 \\
Gemini 3 Flash & intformat & 0.12 & 0.86 & 0.00 & 0.00 & 0.00 & 0.36 & 0.22 \\
Gemini 3 Flash & intf. json & 0.18 & 0.90 & 0.00 & 0.00 & 0.00 & 0.30 & 0.23 \\
Gemini 3 Flash & ascii + img & 0.06 & 0.34 & 0.00 & 0.00 & 0.00 & 0.44 & 0.14 \\
GPT-5 Mini & ascii & 0.02 & 0.44 & 0.00 & 0.00 & 0.52 & 0.44 & 0.24 \\
GPT-5 Mini & intformat & 0.04 & 0.54 & 0.00 & 0.00 & 0.24 & 0.00 & 0.14 \\
GPT-5 Mini & intf. json & 0.02 & 0.64 & 0.00 & 0.00 & 0.24 & 0.14 & 0.17 \\
GPT-5 Mini & ascii + img & 0.00 & 0.38 & 0.00 & 0.00 & 0.30 & 0.16 & 0.14 \\
DeepSeek V3.2 & ascii & 0.00 & 0.40 & 0.00 & 0.00 & 0.10 & 0.12 & 0.10 \\
DeepSeek V3.2 & intformat & 0.02 & 0.74 & 0.00 & 0.00 & 0.00 & 0.14 & 0.15 \\
DeepSeek V3.2 & intf. json & 0.02 & 0.68 & 0.00 & 0.00 & 0.00 & 0.04 & 0.12 \\
\rowcolor{gray!20} qwen3-235b & ascii & 0.00 & 0.04 & 0.00 & 0.00 & 0.00 & 0.02 & 0.01 \\
qwen3-235b & intformat & 0.02 & 0.08 & 0.00 & 0.00 & 0.00 & 0.00 & 0.02 \\
qwen3-235b & intf. json & 0.02 & 0.16 & 0.00 & 0.00 & 0.00 & 0.00 & 0.03 \\
\rowcolor{gray!20} GLM-4.7 Flash & ascii & 0.00 & 0.00 & 0.00 & 0.00 & 0.02 & 0.02 & 0.01 \\
\rowcolor{gray!20} GLM-4.7 Flash & intformat & 0.00 & 0.00 & 0.00 & 0.00 & 0.00 & 0.00 & 0.00 \\
\rowcolor{gray!20} GLM-4.7 Flash & intf. json & 0.00 & 0.00 & 0.00 & 0.00 & 0.00 & 0.02 & 0.00 \\
\rowcolor{gray!20} Llama 4 Maverick & ascii & 0.00 & 0.00 & 0.00 & 0.00 & 0.00 & 0.00 & 0.00 \\
\rowcolor{gray!20} Llama 4 Maverick & intformat & 0.00 & 0.00 & 0.00 & 0.00 & 0.00 & 0.00 & 0.00 \\
\rowcolor{gray!20} Llama 4 Maverick & intf. json & 0.00 & 0.00 & 0.00 & 0.00 & 0.00 & 0.00 & 0.00 \\
\rowcolor{gray!20} Mercury & ascii & 0.00 & 0.00 & 0.00 & 0.00 & 0.00 & 0.02 & 0.00 \\
\rowcolor{gray!20} Mercury & intformat & 0.00 & 0.00 & 0.00 & 0.00 & 0.00 & 0.00 & 0.00 \\
\rowcolor{gray!20} Mercury & intf. json & 0.00 & 0.02 & 0.00 & 0.00 & 0.00 & 0.00 & 0.00 \\
\rowcolor{gray!20} OLMo 3.1 32B & ascii & 0.00 & 0.00 & 0.00 & 0.00 & 0.00 & 0.04 & 0.01 \\
\rowcolor{gray!20} OLMo 3.1 32B & intformat & 0.00 & 0.00 & 0.00 & 0.00 & 0.00 & 0.00 & 0.00 \\
\rowcolor{gray!20} OLMo 3.1 32B & intf. json & 0.00 & 0.00 & 0.00 & 0.00 & 0.00 & 0.00 & 0.00 \\
\rowcolor{gray!20} Qwen3-32B & ascii & 0.00 & 0.00 & 0.00 & 0.00 & 0.00 & 0.00 & 0.00 \\
\rowcolor{gray!20} Qwen3-32B & intformat & 0.00 & 0.00 & 0.00 & 0.00 & 0.00 & 0.00 & 0.00 \\
\rowcolor{gray!20} Qwen3-32B & intf. json & 0.00 & 0.00 & 0.00 & 0.00 & 0.00 & 0.00 & 0.00 \\
\bottomrule
\end{tabular}
\end{table}

\section{Cost}
In this section we provide estimates for inference cost, based only on the output token cost and ignoring the marginal input token cost.

\begin{table}[htbp]
\centering
\small
\setlength{\tabcolsep}{10pt}
\renewcommand{\arraystretch}{1.2}
\begin{tabular}{lrrr}
\toprule
Model & Tokens & Price per 1M tokens (USD) & Full price (USD) \\
\midrule
gpt-5-mini & 49,607,682 & 2.00 & 99.22 \\
qwen3-235b & 36,252,844 & 2.30 & 83.38 \\
gemini-3-flash-preview & 27,014,575 & 3.00 & 81.04 \\
deepseek-v3.2 & 30,766,175 & 0.42 & 12.92 \\
glm-4.7-flash & 26,939,558 & 0.40 & 10.78 \\
olmo-3.1-32b-think & 19,736,186 & 0.50 & 9.87 \\
qwen3-32b & 13,985,471 & 0.24 & 3.36 \\
llama-4-maverick-17b-128e-instruct & 1,148,478 & 0.60 & 0.69 \\
mercury & 513,078 & 1.00 & 0.51 \\
\bottomrule
\end{tabular}
\caption{Cost of running the main benchmark}
\label{tab:cost_main_benchmark}
\end{table}

\begin{table}[htbp]
\centering
\small
\setlength{\tabcolsep}{10pt}
\renewcommand{\arraystretch}{1.2}
\begin{tabular}{lrrr}
\toprule
Model & Tokens & Price per 1M tokens (USD) & Full price (USD) \\
\midrule
gpt-5-mini & 191,749,525 & 2.00 & 383.50 \\
gemini-3-flash-preview & 109,343,674 & 3.00 & 328.03 \\
qwen3-235b & 108,633,225 & 2.30 & 249.86 \\
deepseek-v3.2 & 91,074,375 & 0.42 & 38.25 \\
glm-4.7-flash & 82,757,329 & 0.40 & 33.10 \\
olmo-3.1-32b-think & 58,457,735 & 0.50 & 29.23 \\
qwen3-32b & 43,261,103 & 0.24 & 10.38 \\
llama-4-maverick-17b-128e-instruct & 6,156,117 & 0.60 & 3.69 \\
mercury & 3,383,521 & 1.00 & 3.38 \\
\bottomrule
\end{tabular}
\caption{Cost of running the input format ablations}
\label{tab:cost_input_format_ablations}
\end{table}

\section{Input formats}

% ---------------

\begin{figure}[H]
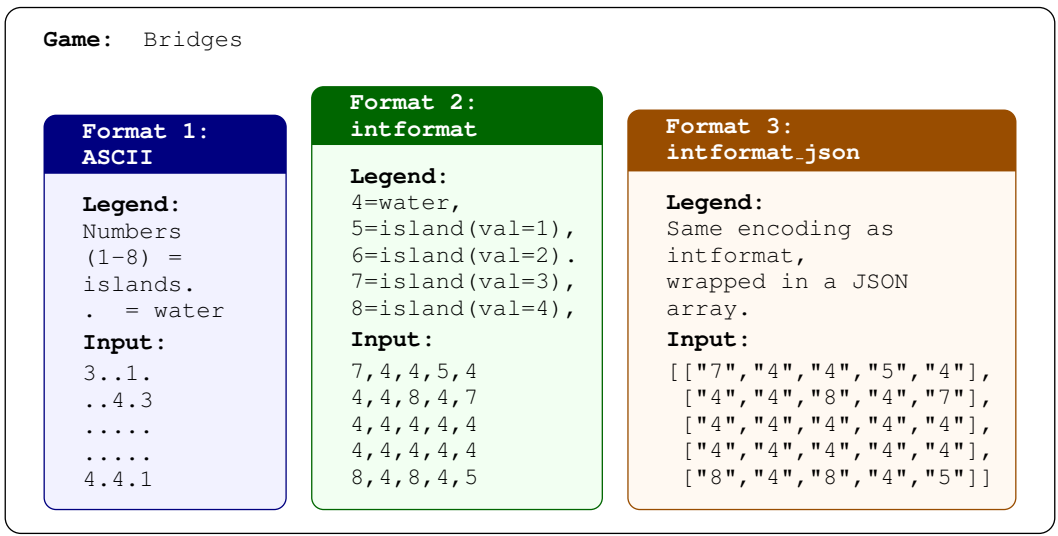

\noindent
\begin{tcolorbox}[
  colback=white,
  colframe=black,
  arc=2mm,
  boxrule=0.5pt,
  width=\textwidth,
  toptitle=1pt,
  bottomtitle=1pt,
  fontupper=\small,
]
\ttfamily
\textbf{Game:} Bridges \\[4pt]

\noindent
\begin{minipage}[t]{0.25\textwidth}
\begin{tcolorbox}[
  colback=blue!5,
  colframe=blue!50!black,
  arc=1.5mm,
  boxrule=0.5pt,
  title={\small\bfseries\ttfamily Format 1: ASCII},
  fonttitle=\small,
  fontupper=\small\ttfamily,
]
\textbf{Legend:}\\
Numbers (1-8) = islands.\\
\texttt{.} = water\\[2pt]
\textbf{Input:}\\[2pt]
3..1.\\
..4.3\\
.....\\
.....\\
4.4.1
\end{tcolorbox}
\end{minipage}
\hfill
\begin{minipage}[t]{0.3\textwidth}
\begin{tcolorbox}[
  colback=green!5,
  colframe=green!40!black,
  arc=1.5mm,
  boxrule=0.5pt,
  title={\small\bfseries\ttfamily Format 2: intformat},
  fonttitle=\small,
  fontupper=\small\ttfamily,
]
\textbf{Legend:}\\
\texttt{4}=water, \texttt{5}=island(val=1),\\
\texttt{6}=island(val=2).\\
\texttt{7}=island(val=3), \texttt{8}=island(val=4),\\[2pt]
\textbf{Input:}\\[2pt]
7,4,4,5,4\\
4,4,8,4,7\\
4,4,4,4,4\\
4,4,4,4,4\\
8,4,8,4,5
\end{tcolorbox}
\end{minipage}
\hfill
\begin{minipage}[t]{0.4\textwidth}
\begin{tcolorbox}[
  colback=orange!5,
  colframe=orange!60!black,
  arc=1.5mm,
  boxrule=0.5pt,
  title={\small\bfseries\ttfamily Format 3: intformat\_json},
  fonttitle=\small,
  fontupper=\small\ttfamily,
]
\textbf{Legend:}\\
Same encoding as intformat,\\
wrapped in a JSON array.\\[2pt]
\textbf{Input:}\\[2pt]
[["7","4","4","5","4"],\\
\phantom{x}["4","4","8","4","7"],\\
\phantom{x}["4","4","4","4","4"],\\
\phantom{x}["4","4","4","4","4"],\\
\phantom{x}["8","4","8","4","5"]]
\end{tcolorbox}
\end{minipage}

\end{tcolorbox}
\caption{The same Bridges puzzle represented in three input formats: ASCII, intformat, and intformat\_json.}
\label{fig:prompt_bridges}
\end{figure}

\section{Error-Analysis Judge Prompt}
\label{app:judge_prompt}

% \begin{figure*}[!ht]
% \noindent
\begin{center}
\begin{tcolorbox}[
  breakable,
  colback=white,
  colframe=black,
  arc=2mm,
  boxrule=0.5pt,
  width=\textwidth,
  toptitle=1pt,
  bottomtitle=1pt,
  fontupper=\tiny,
]
\ttfamily
You are an expert evaluator analyzing reasoning quality in large language models on logic-based puzzle tasks.\\

You will be given the following sections:\\
1. \textbf{ORIGINAL PROMPT GIVEN TO MODEL}: The exact instructions and rules that the AI was given to solve the puzzle.\\
2. \textbf{ORIGINAL PUZZLE PROBLEM}: The puzzle type, difficulty, and the problem board that needed to be solved.\\
3. \textbf{CORRECT SOLUTION}: The actual correct solution to the puzzle for comparison.\\
4. \textbf{REASONING TRACE}: The AI's step-by-step thinking process (chain-of-thought).\\
5. \textbf{FULL RESPONSE}: The AI's complete final output shown to the user.\\

Note: You must analyze the reasoning quality based solely on the reasoning trace and how well the model followed the puzzle rules and instructions.\\

Your task is to analyze the reasoning trace and the final response to identify the main behaviour modes in the model's reasoning.
For each behaviour you report, please provide 15 characters as they appear in one instance of that behaviour from the reasoning trace. This is to enable locating the behaviour in the chain of thought easily. Please put the characters as MARKER:"\{15\_chars\}".\\

\textbf{BEHAVIOUR CLASSIFICATION INSTRUCTIONS}\\

Report zero to three distinct behaviour categories that best represent the model's main behaviour.
Base your judgments on observable behavior in the reasoning trace and final output.
Behaviour categories must be denoted exactly as written below, using the bold constant-style names.\\

\textbf{BEHAVIOUR CATEGORIES}\\

\textbf{REPEATED\_REASONING\_PATHS}\\
The model executes a near-identical sequence of moves or reasoning steps that it has already tried earlier from the same or equivalent board position, without meaningfully varying its approach. Look for:\\
- The model backtracks, then re-executes the same (or near-identical) sequence of moves it already tried from that position.\\
- The model revisits a previously explored configuration and follows essentially the same reasoning path.\\
- \textbf{Important:} Normal backtracking and exploration of DIFFERENT paths from the same state is NOT repeated reasoning. Only flag when the model is genuinely looping---trying the same approach again without meaningful variation.\\
- A model that tries path A, backtracks, then tries path B from the same state is performing normal search, NOT repeated reasoning.\\

\textbf{STATE\_TRACKING\_FAILURE}\\
The model's claimed board state diverges from what its own sequence of actions would produce. Look for:\\
- The model writes out an intermediate grid/board that is inconsistent with the moves it has described (e.g., an element it placed earlier is missing, or the grid shows an element it never placed).\\
- The model states a cell or connection has a certain value, but its own prior actions would have set it to something different.\\
- The model's description of the board at step N contradicts the cumulative effect of all moves through step N.\\
- \textbf{Do NOT flag} cases where the model merely restates an already-placed element without changing state. Only flag genuine divergence between claimed state and the result of stated actions.\\

\textbf{CONSTRAINT\_FORGETTING}\\
The model attempts a specific action that directly violates the puzzle's structural rules. Concrete violation types include:\\
- \textbf{Overcounting:} exceeding a cell's or node's required count (e.g., too many connections on a node, wrong number of edges around a cell).\\
- \textbf{Crossing:} placing an element that crosses or overlaps an existing one where the rules forbid it.\\
- \textbf{Invalid connection:} connecting cells or nodes that cannot be connected per the rules (e.g., non-adjacent nodes, diagonal where only orthogonal is allowed).\\
- \textbf{Overwriting fixed cells:} modifying a pre-filled or immutable element.\\
- \textbf{The violation must be in an explicit action the model proposes to execute}, not merely in speculative reasoning it later discards. If the model considers and then rejects an invalid move in prose, that is NOT constraint forgetting.\\
- Misapplying a logical deduction (drawing a wrong conclusion from correct constraints) is NOT constraint forgetting---it is a reasoning error, not a structural rule violation.\\

\textbf{PREMATURE\_COMMITMENT}\\
The model starts a reasoning path with a move or assumption that is provably incorrect, and persists down that path for multiple steps before either (a) eventually backtracking or (b) incorporating the wrong assumption into its final answer. Look for:\\
- The model's first move in a new exploration branch places an element that differs from the correct solution at that position, and continues for 3+ additional steps down that path.\\
- The model commits to a particular configuration early and continues building on it despite encountering constraints that suggest it may be wrong.\\
- In the most severe form (committed PC), the model's final answer includes elements from this wrong initial commitment.\\
- \textbf{Note:} If the correct solution is available, check whether the model's early moves match it. A move contradicting the correct answer that leads to a long dead-end exploration is premature commitment.\\

\textbf{EXPLICIT\_SURRENDER}\\
The model explicitly acknowledges inability to complete the task by requesting permission to use a programmatic solver, asking for more time, requesting relaxed constraints, or otherwise stating it cannot proceed as required.\\

\textbf{INCOMPLETE\_OUTPUT}\\
The model fails to produce a complete, valid final answer. Examples include partially filled solutions, placeholder symbols (e.g., dots or blanks), returning the original unsolved puzzle, or providing only a fragment of the required output.\\

\textbf{UNJUSTIFIED\_ELIMINATION}\\
The model prematurely eliminates valid possibilities, paths, or configurations without sufficient justification, leading to dead ends or incorrect conclusions about solvability.\\

\textbf{HALLUCINATED\_VALIDATION}\\
The model asserts that a solution is correct, complete, or satisfies all constraints without proper verification, or despite the presence of contradictions, invalid configurations, or failed verification signals.\\

\textbf{REPRESENTATION\_DRIFT}\\
The model's internal representation of the problem changes over time (e.g., reinterpretation of symbols, coordinates, endpoints, or rules), resulting in inconsistent reasoning across different parts of the trace.\\

\textbf{TOPIC\_DRIFT}\\
The model shifts focus away from the original problem domain and does not answer the original task.\\

\textbf{OTHER}\\
Use only if the observed failure does not reasonably fit any category above. If used, briefly describe the additional failure mode and why it is distinct.
\end{tcolorbox}
\captionof{figure}{LLM-as-judge prompt used for error-analysis classification of reasoning traces. The judge receives the original puzzle prompt, problem, correct solution, reasoning trace, and full response, then labels up to three behaviour categories from the taxonomy.}
\label{fig:judge_prompt}
\end{center}
% \end{figure*}

\section{Evaluation Prompts}
\label{app:eval_prompts}

% include \usepackage{tcolorbox}  in the main file start

\begin{center}
\begin{tcolorbox}[colback=white,colframe=black,arc=2mm,boxrule=0.5pt,
  width=\linewidth,
  left=2mm,right=2mm,top=1mm,bottom=1mm, % tighten padding
  fontupper=\small
]
\ttfamily
\textbf{Game:} Bridges (also known as Hashi or Hashiwokakero) \\

Solve the following Bridges puzzle by connecting islands with bridges. You are given a 2D ASCII board representation.\\

\textbf{Legend:} \\

A grid where each cell is either:
- A number (1 - 8) representing an island with that many required bridges. \\
- A dot (.) representing an empty cell/water. \\

Provide the completed grid with bridges represented by: \\
  - `-` for horizontal bridges.\\
  - `|` for vertical bridges.\\
  - `=` for horizontal double bridges.\\
  - `"` for vertical double bridges.\\

\textbf{Rules:} \\

Connect all of the islands with bridges such that:\\
The number of bridges connected to each island matches the number on that island.\\
Bridges only run horizontally or vertically.\\
Bridges must not cross other bridges or islands.\\
A maximum of two bridges can connect any pair of islands.\\
All islands must be part of a single connected group.\\

Think step by step then output only the solved board in json format as shown below.\\

\textbf{Output Format:}\\

Return your final answer exactly like this:\\
```json \\
\{"response": "\{final board state\}"\}```\\

\textbf{Example Puzzle:}\\

\textbf{Input:}\\

3..1.\\
..4.3\\
.....\\
.....\\
4.4.1\\

\textbf{Solution:}\\

```json\\
{"response": "\\
3--1.\\
".4=3\\
".".|\\
".".|\\
4=4.1\\
"}
```\\

\textbf{now solve this puzzle:}
\end{tcolorbox}

\captionof{figure}{Prompt used for evaluating Bridges puzzles}
\label{fig:bridges_prompt}
\end{center}

% -----------------------

\begin{figure*}[!ht]
\noindent
\begin{tcolorbox}[
  colback=white,
  colframe=black,
  arc=2mm,
  boxrule=0.5pt,
  width=\textwidth,
  toptitle=1pt,
  bottomtitle=1pt,
  fontupper=\small,
  % title=\bfseries FlowFree Prompt
]
\ttfamily
\textbf{Game:} Flow Free \\

Solve the following Flow Free puzzle by connecting the same letters. You are given a 2D ASCII board representation.\\

\textbf{Legend:}\\

Letters (A-Z) = colored dots that need to be connected\\
. = empty space that can be filled with letters to create paths\\

\textbf{Rules:}\\

1. You may only fill in dots (.) with letters that already exist on the board.\\
2. You cannot modify or move any of the existing letters.\\
3. Each pair of identical letters must be connected with a continuous, unbroken path of the same letter.\\
4. Paths cannot cross or overlap each other.\\
5. When solved, no dots should remain — the board must be completely filled with letters.\\
6. A path cannot be adjacent to itself horizontally or vertically.\\

Think step by step then output only the solved board in json format as shown below.\\

\textbf{Output Format:}\\

Return your final answer exactly like this:\\
```json \\
\{"response": "\{final board state\}"\}```\\

\textbf{Example Puzzle:}\\

\textbf{Input:}\\
BA...A\\
..E...\\
..D.F.\\
..F..D\\
..C.CE\\
B.....\\

\textbf{Solution:}\\
```json \\
{"response": "\\
BAAAAA\\
BEEDDD\\
BEDDFD\\
BEFFFD\\
BECCCE\\
BEEEEE\\
"}\\
\textbf{now solve this puzzle:}
\end{tcolorbox}
\caption{Prompt used for evaluating FlowFree puzzles}
\label{fig:flowfree_prompt}
\end{figure*}

% -----------------------

\begin{figure*}[!ht]
\noindent
\begin{tcolorbox}[
  colback=white,
  colframe=black,
  arc=2mm,
  boxrule=0.5pt,
  width=\textwidth,
  toptitle=1pt,
  bottomtitle=1pt,
  fontupper=\small,
]
\ttfamily
\textbf{Game:} Galaxies (also known as Spiral Galaxies / Tentai Show) \\
Solve the following Galaxies puzzle by partitioning the grid into rotationally symmetric regions. You are given a 2D ASCII board representation.\\

\textbf{Legend:}\\
Initial grid: o = dot, + = grid vertex, - = horizontal grid line, | = vertical grid line, space ( ) = empty square or non-existent grid line\\
Grid orientation: rows are written top-to-bottom; columns left-to-right. Each row is a string of characters; rows are separated by newline characters (\textbackslash n).\\

\textbf{Rules:} \\
Partition the rectangular grid into connected regions of squares so that:
Every region is 180° rotationally symmetric.\\
Each region contains exactly one dot, and that dot is the region’s centre of symmetry. The dot may be on a square, on an edge between two squares, or at a vertex where four squares meet.\\
Regions are formed by drawing edges on grid lines; the puzzle is complete when the drawn edges separate every pair of squares that belong to different regions.\\
Do not modify or move any of the existing dots or grid lines and vertexes. Only add new horizontal and vertical edges to the grid lines.\\

Think step by step then output only the solved board in json format as shown below.\\

\textbf{Output Format:}\\
Return your final answer exactly like this:\\
```json \\
\{"response": "\{final board state\}"\}```\\

\textbf{Example Puzzle:} \\
\textbf{Input:}
\begin{tabbing}
+-+-+-+-+-+\\
|o\phantom{xxx}o\phantom{xxxx}|\\
+\phantom{x}+\phantom{x}+\phantom{x}+\phantom{x}+\phantom{x}+\\
|\phantom{xxxx}o\phantom{xxxx}|\\
+\phantom{x}+\phantom{x}+\phantom{x}+\phantom{x}+\phantom{x}+\\
|\phantom{xxxxxxxx}o|\\
+\phantom{x}+\phantom{x}+\phantom{x}+\phantom{x}+\phantom{x}+\\
|\phantom{xxxx}o\phantom{xxxx}|\\
+o+o+\phantom{x}+\phantom{x}o\phantom{x}+\\
|\phantom{xxxxxxxxx}|\\
+-+-+-+-+-+\\
\end{tabbing}
\textbf{Solution:}
\begin{tabbing}
```json \\\{"response": "\\
+-+-+-+-+-+\\
|o|\phantom{x}|o|\phantom{xxx}|\\
+-+\phantom{x}+-+\phantom{x}+\phantom{x}+\\
|\phantom{xxxx}o\phantom{xxxx}|\\
+\phantom{x}+\phantom{x}+-+\phantom{x}+-+\\
|\phantom{xxx}|\phantom{x}|\phantom{x}|o|\\
+-+-+\phantom{x}+-+-+\\
|\phantom{x}|\phantom{x}|o|\phantom{xxx}|\\
+o+o+\phantom{x}+\phantom{x}o\phantom{x}+\\
|\phantom{x}|\phantom{x}|\phantom{x}|\phantom{xxx}|\\
+-+-+-+-+-+\\
"\}```\\
\end{tabbing}

\textbf{now solve this puzzle:}\
\end{tcolorbox}
\caption{Prompt used for evaluating Galaxies puzzles}
\label{fig:prompt_galaxies}
\end{figure*}

% --------------------

% -----------------------

\begin{figure*}[!ht]
\noindent
\begin{tcolorbox}[
  colback=white,
  colframe=black,
  arc=2mm,
  boxrule=0.5pt,
  width=\textwidth,
  toptitle=1pt,
  bottomtitle=1pt,
  fontupper=\small,
]
\ttfamily
\textbf{Game:} Pattern \\
Solve the following Pattern puzzle. You are given a 2D ASCII board representation. \\
\textbf{Legend:}\\
"+" for vertices, "-" for horizontal edges, "|" for vertical edges, -Spaces for empty cells\\
-Digits on top and left sides of the grid (called "clues") indicating how many cells in that row or column should form a connected pattern. \\
\textbf{Rules:}\\
Patterns are formed by placing "\#\#" into cells that should be filled and ".." into cells that should stay empty. For example when there are two clues of "3" in a row, it means that there should be two separate groups of three connected \#\# cells in that row. \\
Only place \#\# and .. in empty cells (spaces). Do not modify or move any of the existing +, -, |, or numbers. Each row and column must contain exactly the number of connected \#\# groups as indicated by the clues. The pattern must be continuous (all \#\# cells must be connected horizontally or vertically).\\
Think step by step then output only the solved board in json format as shown below.\\

\textbf{Output Format:}\\

Return your final answer exactly like this:\\
```json \\
\{"response": "\{final board state\}"\}```\\

\textbf{Example Puzzle:}\\

\textbf{Input:}\\
\begin{tabbing}
\phantom{xxxxxx}2\phantom{xx}1\phantom{xx}4\phantom{xx}2\phantom{xx}4\\
\phantom{xxxxx}+--+--+--+--+--+\\
\phantom{xxxx}3|\phantom{xx}|\phantom{xx}|\phantom{xx}|\phantom{xx}|\phantom{xx}|\\
\phantom{xxxxx}+--+--+--+--+--+\\
\phantom{xxxx}3|\phantom{xx}|\phantom{xx}|\phantom{xx}|\phantom{xx}|\phantom{xx}|\\
\phantom{xxxxx}+--+--+--+--+--+\\
\phantom{xx}1\phantom{x}1|\phantom{xx}|\phantom{xx}|\phantom{xx}|\phantom{xx}|\phantom{xx}|\\
\phantom{xxxxx}+--+--+--+--+--+\\
1\phantom{x}1\phantom{x}1|\phantom{xx}|\phantom{xx}|\phantom{xx}|\phantom{xx}|\phantom{xx}|\\
\phantom{xxxxx}+--+--+--+--+--+\\
\phantom{xxxx}2|\phantom{xx}|\phantom{xx}|\phantom{xx}|\phantom{xx}|\phantom{xx}|\\
\phantom{xxxxx}+--+--+--+--+--+\\
\end{tabbing}
\textbf{Solution:}\\
\begin{tabbing}
```json \{"response":"\\
\phantom{xxxxxx}2\phantom{xx}1\phantom{xx}4\phantom{xx}2\phantom{xx}4\\
\phantom{xxxxx}+--+--+--+--+--+\\
\phantom{xxxx}3|..|..|\#\#|\#\#|\#\#|\\
\phantom{xxxxx}+--+--+--+--+--+\\
\phantom{xxxx}3|..|..|\#\#|\#\#|\#\#|\\
\phantom{xxxxx}+--+--+--+--+--+\\
\phantom{xx}1\phantom{x}1|..|..|\#\#|..|\#\#|\\
\phantom{xxxxx}+--+--+--+--+--+\\
1\phantom{x}1\phantom{x}1|\#\#|..|\#\#|..|\#\#|\\
\phantom{xxxxx}+--+--+--+--+--+\\
\phantom{xxxx}2|\#\#|\#\#|..|..|..|\\
\phantom{xxxxx}+--+--+--+--+--+\\
"\}```\\
\end{tabbing}
\textbf{now solve this puzzle:}

\end{tcolorbox}
\caption{Prompt used for evaluating Pattern puzzles}
\label{fig:prompt_pattern}
\end{figure*}

% --------------------

\begin{figure*}[!ht]
\noindent
\begin{tcolorbox}[
  colback=white,
  colframe=black,
  arc=2mm,
  boxrule=0.5pt,
  width=\textwidth,
  toptitle=1pt,
  bottomtitle=1pt,
  fontupper=\small,
]
\ttfamily
\textbf{Game:} Undead \\
Solve the following Undead puzzle. You are given a 2D ASCII board representation.\\

\textbf{Legend:}\\
- '.' for empty squares.\\
- '/' or '\textbackslash' for diagonal mirrors that reflect monster visibility diagonally.\\
- Edge clues (numbers) indicating visible monsters from that direction.\\
- Fill the grid with letters: G, V, or Z.\\

\textbf{Rules:}\\
1. The puzzle is a grid of squares. Some squares contain diagonal mirrors ('/' or '\textbackslash') which cannot hold monsters.\\
2. All non-mirror squares must be filled with exactly one monster:\\
- Ghost (G): visible only in mirrors.\\
- Vampire (V): visible only directly.\\
- Zombie (Z): visible in both direct view and mirrors.\\
3. Total counts of each monster type are provided.\\
4. Numbers around the edges of the grid indicate how many monsters are visible along that row or column from that position, counting diagonal reflections.\\
5. If a reflected line of sight crosses the same monster multiple times, count each occurrence.\\
6. Mirrors reflect light in both directions.\\
Think step by step then output only the solved board in json format as shown below.\\

\textbf{Output Format:}\\
Return your final answer exactly like this:\\
```json \\
\{"response": "\{final board state\}"\}```\\

\textbf{Example Puzzle:}\\

\textbf{Input:}
\begin{tabbing}
G: 2 V: 6 Z: 0\\
\\
\phantom{x}\phantom{x}\phantom{x}2\phantom{x}0\phantom{x}0\phantom{x}0\phantom{xx}\\
\phantom{x}2\phantom{x}.\phantom{x}\textbackslash\phantom{x}\textbackslash\phantom{x}/\phantom{x}0\\
\phantom{x}3\phantom{x}.\phantom{x}.\phantom{x}.\phantom{x}.\phantom{x}3\\
\phantom{x}0\phantom{x}\textbackslash\phantom{x}\textbackslash\phantom{x}\textbackslash\phantom{x}\textbackslash\phantom{x}0\\
\phantom{x}1\phantom{x}.\phantom{x}\textbackslash\phantom{x}.\phantom{x}.\phantom{x}2\\
\phantom{x}\phantom{x}\phantom{x}1\phantom{x}0\phantom{x}0\phantom{x}1\phantom{xx}\\
\end{tabbing}
\textbf{Solution:}
\begin{tabbing}
```json \{"response":"\\

G: 2 V: 6 Z: 0\\
\\

\phantom{x}\phantom{x}\phantom{x}2\phantom{x}0\phantom{x}0\phantom{x}0\phantom{xx}\\
\phantom{x}2\phantom{x}V\phantom{x}\textbackslash\phantom{x}\textbackslash\phantom{x}/\phantom{x}0\\
\phantom{x}3\phantom{x}G\phantom{x}V\phantom{x}V\phantom{x}V\phantom{x}3\\
\phantom{x}0\phantom{x}\textbackslash\phantom{x}\textbackslash\phantom{x}\textbackslash\phantom{x}\textbackslash\phantom{x}0\\
\phantom{x}1\phantom{x}V\phantom{x}\textbackslash\phantom{x}G\phantom{x}V\phantom{x}2\\
\phantom{x}\phantom{x}\phantom{x}1\phantom{x}0\phantom{x}0\phantom{x}1\phantom{xx}\\
"\}```\\
\end{tabbing}
\textbf{now solve this puzzle:}
\end{tcolorbox}
\caption{Prompt used for evaluating Undead puzzles}
\label{fig:undead_prompt}
\end{figure*}

% -----------------------

\begin{figure*}[!ht]
\noindent
\begin{tcolorbox}[
  colback=white,
  colframe=black,
  arc=2mm,
  boxrule=0.5pt,
  width=\textwidth,
  toptitle=1pt,
  bottomtitle=1pt,
  fontupper=\tiny,
]
\ttfamily
\textbf{Game:} Loopy (Slitherlink) \\

Solve the following Loopy (Slitherlink) puzzle given as a 2D ASCII grid.\\

\textbf{Legend:}\\
You are given numbers arranged in a grid that represent the number of loop edges that should be adjacent to their cell. Use the following symbols to solve the puzzle:\\
- x for no edge in that position, | for horizontal and vertical edges, space ( ) for empty cells where no number is given\\

\textbf{Rules:}\\
Draw a single continuous loop using some subset of the possible edges.\\
The loop must: Pass along - and | positions only.\\
Satisfy all cell clues.\\
Have no branches (each vertex used by the loop has degree 2).\\
In the \textbf{Solution:}\\
Use - and | to show edges that are part of the loop and x to show edges that are not part of the loop.
The structure of the puzzle is as follows: empty row, row with clues, empty row, row with clues, ... ending with an empty row.\\
The available positions for horizontal edges (-) are in the empty rows in every second column (the columns where clues can appear).\\
The available positions for vertical edges (|) are in the rows with clues in the first, third, fifth, etc. columns (the columns where no clues appear).\\
In each of the possible positions for edges, you must determine whether to place an edge (- or |) or to mark it as not part of the loop (x).\\

Stay within the grid boundaries denoted as +. Do not modify the grid boundaries.\\

Think step by step then output only the solved board in json format as shown below.\\

\textbf{Output Format:}\\

Return your final answer exactly like this:\\
```json \\

{"response": "{final board state}"} \\

\textbf{Example Puzzle:} \\

\textbf{Input:}\\
+++++++++++++  \\
+ \phantom{xxxxxxxxx} + \\
+ 2 1\phantom{xx} 3\phantom{xx} + \\
+ \phantom{xxxxxxxxx}          + \\
+ 2 \phantom{xx}1\phantom{xxx}2 + \\
+ \phantom{xxxxxxxxx} + \\
+ 3 \phantom{xxxxxxx}      + \\
+  \phantom{xxxxxxxxx}         + \\
+  \phantom{xx}0 \phantom{xx}1 3 + \\
+   \phantom{xxxxxxxxx}        + \\
+   \phantom{xx}3      \phantom{xxxxx} + \\
+  \phantom{xxxxxxxxx}         + \\
+++++++++++++ \\

\textbf{Solution:} \\

```json
{"response": "\\
+++++++++++++\\
+ - - - x - +\\
+|2x1x |3| |+\\
+ x x x - x +\\
+|2x x1x x2|+\\
+ - x - - - +\\
+x3| | x x x+\\
+ - x - - - +\\
+| x0x x1x3|+\\
+ - x - x - +\\
+x |3| | | x+\\
+ x - x - x +\\
+++++++++++++"} \\
``` \\

\textbf{now solve this puzzle:}
\end{tcolorbox}
\caption{Prompt used for evaluating Loopy puzzles}
\label{fig:prompt_loopy}
\end{figure*}

\end{document}